\documentclass[final,5p,times,twocolumn,authoryear]{elsarticle}
\usepackage{url,hyperref,microtype,subcaption}
\usepackage[onehalfspacing]{setspace}
\usepackage[thinc]{esdiff} % For nice derivative commands.
\usepackage{amssymb}
\usepackage{amsmath}
\usepackage{bm}
\usepackage{xcolor}
\usepackage{setspace}
\usepackage{geometry}
\usepackage{tikz}
\usepackage[ruled,vlined]{algorithm2e}
\usepackage{algorithmic}
\usepackage{booktabs} % For \toprule \midrule and \bottomrule commands
\hypersetup{pdfauthor={Name}}
\definecolor{green}{rgb}{0,0.6,0}
\newcommand{\norm}[1]{\left\lVert#1\right\rVert} % Typesetting norms.
\newcommand{\vect}[1]{\bm{#1}} % Typesetting vectors.

\newcommand{\greenline}{\raisebox{0.8pt}{\textcolor{green}{\textbf{---}}}}
\newcommand{\redline}{\raisebox{0.8pt}{\textcolor{red}{\textbf{---}}}}
\newcommand{\blueline}{\raisebox{0.8pt}{\textcolor{blue}{\textbf{---}}}}
\usepackage{siunitx}
\usepackage[switch]{lineno}

\usepackage{placeins}

\journal{Elsevier}
\begin{document}
\begin{frontmatter}
%\title{Deep neural network enabled corrective source term approach applied to 2D heat diffusion modelling} 
\title{Combining physics-based and data-driven techniques for reliable hybrid analysis and modeling using the corrective source term approach}

\author[sindresaddress1,sindresaddress2]{Sindre Stenen Blakseth}
\ead{sindre.blakseth@sintef.no}

\author[adilsaddress,trondsaddress2]{Adil Rasheed\corref{mycorrespondingauthor}}
\cortext[mycorrespondingauthor]{Adil Rasheed}
\ead{adil.rasheed@ntnu.no}

\author[trondsaddress1,trondsaddress2]{Trond Kvamsdal}
\ead{trond.kvamsdal@ntnu.no}

\author[omersaddress]{Omer San}
\ead{osan@okstate.edu }

\address[sindresaddress1]{Department of Physics, Norwegian University of Science and Technology}
\address[sindresaddress2]{Currently at the Department of Gas Technology, SINTEF Energy Research}
\address[adilsaddress]{Department of Engineering Cybernetics, Norwegian University of Science and Technology}
\address[trondsaddress1]{Department of Mathematical Sciences, Norwegian University of Science and Technology}
\address[trondsaddress2]{Department of Mathematics and Cybernetics, SINTEF Digital}
\address[omersaddress]{School of Mechanical and Aerospace Engineering, Oklahoma State University}

%\doublespacing
\begin{abstract}
Upcoming technologies like digital twins, autonomous, and artificial intelligent systems involving safety-critical applications require models which are accurate, interpretable, computationally efficient, and generalizable. Unfortunately, the two most commonly used modeling approaches, physics-based modeling (PBM) and data-driven modeling (DDM) fail to satisfy all these requirements. In the current work, we demonstrate how a hybrid approach combining the best of PBM and DDM can result in models which can outperform them both. We do so by combining partial differential equations based on first principles describing partially known physics with a black box DDM, in this case, a deep neural network model compensating for the unknown physics. First, we present a mathematical argument for why this approach should work and then apply the hybrid approach to model two dimensional heat diffusion problem with an unknown source term. The result demonstrates the method's superior performance in terms of accuracy, and generalizability. Additionally, it is shown how the DDM part can be interpreted within the hybrid framework to make the overall approach reliable.  
\end{abstract}

\end{frontmatter}
%\linenumbers
%\doublespacing

\begin{table*}[h]
	\centering
	\caption*{}
	\resizebox{1\textwidth}{!}{
    	\begin{tabular}{llll}
    		 &                \textbf{Abbreviations}            & \textbf{Symbols} & \\
    		\midrule
    		BC      & Boundary Condition                & $T$       & Temperature \\
    		CoSTA\hspace{1.0em}   & Corrective Source Term Approach   & $\vect{T}_{\mathrm{ref}}$ & Reference temperature field \\
    		DDM     & Data-Driven Model(ing)            & $\vect{T}_{\mathrm{p}}$ / $\vect{T}_{\mathrm{d}}$ / $\vect{T}_{\mathrm{h}}$ & Temperature field predicted by PBM / DDM / CoSTA \\
    		DNN     & Deep Neural Network               & $T_e$, $T_w$, $T_n$, $T_s$ & Boundary temperatures \\
    		DT      & Digital Twin                      & $\hat{\sigma}$ / $\vect{\hat{\sigma}}_{\textsc{nn}}$ & Reference / DNN-generated corrective source term \\
    		FC      & Fully Connected                   & $\vect{\hat{\sigma}}_P$ / $\vect{\hat{\sigma}}_k$ & Source term correcting error in modeling of $P$ / $k$ \\
    		FVM     & Finite Volume Method              & $\mathcal{N}_{\Omega}$ / $\mathcal{N}_{\partial\Omega}$ & Operators defining general PDE \\
    		HAM     & Hybrid Analysis and Modeling    & $f$ / $g$ & Right-hand-side functions defining general PDE \\
    		MSE     & Mean Squared Error                & $u$ & True solution of general PDE \\
    		MMS     & Method of Manufactured Solutions\hspace{4.0em}  & $k$ & Thermal conductivity \\
    		NN      & Neural Network                    & $P$ & Internal heat generation rate \\
    		PBM     & Physics-Based Model(ing)          & $c_V$ & Specific heat capacity at constant volume \\
    		PDE     & Partial Differential Equation     & $\rho$ & Density \\
    		PGNN    & Physics-Guided Neural Network     & $\alpha$ & General system parameter \\
    		PINN    & Physics-Informed Neural Network   & $x$ / $y$ / $t$ & Spatial and temporal coordinates \\
    		ROM     & Reduced-Order Model               & $E_{\mathrm{p}}$ / $E_{\mathrm{d}}$ / $E_{\mathrm{h}}$ & Relative $\ell_2$-errors of PBM / DDM / CoSTA \\
    		        &                                   & $\epsilon_P$ / $\epsilon_k$ & Error in modeling of $P$ / $k$ \\
    		        &                                   & $\xi$ & Convenience function (see Eq.~\eqref{eq:ksi}) \\
    		        &                                   & $\tilde{}$ \, / \, $\hat{\tilde{}}$ & Approximation / Corrected approximation \\
    		        &                                   & $_{j,i}$ \, / \, $^n$ & Grid cell indices / Time level \\
    		        &                                   & DNN$_{\sigma}$ / DNN$_T$\hspace{1.0em} & DNN predicting               $\hat{\sigma}$ / $T$ \\
    		\bottomrule
    	\end{tabular}}
	\label{tab:nomenclature}
\end{table*}

\section{Introduction}
\label{sec:introduction}
The development and industrial utilization of digital twins (DTs) is an important trend facilitated by the increased digitalization following Industry 4.0, in addition to recent advances within data processing, computational infrastructure and big data cybernetics. DTs \citep{rasheed2020dtv} can be defined as virtual representations of physical assets and their applications include real-time prediction, optimization, monitoring, control, and improved decision making. For DTs to be successful in these applications, high-quality modeling techniques are paramount. In particular, DTs require models which are generalizable, trustworthy, self-evolving and computationally efficient while maintaining good accuracy \citep{san2021hybrid}. Unfortunately, it has proven difficult to attain all four of these modelling characteristics using either of the two traditional modeling paradigms: physics-based modeling (PBM) and data-driven modeling (DDM).

\begin{figure}[h]
	\includegraphics[width=0.95\linewidth]{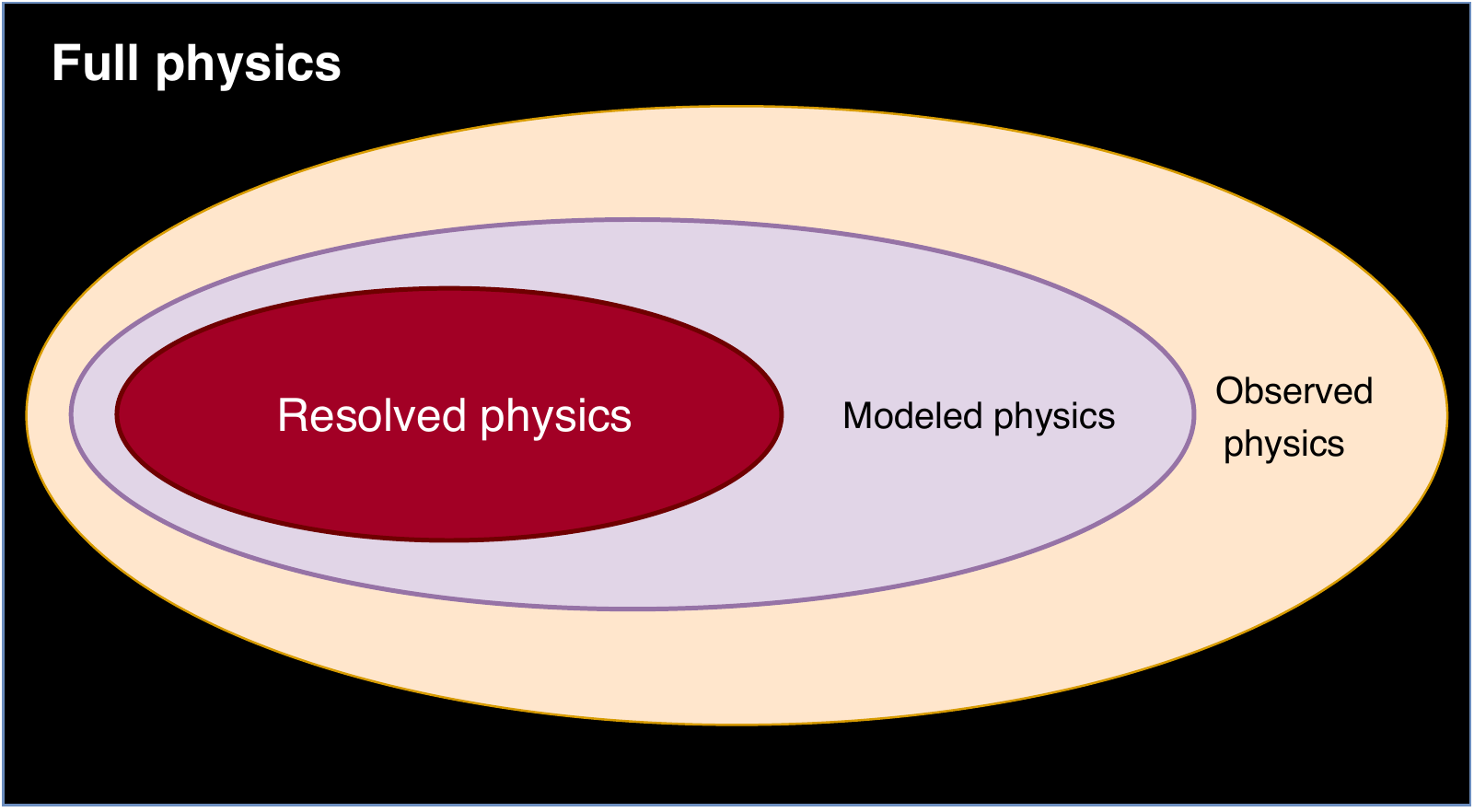}
	\caption{PBM: black part corresponds to unknown / unmodeled physics, orange ellipse corresponds to observed physics, purple ellipse corresponds to actually modelled physics while red ellipse signifies the actual physics solved for. 
	}
	\label{fig:pbm} 
\end{figure}
PBM (Figure \ref{fig:pbm}) generally describes the system to be modeled using a set of governing equations representing known and understood physical phenomena. However, the governing equations generally do not reflect the complete physics of the system, as some relevant physics may be unobserved, not understood, or neglected as a simplifying assumption. In addition, further loss of physics may result from solving the governing equations using numerical methods with finite precision. As such, PBM generally does not capture the complete physics of the system being modeled. However, they are still considered trustworthy because we know exactly which physical phenomena are included in the model and can bound their numerical errors. PBM also tends to generalize well, as it is typically not fine-tuned for specific applications. However, solving the governing equations can be computationally expensive. Another disadvantage of PBM is that it is generally static in the sense that it does not automatically get updated to account for new scenarios encountered after model deployment.

\begin{figure}[h]
	\includegraphics[ width=0.95\linewidth]{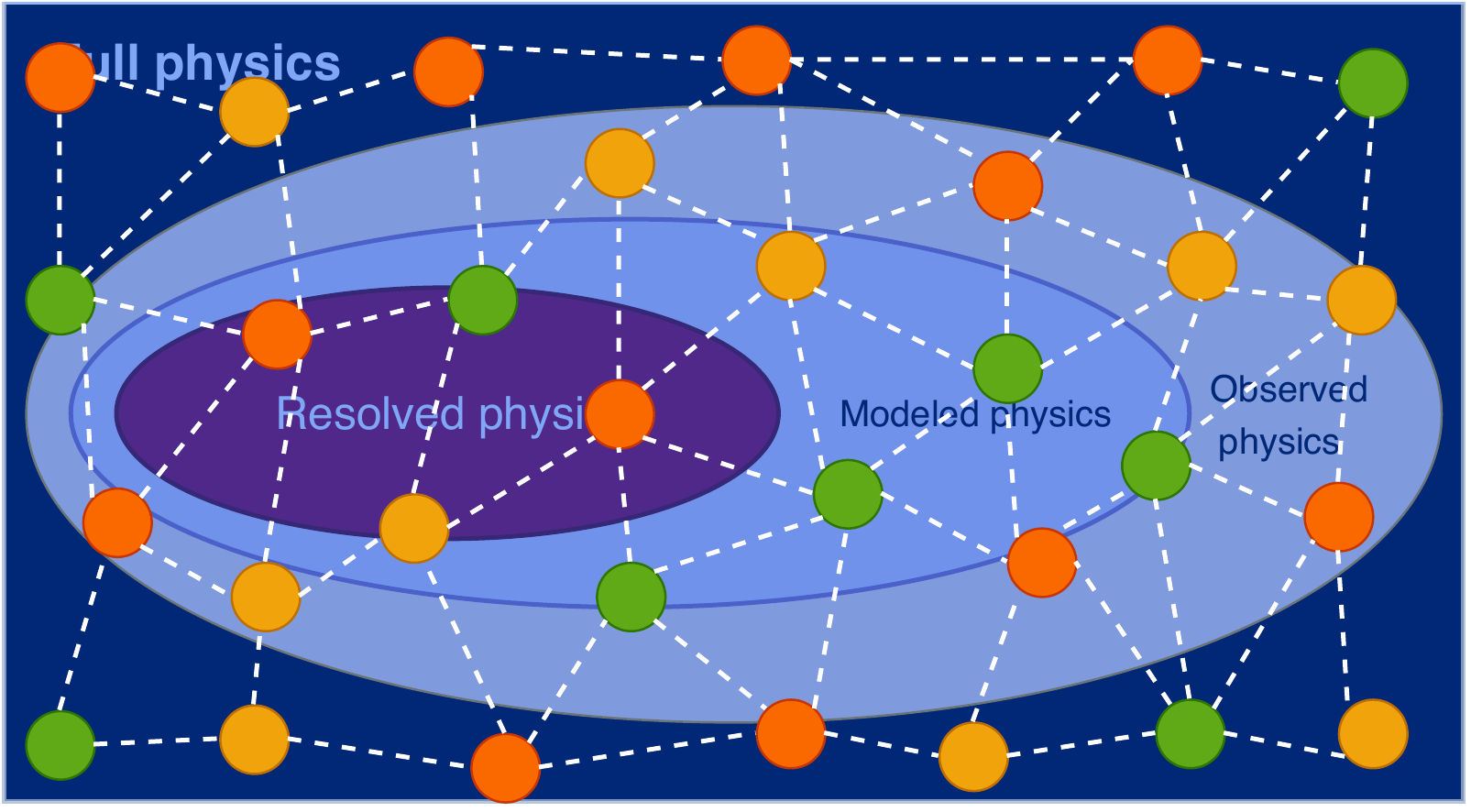}
	\caption{DDM: It is assumed that since data is a manifestation of both known and unknown physics, models trained on the data will implicitly captured full physics. 
	}
	\label{fig:ddm} 
\end{figure}
DDM (Figure \ref{fig:ddm}) exhibits the opposite traits of PBM, as DDM is generally inexpensive to run and can be continuously updated using new data even after deployment. Furthermore, the fact that observations may include neglected/unknown physics means that DDM, if tuned and trained perfectly, can reflect the complete physics. However, under realistic (imperfect) conditions, DDM will be biased towards the data samples on which the model is trained. This limits the models' generalizability, especially to extrapolation scenarios. Additionally, even with recent research focused on increasing the explainability of DDM, it can be challenging to establish precisely what physics are modeled by a DDM approach. This black-box-like nature greatly hurts the trustworthiness of DDM and is the primary barrier keeping DDM from entering high-stakes and safety-critical applications.

From an analysis of the strong and weak points of PBM and DDM as described above, it is clear that neither modeling approach is ideal for use in DTs. However, we see that all four modeling characteristics identified by \citet{san2021hybrid} can be attained by combining PBM and DDM in a way that retains their strengths while eliminating their weaknesses. This is exactly the philosophy behind the emerging Hybrid Analysis and Modeling (HAM) paradigm. As shown in Figure \ref{fig:ham}, HAM utilizes PBM to the maximum extent possible, and only compensates for the unmodelled / unknown physics using DDM. Recent works have explored many interesting approaches to HAM, most of which falls into one of the following categories: 
\begin{figure}[h]
	\includegraphics[width=0.95\linewidth]{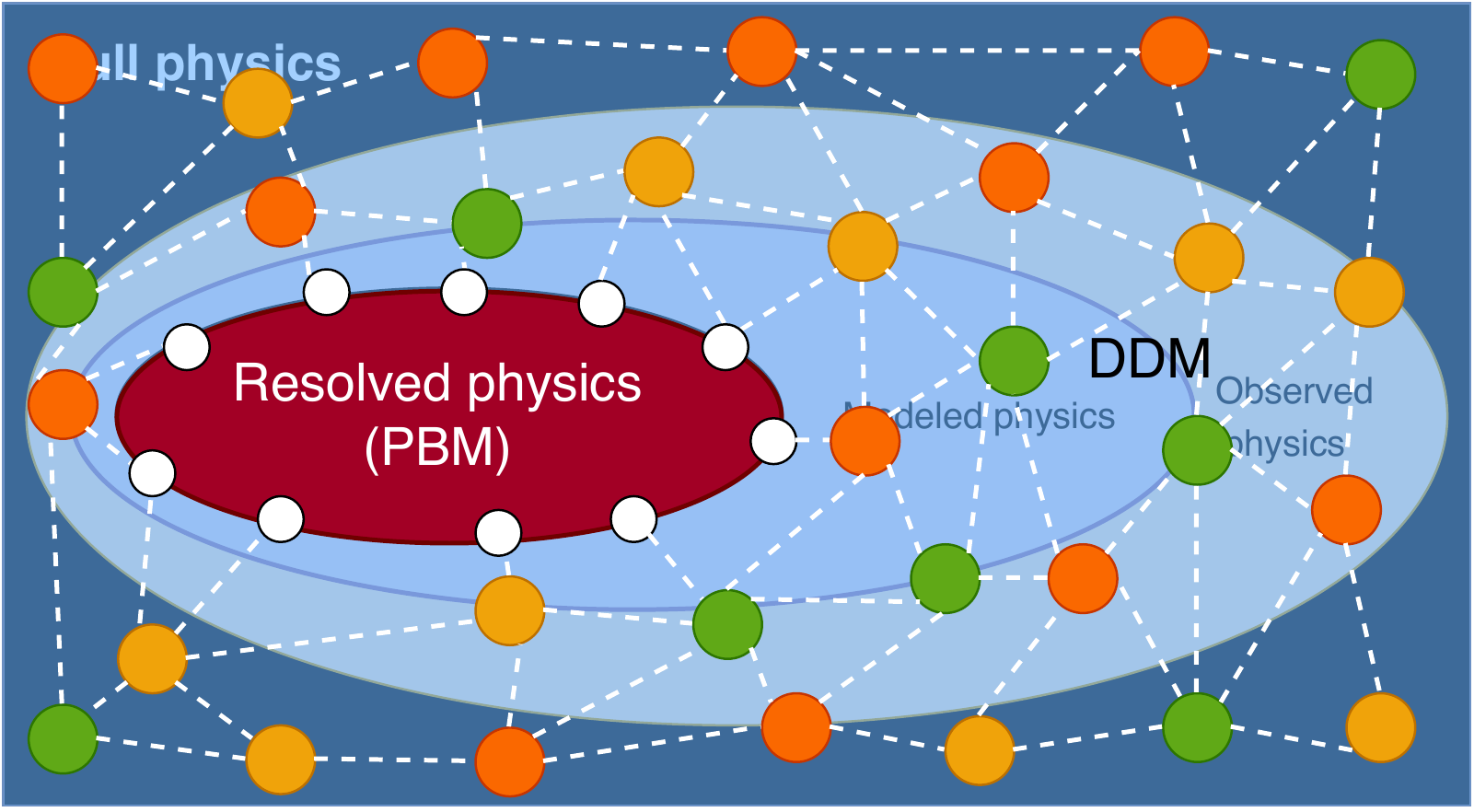}
	\caption{Hybrid analysis and modeling: It maximizes the utilization of the well known PBM while correcting for the unknown using DDM. In the CoSTA, PBM is described by partial differential equations and DDM is a DNN.
	}
	\label{fig:ham} 
\end{figure}
\begin{enumerate}
    \item PBM embedded inside neural networks (NNs): Examples of these are embedding a differentiable convex optimisation solver \citep{amos_2017_optnet} or a rigid body simulator \citep{avilabelbuteperes_2018_end} in a neural network. A common challenge with these methods is that they are computationally expensive not only for training but also for inference.
    \item Reduced Order Model (ROM): The ROM approach \citep{quarteroni2014reduced} involves projecting complex partial differential equations (PDEs) onto a reduced dimensional space based on the singular value decomposition of the offline high fidelity simulation snapshots resulting in a set of ordinary differential equations which are fast to solve. \citet{XIANG2022117641} applied ROM to model heat transfer in a battery pack of an electric vehicle, while \citet{GEORGAKA2020104615} applied it to model turbulent heat transfer problems, and \citet{LI2020118783} modelled steady-state and transient heat transfer in fractured geothermal reservoir. A recent review conducted by \citet{shady2021ocf} gives a detailed overview of the various ROM approaches. Despite the huge potential ROM holds, their development requires knowledge of the equation governing the process to be modelled. 
    \item Physics-informed neural networks (PINN): The work of \citet{raissi2019physics} involves penalizing the cost function of the neural network with the residual of governing equations representing physical laws. \citet{PENWARDEN2021110844} propose a particular multifidelity approach applied to PINNs that exploits low-rank structure. In the context of heat transfer, PINN has been used by \citet{HE2021102719} for solving both direct and inverse heat conduction problems. The penalization of the cost function can result in challenges during the optimization process due to the increased complexity of the cost function. Moreover, the exact form of the governing equations is a prerequisite for the method to work well.   
    \item Data-driven equation discovery: Sparse regression based on $l_1$ regularization \citep{BAKARJI2021110219, Champion22445} and symbolic regression based on gene expression programming have been shown to be very effective in complex equation discovery directly from data \citep{vaddireddy2020fes}. \citet{XU2021110592} 
    demonstrated deep-learning based discovery of partial differential equations in integral form from sparse and noisy data. However, the limitations of this approach are that either large number of additional features are required to be handcrafted (in case of sparse regression) based on prior knowledge or the resulting models are unstable and prone to overfitting (in case of symbolic regression based on gene expression programming). In the case of using deep learning, interpretability remains elusive.
    \item Physics-guided neural network (PGNN): This concept has been recently introduced to improve the training and predictions of deep neural networks (DNN). Partial knowledge, prior information, or results from highly simplified (and hence incomplete) models are injected into an intermediate hidden layer of the neural network \citep{haakon2022pgnn}. The injection helps in improving accuracy, reducing model uncertainty and enabling more robust training. The approach has been used to combine information from simplified analytical models or low-fidelity models with noisy data obtained either from experiments or high-fidelity simulations through a neural network. It has been shown that this multi-fidelity information fusion framework produces physically consistent models that achieve better generalizability than purely DDM \citep{pawar2021pgml, pawar2021msw}. However, since PGNN is just a special kind of NN, misbehaviour of the neural network in unseen conditions and lack of interpretability may still be an issue in high-stakes application.
\end{enumerate}

It is clear from the discussion so far that most of the HAM approaches mentioned above have some shortcomings. Some of the approaches, like PBM embedded inside NNs, tend to be computationally expensive. Others, such as ROM and PINN, require the exact form of the equations governing the physics to be modelled. Data-driven equation discovery based on symbolic regression can be unstable and not fit for interpretation, while PGNN also offers limited interpretability and few opportunities for NN sanity checks.
These observations have motivated our recent work \cite{blakseth2022dnn} on a different HAM approach -- the Corrective Source Term Approach (CoSTA) -- where a PBM is augmented with a data-driven component. More specifically, a DNN-generated corrective source term is added to the (discretized) governing equation(s) of the PBM such as to correct any errors present in the original PBM, as illustrated in Figure~\ref{fig:CoSTA_overview}. These errors may e.g.\ stem from partial knowledge, discretization, and/or inaccurate parameter estimation. An important difference between CoSTA and the other HAM approaches discussed above is that CoSTA utilizes PBM to the greatest extent possible. 

In \cite{blakseth2022dnn}, CoSTA was demonstrated to work for simple, one-dimensional heat transfer problems. In the current work we extend the work to two dimensions and take a closer look at the interpretability of the DNN-generated source term. The main contribution of this work can be enumerated as follows:

\begin{itemize}
    \item Provide a brief presentation of the approach's underlying mathematical foundation
    \item Apply the approach to model a wide variety of two-dimensional heat diffusion phenomena
    \item Demonstrate and discuss how the DNN-generated correction term can be interpreted in a physics context, thereby increasing the explainability and reliability of the approach.
\end{itemize}

In Section~\ref{sec:CoSTA_rationale}, we present a profound mathematical foundation of the approach. We then continue with a discussion on heat diffusion modeling and its importance in Section~\ref{sec:modelling}. Physics-based and data-driven heat diffusion models are presented in Sections~\ref{subsec:PBM} and~\ref{subsec:DDM}, respectively. In Section~\ref{subsec:HAM}, we explain in detail how to combine these models using the proposed hybrid approach, as illustrated in Figure~\ref{fig:CoSTA_overview}.

Section~\ref{sec:setup} is devoted to explaining the setup of our numerical experiments -- including the manufactured solutions considered, our DNN architecture and hyperparameter choices, and our data generation, training and testing procedures. Our experimental results are presented and discussed in Section~\ref{sec:resultsanddiscussion} before the article is concluded in Section~\ref{sec:conclusionandfuturework} with a brief summary and an outlook on future work.

\begin{figure}
	\centering 
	\begin{tikzpicture}[scale=0.2]
	    
	    \draw[color=white, fill=green!15, opacity=0.8] (-18, -5.5) rectangle (19, 5.5);
	    \draw[line width=0.1cm, color=green!80!black, fill=green!0, opacity=0.3] (-18, -5.5) rectangle (19, 5.5);
	    
	    \draw[line width=0.1cm, color=white,fill=white] (-15, -2.5) rectangle (4, 2.5);
	    \draw[line width=0.1cm, color=red!40,fill=red!15,opacity=0.6] (-15, -2.5) rectangle (4, 2.5);
	    
	    \draw[line width=0.1cm, color=white, fill=white] (8, -2.5) rectangle (16, 2.5);
	    \draw[line width=0.1cm, color=blue!40,fill=blue!15,opacity=0.6] (8, -2.5) rectangle (16, 2.5);
	    
	    \node[scale=0.7*2, opacity=0.8] at (-11, -4.25)  {\textcolor{green!40!black}{CoSTA}};
	    \node[scale=0.7*2] at (-11.8, 3.75)  {\textcolor{red!45}{PBM}};
	    
	    \node[scale=0.7*2] at (12, 3.75)  {\textcolor{blue!40}{DDM}};
	    
	    \node[scale=2] at (-11, 0.3) {$\widetilde{\mathcal{N}}$};
	    \node[scale=2] at (-7.5, 0.2) {$\hat{\tilde{u}}$};
	    \node[scale=2] at (-3, -0.4) {$=$};
	    \node[scale=2] at (1.5, -0) {$\tilde{f}$};
	    \node[scale=2] at (6, -0.1) {$+$};
	    \node[scale=2] at (12, -0.2) {$\hat{\sigma}_{\textsc{nn}}$};
        
    \end{tikzpicture}
	\caption{CoSTA combines PBM and DDM into a unified model by adding a DNN-generated corrective source term to the governing equation of the PBM.}
	\label{fig:CoSTA_overview}
\end{figure}

\section{Theory}\label{sec:CoSTA_rationale}
This section briefly presents the mathematical justification of the hybrid approach, henceforth called the Corrective Source Term Approach (CoSTA), originally introduced in \cite{blakseth2021ica, blakseth2022dnn} for modeling systems governed by linear partial differential equations (PDEs). \textcolor{red}{The presentation largely follows \cite{blakseth2021ica}.}

First, suppose we want to solve the following general problem, defined on a domain $\Omega$ with boundary $\partial \Omega$:
\begin{alignat}{3}
    \label{eq:exact_in}
    \mathcal{N}_{\Omega} u &= f\ &&\mbox{in}\ &&\Omega, \\
    \mathcal{N}_{\partial\Omega} u &= g\ &&\mbox{on}\ \partial&&\Omega.
    \label{eq:exact_BC}
\end{alignat}
Here, $u$ is the unknown of the problem, $\mathcal{N}_{\Omega}$ and $\mathcal{N}_{\partial\Omega}$ are linear operators\footnote{For $u$ to be uniquely defined, $\mathcal{N}_{\partial\Omega}$ must be the unity mapping along a portion of $\partial\Omega$ of length greater than zero.} acting on $u$, and $f$ and $g$ are some functions. With this formulation, we also capture scenarios where there are multiple governing equations. In such scenarios, $u$ is a vector, and $f$ and $g$ are vector-valued functions.

Assume now that we have a PBM designed to predict $u$, and let $\tilde{u}$ denote the PBM's prediction of the true solution $u$.

If $\tilde{u}\neq u$, there is some error in the PBM, and this error must stem from one or more of the following sources:
\begin{enumerate}
    \item The true function $f$ in Equation~\eqref{eq:exact_in} is unknown, so it is approximated by $\tilde{f}$.
    \item The true operator $\mathcal{N}_{\Omega}$ in Equation~\eqref{eq:exact_in} is unknown, so it is approximated by $\widetilde{\mathcal{N}}_{\Omega}$.
    \item The true function $g$ in Equation~\eqref{eq:exact_BC} is unknown, so it is approximated by $\tilde{g}$.
    \item The true operator $\mathcal{N}_{\partial\Omega}$ in Equation~\eqref{eq:exact_BC} is unknown, so it is approximated by $\widetilde{\mathcal{N}}_{\partial\Omega}$.
    \item A combination of the above.
    \item The true governing equation~\eqref{eq:exact_in} and the true boundary conditions~\eqref{eq:exact_BC} are known, but cannot be solved analytically. To obtain a prediction $\tilde{u}$, we must therefore solve some approximation of the true system constituted by Equations~\eqref{eq:exact_in} and~\eqref{eq:exact_BC}, which effectively puts us in one of the other cases. For example, one could approximate the true operator $\mathcal{N}_{\Omega}$ with some numerical operator $\mathcal{N}_{\mathrm{num}}$ e.g.\ based on finite-difference approximations, which is equivalent to Case 2.
\end{enumerate}

We observe that Cases 3 and 4 are analogous to Cases 1 and 2 because $\mathcal{N}_{\partial\Omega}$ and $g$ play exactly the same roles in Equation~\eqref{eq:exact_BC} as $\mathcal{N}_{\Omega}$ and $f$ do in Equation~\eqref{eq:exact_in}.

Since Case 6 is also mathematically equivalent to one of the other cases, it suffices to consider Cases 1 and 2, and combinations thereof.

Suppose now that the PBM-predicted solution $\tilde{u}$ is given as the solution of the following system:
\begin{alignat}{3}
    \label{eq:PBM_in}
    \widetilde{\mathcal{N}}_{\Omega} \tilde{u} &= \tilde{f}\ &&\mbox{in}\ &&\Omega, \\
    \mathcal{N}_{\partial\Omega} \tilde{u} &= g\ &&\mbox{on}\ \partial&&\Omega.
    \label{eq:PBM_BC}
\end{alignat}
This formulation encompasses both Case 1 (for $\widetilde{\mathcal{N}}_{\Omega} = {\mathcal{N}}_{\Omega}$ and $\tilde{f} \neq f$), Case 2 (for $\widetilde{\mathcal{N}}_{\Omega} \neq {\mathcal{N}}_{\Omega}$ and $\tilde{f} = f$), and combinations thereof (for $\widetilde{\mathcal{N}}_{\Omega} \neq {\mathcal{N}}_{\Omega}$ and $\tilde{f} \neq f$).
Furthermore, suppose we modify the system above by adding a source term $\hat{\sigma}$ to Equation~\eqref{eq:PBM_in}, and let the solution of the modified system be denoted $\hat{\tilde{u}}$. Then, the modified system reads
\begin{alignat}{3}
    \label{eq:PBM_mod_in}
    \widetilde{\mathcal{N}}_{\Omega} \hat{\tilde{u}} &= \tilde{f} + \hat{\sigma} \ &&\mbox{in}\ &&\Omega, \\
    \mathcal{N}_{\partial\Omega} \hat{\tilde{u}} &= g\ &&\mbox{on}\ \partial&&\Omega.
    \label{eq:PBM_mod_BC}
\end{alignat}
and the following theorem holds.

\paragraph{Theorem} Let $\hat{\tilde{u}}$ be a solution of Equations~\eqref{eq:PBM_mod_in} and~\eqref{eq:PBM_mod_BC}, and let $u$ be a solution of Equations~\eqref{eq:exact_in} and~\eqref{eq:exact_BC}. Then, for all operators $\widetilde{\mathcal{N}}_{\Omega}$, ${\mathcal{N}}_{\Omega}$, $\widetilde{\mathcal{N}}_{\partial\Omega}$ and $\mathcal{N}_{\partial\Omega}$ and all functions $f$, $\tilde{f}$, $g$ and $\tilde{g}$ such that $\hat{\tilde{u}}$ and $u$ are uniquely defined, there exists a function $\hat{\sigma}$ such that $\hat{\tilde{u}} = u$.

\textit{Proof}:
Define the residual $r$ of the PBM's governing equation~\eqref{eq:PBM_in} as\footnote{Note that our definition is in some sense opposite of common practice; we have defined the residual by inserting the true solution into the approximate equation rather than inserting the approximate solution into the true equation. The latter is the conventional approach, and is used e.g.\ in truncation error analysis \citep[chapter~8]{leveque2002fvm}. The reason for our choice is two-fold: 1) It yields the simplest proof of the theorem. 2) When observing a real-world system, it is often easier to measure its state than to find the exact governing equation describing said state. }
\begin{equation}
    r = \widetilde{\mathcal{N}}_{\Omega} u - \tilde{f}.
    \label{eq:residual}
\end{equation}
If we set $\hat{\sigma} = r$ in Equation~\eqref{eq:PBM_mod_in}, we then obtain
\begin{align*}
    \widetilde{\mathcal{N}}_{\Omega} \hat{\tilde{u}} &= \tilde{f} + \hat{\sigma} \\
    &= \tilde{f} + \widetilde{\mathcal{N}}_{\Omega} u - \tilde{f} \\
    &= \widetilde{\mathcal{N}}_{\Omega} u \\
    \implies \quad \hat{\tilde{u}} &= u \quad \quad \quad \quad \quad \quad \blacksquare
\end{align*}

The theorem above proves that, for any error in the PBM's governing equation~\eqref{eq:PBM_in}, there always exists a corrective source term $\hat{\sigma}$ which we can add to that equation such that the solution $\hat{\tilde{u}}$ of the modified governing equation~\eqref{eq:PBM_mod_in} is equal to the true solution $u$. Furthermore, any error in the PBMs boundary conditions (Equation~\eqref{eq:PBM_BC}) can be corrected analogously, since Equations~\eqref{eq:PBM_mod_in} and~\eqref{eq:PBM_BC} have the same functional form. Thus, the true solution $u$ of the true governing equations can always be retained by modifying an erroneous PBM with a corrective source term. This observation is the principal theoretical justification of CoSTA. It is worth pointing out that, so far, we have not made any assumption regarding the operators $\mathcal{N}_{\Omega}$ and $\mathcal{N}_{\partial\Omega}$ except that they are linear. Hence, the approach should be applicable across a wide array of physical problems that can be cast in the form of the above equations. The broad \emph{applicability} of the approach is not to be confused with its \emph{generalizability}, which is its ability to provide accurate predictions for previously unseen states of some particular system. The case study presented in the following is aimed at demonstrating the approach's generalizability.

\section{Heat Diffusion Modeling}
\label{sec:modelling}
To demonstrate the potential of CoSTA, we choose to study two dimensional heat diffusion problems. The main motivation for choosing such problems is two-fold. Firstly, temperature can give insight into a wide variety of physical phenomena.\footnote{Trivial examples include using temperature to evaluate the power output of a heater, or to indicate an impending malfunction due to overheating.} Secondly, cost-effective and non-intrusive measurement techniques (e.g.\ based on thermal cameras) exist to make high-resolution spatio-temporal temperature measurements. In real-world applications, such techniques can be used to obtain the reference data needed for training the DNN used by CoSTA.

In this following sections, we describe the PBM, DDM and CoSTA models used in our numerical experiments on 2D heat diffusion. These models are presented in Sections~\ref{subsec:PBM},~\ref{subsec:DDM} and~\ref{subsec:HAM}, respectively. But first, we shall briefly describe the heat equation in Section~\ref{subsec:heat_equation}.

\subsection{The Heat Equation}\label{subsec:heat_equation}

The heat equation, which describes heat conduction through solid materials, can be written as
\begin{equation}
    \int\limits_V \rho c_V \diffp{T}{t} \mathrm{d}V
    = \int\limits_{\partial V} \left(k\vect{\nabla} T\right) \cdot\vect{\hat{n}}\,\mathrm{d}A + \int\limits_V P\ \mathrm{d}V
    \label{eq:general_integral_form}
\end{equation}
for a stationary system with volume $V$, surface $\partial V$, surface unit normal $\vect{\hat{n}}$, density $\rho$, specific heat capacity at constant volume $c_V$, conductivity $k$, internal heat generation rate $P$ and temperature $T$.
We take Equation~\eqref{eq:general_integral_form} to be the true governing equation for all systems considered in the present work.
Comparing with the general formulation used in Section~\ref{sec:CoSTA_rationale}, Equation~\eqref{eq:general_integral_form} corresponds to Equation~\eqref{eq:exact_in} with

\begin{equation}
    u = T, \quad
    \mathcal{N}_{\Omega} u = \int\limits_V \rho c_V \diffp{T}{t} \mathrm{d}V
    - \int\limits_{\partial V} \left(k\vect{\nabla} T\right) \cdot\vect{\hat{n}}\,\mathrm{d}A
\end{equation}
and
\begin{equation}
    f = \int\limits_V P\ \mathrm{d}V.
\end{equation}

To have a complete formulation of the system at hand, we also need to formulate the boundary conditions (BCs) of the system. In this work, we only consider Dirichlet BCs, which means that the temperature at the domain boundary is specified by some function(s). For the 2D systems considered in this work, the Dirichlet BCs can be formulated as
\begin{equation}
\begin{split}
    T(x_e, y, t) = T_e(y, t), \quad
    T(x_w, y, t) = T_w(y, t), \\ 
    T(x, y_n, t) = T_n(x, t), \quad
    T(x, y_s, t) = T_s(x, t),
\end{split}
    \label{eq:Dirichlet_BCs}
\end{equation}
where the subscripts $_e$, $_w$, $_n$ and $_s$ denote quantities evaluated at, respectively, the eastern (right), western (left), northern (upper) and southern (bottom) domain boundaries, and $T_e$, $T_w$, $T_n$ and $T_s$ are the functions specifying the boundary temperature. Again comparing with Section~\ref{sec:CoSTA_rationale}, we see that Equation~\eqref{eq:Dirichlet_BCs} is equivalent to Equation~\eqref{eq:exact_BC} with $u=T$, $\mathcal{N}_{\partial \Omega}$ as the unity operator, and $g$ being equal to $T_e$, $T_w$, $T_n$ or $T_s$ depending on whether we are on the eastern, western, northern or southern part of $\partial \Omega$.

\subsection{Physics-Based Modeling}\label{subsec:PBM}

We now want to obtain a PBM for Equations~\eqref{eq:general_integral_form} and~\eqref{eq:Dirichlet_BCs}. By limiting ourselves to 2D systems and assuming $k$, $\rho$ and $c_V$ to be constant\footnote{In our numerical experiments, we consider scenarios where the assumption of constant $k$ does not hold. Assuming constant $k$ thereby allows us to synthesize modelling error in the PBM.}, we are able to rewrite Equation~\eqref{eq:general_integral_form} as
\begin{equation}
\begin{split}
    \int\limits_{y_{s}}^{y_{n}}\int\limits_{x_{w}}^{x_{e}} \diffp{T}{t} \mathrm{d}x\mathrm{d}y
    = \kappa \left( \left( \diffp{T}{x} \right)_{e}
    - \left( \diffp{T}{x} \right)_{w} + \left( \diffp{T}{y} \right)_{n}
    - \left( \diffp{T}{y} \right)_{s} \right) \\
    + \int\limits_{y_{s}}^{y_{n}}\int\limits_{x_{w}}^{x_{e}} \sigma\, \mathrm{d}x\mathrm{d}y,
\end{split}
    \label{eq:2D_FVM_starting_point}
\end{equation}
where $\kappa = k/(\rho c_V)$ and $\sigma = P/(\rho c_V)$.
Equation~\eqref{eq:2D_FVM_starting_point} can be solved numerically using the Implicit Euler FVM, which can be expressed on the following matrix form for two successive time levels $n$ and $n+1$:
\begin{equation}
    \mathbb{A}\vect{T}_{\mathrm{p}}^{n+1} = \vect{b}\left(\vect{T}_{\mathrm{p}}^n\right).
    \label{eq:2D_matrix_form}
\end{equation}
For a domain that is discretized with $N_j$ grid cells in the $x$-direction and $N_i$ grid cells in the $y$-direction, $\mathbb{A}$ is a banded $(N_jN_i\times N_jN_i)$-matrix with five non-zero diagonals, while $\vect{T}_{\mathrm{p}}$ and $\vect{b}$ are $N_jN_i$-dimensional vectors. The components of $\vect{T}_{\mathrm{p}}$ describe the temperature at the grid cell centers, as predicted by the PBM. The components are ordered such that the first $N_j$ components describe the temperature at bottom-most row of cell centers (from left to right), the subsequent $N_j$ components correspond to the second row from the bottom (still from left to right), and so on. Precise definitions of $\mathbb{A}$ and $\vect{b}$ can be found in~\citep{blakseth2021ica}. Here, we highlight that $\mathbb{A}$ depends on the conductivity $k$, while $\vect{b}$ depends on the heat generation rate $P$ and the system's boundary conditions (cf. Equation~\eqref{eq:Dirichlet_BCs}) in addition to the predicted temperature distribution at the old time level $n$, $\vect{T}_{\mathrm{p}}^n$. 
Comparing with Section~\ref{sec:CoSTA_rationale}, we see that Equation~\eqref{eq:2D_matrix_form} is equivalent to Equation~\eqref{eq:PBM_in} with
\begin{equation}
    \tilde{u} \leftrightarrow \vect{T}_{\mathrm{p}}^{n+1}, \quad
    \tilde{\mathcal{N}}_{\Omega} \tilde{u} \leftrightarrow \mathbb{A} \vect{T}^{n+1},
    \quad \mathrm{and} \quad
    \tilde{f} \leftrightarrow \vect{b}.
\end{equation}

In our numerical experiments, we use the LAPACK routine (accessed through the SciPy library) to solve the system~\eqref{eq:2D_matrix_form}. However, using a specialized solver for sparse, banded systems is advised for problems that are more computationally demanding than those considered herein.

\subsection{Data-Driven Modeling}\label{subsec:DDM}

The crux of DDM is to learn physics directly from observational data. For transient systems, this can be done by training a DNN to learn a mapping between two subsequent observations of the system state. For the heat diffusion problems considered herein, we take an observed state to be a vector $\vect{T}_{\mathrm{ref}}^n$ describing the true temperature at the center of the grid cells used to define the PBM, as described in Section~\ref{subsec:PBM}. The mapping we want the DNN to learn is then given by
\begin{align}
    \label{eq:ideal_DDM}
    \mathrm{DNN}_T : \mathbb{R}^{(N_j+2)(N_i+2)} &\rightarrow \mathbb{R}^{N_jN_i} \ \ \mathrm{such\ that} \ \ \vect{T}_{\mathrm{d}}^{n+1} = \vect{T}_{\mathrm{ref}}^{n+1},\\
    \vect{T}_{\mathrm{d}}^{n} &\mapsto \vect{T}_{\mathrm{d}}^{n+1} \nonumber
\end{align}
where $\vect{T}_{\mathrm{d}}^{n+1}$ refers to the temperature profile predicted by the DDM at time level $n+1$. The dimensionality of the DNN output is lower than that of the DNN input because the input vector contains boundary temperatures while the output vector does not. Since we consider Dirichlet BCs where the boundary temperatures are known, we need not have the DNN predict the boundary temperatures. However, it is still potentially useful to include them as DNN input with the aim of making the DNN's learning task easier. In an effort to reduce notational complexity, we use the same notation to denote both vectors with and without boundary information. Furthermore, we will use the notation DNN$_T$ to refer to both the mapping defined by Equation~\eqref{eq:ideal_DDM} and any DNN trained to approximate that mapping.

Our reason for choosing DNN-based DDM over other applicable DDMs is that DNNs have the ability to approximate any nonlinear mapping, as guaranteed by the universal approximation theorem. Notice also that if we know the true initial condition of a system, i.e., if we can set $\vect{T}_{\mathrm{d}}^{0} = \vect{T}_{\mathrm{ref}}^{0}$, then we will have $\vect{T}_{\mathrm{d}}^{n} = \vect{T}_{\mathrm{ref}}^{n}\ \forall n \geq 0$ if the mapping~\eqref{eq:ideal_DDM} is learnt perfectly.\footnote{It should be noted that, depending on the discretization used, Equation~\eqref{eq:ideal_DDM} may not constitute a well-defined mapping. Particular care should be taken for systems governed by so-called hyperbolic PDEs which permit discontinuous solutions. However, Equation~\eqref{eq:ideal_DDM} is a well-defined mapping for the cases considered in our numerical experiments.}

To summarize, the DDM used in this work is a DNN denoted DNN$_T$ which is trained to predict $\vect{T}_{\mathrm{ref}}^{n+1}$ given $\vect{T}_{\mathrm{ref}}^n$ for any time level $n$. During testing, the output of DNN$_T$ at time level $n$ is used as its input at time level $n+1$, i.e.\ $\vect{T}_{\mathrm{d}}^{n+1} = \mathrm{DNN}_T ( \vect{T}_{\mathrm{d}}^n )$. Since we use $\vect{T}_{\mathrm{d}}^{0} = \vect{T}_{\mathrm{ref}}^{0}$, we will have $\vect{T}_{\mathrm{d}}^{n} = \vect{T}_{\mathrm{ref}}^{n}\ \forall n \geq 0$ if DNN$_T$ is perfectly trained.

\subsection{Hybrid Analysis and Modeling with CoSTA}\label{subsec:HAM}
%\begin{figure}[tb]
%	\includegraphics[width=\linewidth]{Figures/ham.pdf}
%	\caption{CoSTA: The region within the red ellipse denotes known knowledge. The region outside denotes known unknown, unknown unknown, modelling error due to assumptions, numerical error. The basic idea behind CoSTA is to maximize the utilization of the first principle PBM and then compensate for the rest using a DDM.   
%	}
%	\label{fig:ham} 
%\end{figure}
In this section, we will briefly explain how we use CoSTA to model 2D heat diffusion. The PBM on which we base our CoSTA model is the Implicit Euler FVM described in Section~\ref{subsec:PBM}. In Section~\ref{sec:CoSTA_rationale}, we established that the ideal corrective source term is given generally as the residual defined in Equation~\eqref{eq:residual}. For the Implicit Euler FVM, we recall that $\tilde{\mathcal{N}}_{\Omega} \leftrightarrow \mathbb{A}$ and $f \leftrightarrow \vect{b}$. Moreover, we have $u \leftrightarrow \vect{T}_{\mathrm{ref}}^{n+1}$, such that the ideal corrective source term for the Implicit Euler FVM reads
\begin{equation}
    \vect{\hat{\sigma}}^{n+1}= \mathbb{A} \vect{T}_{\mathrm{ref}}^{n+1} - \vect{b} \left( \vect{T}_{\mathrm{ref}}^n \right).
    \label{eq:FVM_corr_src_def}
\end{equation}

As in Section~\ref{sec:CoSTA_rationale}, we use the corrective source term to define a modified governing equation whose solution is exactly equal to the reference solution at all grid nodes and at all time levels. We use a subscript $_{\mathrm{h}}$ (for ``hybrid analysis and modelling'') to denote the solution of the modified system, and we write the modified system as
\begin{equation}
    \mathbb{A}\vect{T}_{\mathrm{h}}^{n+1} = \vect{b} \left( \vect{T}_{\mathrm{h}}^n \right) + \vect{\hat{\sigma}}^{n+1}.
    \label{eq:FVM_mod}
\end{equation}

For a posteriori analyses, these equations can be used directly. However, for a priori predictions, $\vect{T}_{\mathrm{ref}}^{n+1}$ is unknown. We therefore use a DNN-generated corrective source term $\vect{\hat{\sigma}}_ {\textsc{nn}}^{n+1}$ to approximate the true corrective source term $\vect{\hat{\sigma}}^{n+1}$. As input to the DNN generating $\vect{\hat{\sigma}}_{\textsc{nn}}^{n+1}$, we use a predictor $\vect{\tilde{T}}_{\mathrm{h}}^{n+1}$ defined by
\begin{equation}
    \mathbb{A}\vect{\tilde{T}}_{\mathrm{h}}^{n+1} = \vect{b}\left( \vect{T}_{\mathrm{h}}^n \right),
\end{equation}
with $\mathbb{A}$ and $\vect{b}$ defined as in Equation~\eqref{eq:2D_FVM_starting_point}.
Our choice of DNN input was inspired by predictor-corrector schemes used for numerical integration. We make no claim that this choice is optimal, but observe that it has worked well in our numerical experiments. With this choice of input, we want to train the DNN of the CoSTA model to approximate the following mapping:
\begin{align}
    \label{eq:ideal_hybrid_2D}
    \mathrm{DNN}_{\sigma} : \mathbb{R}^{(N_j+2)\cdot(N_i+2)} &\rightarrow \mathbb{R}^{N_j\cdot N_i} \quad \mathrm{such\ that} \quad \vect{\hat{\sigma}}_{\textsc{nn}}^{n+1} = \vect{\hat{\sigma}}^{n+1}.\\
    \widetilde{\vect{T}}_{\mathrm{h}}^{n+1} &\mapsto \vect{\hat{\sigma}}_{\textsc{nn}}^{n+1} \nonumber
\end{align}
As for the DDM mapping~\eqref{eq:ideal_DDM}, the dimensionality reduction originates from the use of Dirichlet BCs. Furthermore, we use the notation DNN$_{\sigma}$ to refer to both the mapping~\eqref{eq:ideal_hybrid_2D} and any DNN trained to approximate that mapping.

\section{Experimental setup and procedures}
\label{sec:setup}

\subsection{Data Generation}
\label{subsec:manufacturedsolution}

%Although it would have been best to apply and evaluate the method on real measurement data, we chose to use synthetic data in the current work. This is primarily due to two reasons. Firstly, it would be difficult to change the simulated source terms in an experimental setup. Knowing the analytical solution of the equations helps in conducting a more rigorous error analysis presented in the results section. This will help in objectively evaluating the value of the proposed approach.
We use the method of manufactured solutions (MMS) to generate data for our numerical experiments.
Our motivation for using synthetic data generated using MMS instead of real data, is to make the analysis of CoSTA's accuracy and interpretability as rigorous as possible. Real data inevitably contains some noise, which would make it difficult to differentiate the models' accuracy in scenarios where several models perform well. Moreover, when the reference data is noisy, the true corrective source term $\hat{\sigma}$ (as defined by Equation~\eqref{eq:PBM_mod_in}) is not known precisely. This would make it difficult to assess our hypothesis that the DNN-generated corrective source term $\hat{\sigma}_{\textsc{nn}}$ can be interpreted to obtain useful information. In the present work, we feel it was important to keep the number of error sources to a minimum, such as to keep the analysis as straight-forward as possible.

The main concept of MMS is to prescribe some convenient but otherwise arbitrary function as the solution of the governing equation (the heat equation~\eqref{eq:general_integral_form} in our case). All of the parameters of the equation except for one (typically the heat generation rate $P$ for the heat equation) are also prescribed. The final parameter is then calculated by inserting the prescribed solution and parameters into the governing equation. This way, it is easy to obtain analytical solutions for any governing equation. Thus, MMS is a powerful tool for generating synthetic data for numerical experiments without the use of any expensive high-fidelity solvers.

Due to the integrals, Equation~\eqref{eq:general_integral_form} is not convenient for use with MMS. Therefore we use instead the so-called differential form of the heat equation when generating data for our numerical experiments. For smooth temperature profiles, the differential form is equivalent to Equation~\eqref{eq:general_integral_form}, and it reads
\begin{equation}
    \rho c_V \diffp{T}{t} = \diffp{}{x} \left( k \diffp{T}{x} \right) + \diffp{}{y} \left( k \diffp{T}{y} \right) + P.
\end{equation}
Our data generation procedure is then to prescribe $T$, $k$, $\rho$ and $c_V$ in the equation above and calculate the $P$ required for the equation to be satisfied. For simplicity we always prescribe $\rho$ and $c_V$ to unity in this work.

We consider a total of four different manufactured solutions $T_{\mathrm{ref}}$ as listed in Table~\ref{tab:manufactured_solutions_2D}. The corresponding chosen $k$ and calculated $P$ are also included in the table. All the manufactured solutions are parametrized by a parameter $\alpha$ which allows us to generate several time series using the same manufactured solutions. We consider a total of 22 different $\alpha$-values, such that we get 22 unique time series for each manufactured solutions. We emphasize that this way of choosing uniformly spaced values of $\alpha$ is not optimal. If we were to use resource-intensive experimentation for date generation then  a better approach would have been to use concepts from Design of Experiments to minimize the number of experiments while still generating informative data.

As shown in Table~\ref{tab:alphas}, 16 of these were used for DNN training, 2 were used for DNN validation, and 4 were used for model testing. Each time series was discretized using 5001 time levels on the temporal domain $[\SI{0}{\second}, \SI{5}{\second}]$ and 20$\times$20 grid cells on the spatial domain $[\SI{0}{\meter}, \SI{1}{\meter}]\times[\SI{0}{\meter}, \SI{1}{\meter}]$. All models operate on flattened data, meaning that any discretized 2D temperature field is represented by a 1D vector in the models.
\begin{table*}[h]
	\centering
	\caption{Manufactured solutions $T_{\mathrm{ref}}$ used for our experiments Each solution is taken to be defined on the spatial domain $[\SI{0}{\meter}, \SI{1}{\meter}]\times[\SI{0}{\meter}, \SI{1}{\meter}]$ and the temporal domain $[\SI{0}{\second}, \SI{5}{\second}]$. $P$ and $k$ are given in their respective SI units, while $T$ is given in degrees Celsius.}
	\resizebox{\textwidth}{!}{
    	\begin{tabular}{llll}
    		\toprule
    		Label & $T_{\mathrm{ref}}(x,y,t;\alpha)$ & $P(x,y,t;\alpha)$ & $k(x,y,t;\alpha)$\\
    		\midrule
    		2P1  & 
    		$t + 0.5\alpha(x^2 + y^2) + x$ & 
    		$(1 - 2\alpha)$  &
    		1
    		\\
    		2P2  &
    		$1 + \sin{(2\pi t + \alpha)}\cos{(2\pi x)}\cos{(2\pi y)}$  &
    		$2\pi \cos{(2\pi x)}\cos{(2\pi y)} \left( \cos{(2\pi t + \alpha)} + 4\pi \sin{(2\pi t + \alpha)} \right)$ & 
    		1 
    		\\
    		\midrule
    		2k1 &
    		$t+\alpha x + y^2$ &
    		$-(1+\alpha+2x+4y)$ &
    		$1+x+y$
    		\\
    		2k2 &
    		$\alpha + (t+1)\cos{(2\pi x)}\cos{(4\pi y)}$ &
    		$\cos{(2\pi x)}\cos{(4\pi y)}\left( 1 + 40\pi^2(t+1)\left( 1 + \sin{(1\pi x)}\sin{(4\pi y)} \right)\right)$ &
    		$2+\sin{(2\pi x)}\sin{(4\pi y)}$
    		\\
    		\bottomrule
    	\end{tabular}
	}
	\label{tab:manufactured_solutions_2D}
\end{table*}

\begin{table}[h]
	\centering
	\caption{Parametrization: Selection of $\alpha$-values corresponding to the training, validation and testing time series used in our experiments. Note that in the test set $\mathcal{A}_{\mathrm{test}}$, two values of $\alpha=-0.5$ and $2.5$ correspond to extrapolation scenarios while $\alpha=0.7$ and $1.5$ correspond to interpolation scenarios.}
	\begin{tabular}{lll}
		\toprule
		Purpose & Set of $\alpha$-values & Symbol     \\
		\midrule
		Training & $\{0.1, 0.2, \dots, 2.0\}\backslash\{0.7, 0.8, 1.1, 1.5\}$ & $\mathcal{A}_{\mathrm{train}}$ \\
		Validation & \{0.8, 1.1\} & $\mathcal{A}_{\mathrm{val}}$ \\
		Testing  & $\{-0.5, 0.7, 1.5, 2.5\}$ & $\mathcal{A}_{\mathrm{test}}$ \\
		\bottomrule
	\end{tabular}
	\label{tab:alphas}
\end{table}
\subsection{DNN Setup and Training Routines}
\begin{figure}[h]
	\includegraphics[trim=150 570 66 150,clip, width=\linewidth]{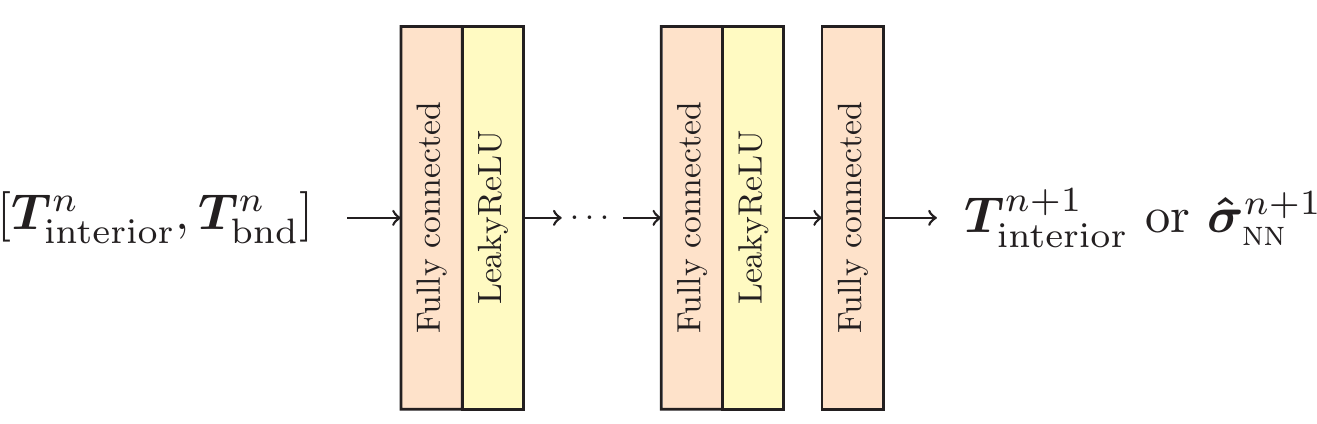}
	\caption[]{The fully connected DNN architecture used in the numerical experiments of the present work. Temperatures at both the domain interior and the boundary are given as model input. The data is processed by a number of fully connected layers with LeakyReLU activation functions. A fully connected layer without activation functions is used to generate the final output. For DDM, the output is the temperature in the domain interior\footnotemark at the subsequent time level. For CoSTA, the output is the corrective source term $\vect{\hat{\sigma}}_{\textsc{nn}}^{n+1}$. Adapted from \cite{blakseth2022dnn}.}
	\label{fig:arch} 
\end{figure}

\footnotetext{We assume the BCs are known, so there is no need for the DNN to predict boundary temperatures.}

\begin{table}[h]
	\centering
	\caption{The DNN hyperparameters used in our experiments.}
	\begin{tabular}{ll}
		\toprule
		Parameter               & Value     \\
		\midrule
		Loss function           & MSE       \\
		Learning rate           & 1e-5      \\
		Optimizer               & Adam      \\
		Batch size              & 32        \\
		\# hidden FC layers     & 4         \\
		Hidden FC layer width   & 80        \\
		LeakyReLU slope         & 0.01      \\
		Validation period       & 1e2       \\
		Overfit limit           & 20        \\
		\bottomrule
	\end{tabular}
	\label{tab:hyperparameters}
\end{table}
The fully-connected DNN architecture we use is illustrated in Figure~\ref{fig:arch}, and our hyperparameter choices are listed in Table~\ref{tab:hyperparameters}. To be compatible with our chosen spatial discretization, the DNN input and output layers must consist of 484 and 400 nodes respectively.\footnote{For our chosen discretization, the number of grid cells is 400, and the number of boundary nodes is 84.} The training procedures for the DNNs of DDM and CoSTA are illustrated in Figure~\ref{subfig:training} for a single data example $(\vect{T}_{\mathrm{ref}}^n, \vect{T}_{\mathrm{ref}}^{n+1})$. 

\subsection{Testing}

In each of our numerical experiments, we consider one of the manufactured solutions listed in Table~\ref{tab:manufactured_solutions_2D}, and attempt to replicate the four time series corresponding to $\alpha \in \mathcal{A}_{\mathrm{test}}$. For each time series, we inform the models of the true initial condition and the true boundary conditions.

To synthesize modeling error in the PBM, we set $P=0$ in the PBM when modelling Systems~2P1 and~2P2. However, for Systems~2k1 and~2k2, we inform the PBM of the true $P$. In these cases, modeling error is instead synthesized by the assumption of constant $k$. More specifically, we set $k=1$ in PBM for all the numerical experiments considered herein. Our hypothesis is that the corrective source term in the CoSTA model will correct for the modeling error synthesized in the PBM, irrespective of the whether the error stems from an incorrect $P$ or an incorrect $k$. We highlight that no modeling error is synthesized in the DDM model.

\begin{figure*}[h]
	\begin{subfigure}{\linewidth}
		\centering 
		\includegraphics[trim={1.7cm 15.5cm 2.5cm 11cm},clip,width=\textwidth]{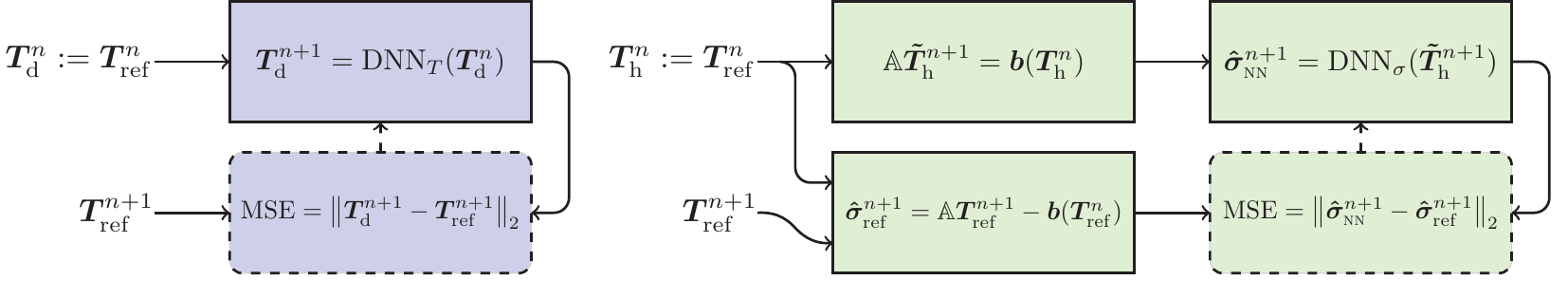}
		\caption{Training procedures for the DNNs used in CoSTA/DDM (left) and HAM (right). Note that, for all $n$, we use the reference profile as input during training. As such, the models are only trained to make \emph{local} (i.e.\ single-step) corrections. Moreover, they are not trained to recognize errors in their own predictions at previous time levels.}
		\label{subfig:training}
	\end{subfigure}%
	\\
	\begin{subfigure}{\linewidth}
	\vspace{0.5em}
		\centering 
		\includegraphics[trim={1.7cm 20.0cm 2.5cm 4.8cm},clip,width=\textwidth]{Figures/flow.pdf}
		\caption{Time stepping procedures for PBM (top), CoSTA/HAM (middle) and DDM (bottom). For all models, the output at one time level will be used as input at the subsequent time level.}
		\label{subfig:testing}
	\end{subfigure}%
	\caption{Training and time stepping procedures for the three modeling approaches PBM (red), DDM (blue) and HAM (green). Note that PBM is not included in (a) because it does not require any training. Figure adapted from \cite{blakseth2022dnn}.}
	\label{fig:traintest} 
\end{figure*}

For assessing the quality of the three models' predictions, we use the relative $\ell_2$-norms
\begin{equation}
\begin{split}
    & E_\mathrm{p} = \frac{\norm{\vect{T}_{\mathrm{p}}^n - \vect{T}_{\mathrm{ref}}^n}_2}{\norm{\vect{T}_{\mathrm{ref}}^n}_2}, \quad
      E_\mathrm{d} = \frac{\norm{\vect{T}_{\mathrm{d}}^n - \vect{T}_{\mathrm{ref}}^n}_2}{\norm{\vect{T}_{\mathrm{ref}}^n}_2}, \\
    & E_\mathrm{h} = \frac{\norm{\vect{T}_{\mathrm{h}}^n - \vect{T}_{\mathrm{ref}}^n}_2}{\norm{\vect{T}_{\mathrm{ref}}^n}_2}, \quad
\end{split}
    \label{eq:error_norms}
\end{equation}
where
\begin{equation}
    \norm{\vect{v}}_2 = \left( \sum\limits_{i=1}^D v_i^2 \right)^{1/2}
\end{equation}
for any $D$-dimensional vector $\vect{v}$. The training and testing procedures are illustrated in Figure \ref{fig:traintest}. Algorithmic representations of the training and testing are given in Algorithms~1 and~2 of \cite{blakseth2022dnn}.\footnote{Keep in mind that, while the procedures used here and in \cite{blakseth2022dnn} are completely analogous, there are some technical difference. An obvious example is the definitions of the temperature vectors. Moreover, \cite{blakseth2022dnn} use $\hat{q}$ to refer to the heat generation rate $P$ which we here denote $P$.}

\section{Results and discussion}
\label{sec:resultsanddiscussion}

In this section, we present and discuss the results of our four numerical experiments. The experiments concerning solutions 2P1 and 2P2, where modelling error due to an unknown $P$ is the dominant error source in the PBM, are considered in Section~\ref{subsec:exp_2D_q}. The experiments concerning solutions 2k1 and 2k2, where an incorrectly modelled $k$ is the primary PBM error source, are considered thereafter in Section~\ref{subsec:exp_2D_k}. Finally, the section is concluded by a discussion on the interpretability of the corrective source term in Section~\ref{subsec:interpretation}.

In Sections~\ref{subsec:exp_2D_q} and~\ref{subsec:exp_2D_k}, the results are grouped in interpolation scenarios $\alpha \in \{0.7, 1.5\}$ and extrapolation scenarios $\alpha \in \{-0.5, 2.5\}$. 
We make a distinction between the interpolation and extrapolation scenarios during testing because data-driven models tend to do relatively better in interpolation compared to the extrapolation scenarios. This is due to the fact that the test data corresponding to interpolations are better represented by the training data. Since by design, PBMs do not differentiate between the two scenarios, it is expected that a hybrid approach will inherit this strength of the PBM, and consequently perform better than pure DDM in the extrapolation scenarios.

Both result sections begin with a discussion on the interpolation scenarios, while the extrapolation scenarios are considered thereafter. For each $\alpha$-value and each manufactured solution, we display the temporal development of the $\ell_2$-errors $E_{\mathrm{p}}$, $E_{\mathrm{d}}$ and $E_{\mathrm{h}}$ defined in Equation~\eqref{eq:error_norms} (cf. Figures~\ref{fig:2P_interp_errors},~\ref{fig:2P_extrap_errors},\ref{fig:2k_interp_errors} and~\ref{fig:2k_extrap_errors}). Additionally, we also display the relative error fields
$(\vect{T}_{\mathrm{p}}^{N_t-1} - \vect{T}_{\mathrm{ref}}^{N_t-1}) / \vect{T}_{\mathrm{ref}}^{N_t-1}$, $(\vect{T}_{\mathrm{d}}^{N_t-1} - \vect{T}_{\mathrm{ref}}^{N_t-1}) / \vect{T}_{\mathrm{ref}}^{N_t-1}$, and $(\vect{T}_{\mathrm{h}}^{N_t-1} - \vect{T}_{\mathrm{ref}}^{N_t-1}) / \vect{T}_{\mathrm{ref}}^{N_t-1}$, where all subtractions and divisions are applied component-wise.\footnote{For these illustrations, we use the imshow function of Matplotlib, which interpolates the discrete differences to produce smooth error fields.}

\subsection{Experiments with Unknown Source Term}\label{subsec:exp_2D_q}

In this section, we consider two experiments where the source term $P$ of the heat equation is assumed unknown. From a physical point of view, this can be interpreted as some unknown heating within the system. For example, $P$ can correspond to an unknown power output of a heater in a room, to heat generated from electrical resistance in a system influenced by electrical currents, or to heat generated by friction inside a system with moving components. The manufactured solutions studied in this section are Solutions~2P1 and~2P2 (cf. Table~\ref{tab:manufactured_solutions_2D}).
We discuss the results for the interpolation scenarios ($\alpha \in \{0.7, 1.5\}$) first and the extrapolation scenarios ($\alpha \in \{-0.5, 2.5\}$) thereafter.

\begin{figure}[b!]
	\begin{subfigure}[b]{0.5\linewidth}
		\centering 
		\includegraphics[width=\textwidth]{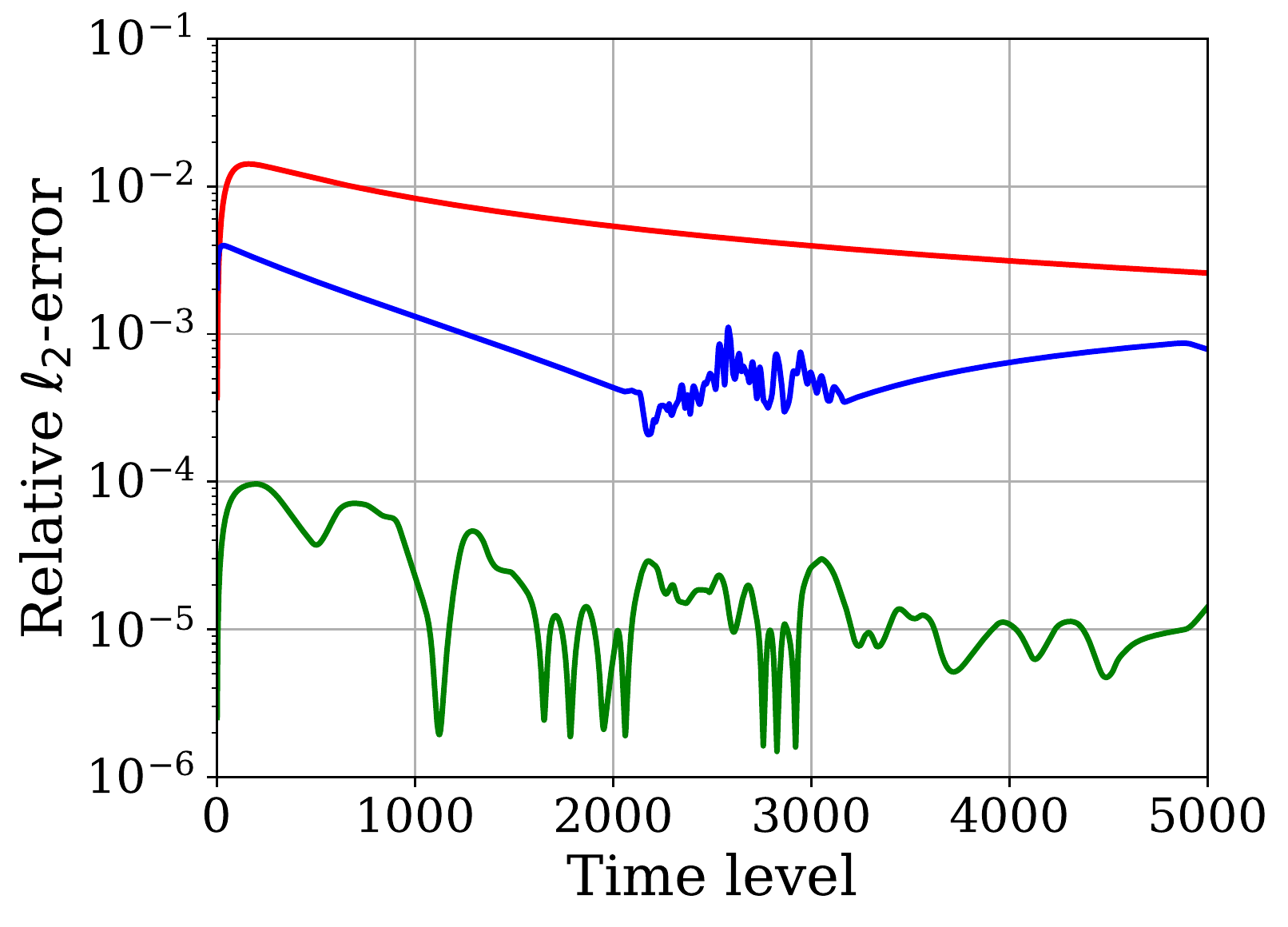}
		\vspace{-1.7em}
		\caption{2P1, $\alpha = 0.7$.}
		\vspace{0.4em}
		\label{subfig:2P1_error_a0.7}
	\end{subfigure}%
	\begin{subfigure}[b]{0.5\linewidth}
		\centering 
		\includegraphics[width=\textwidth]{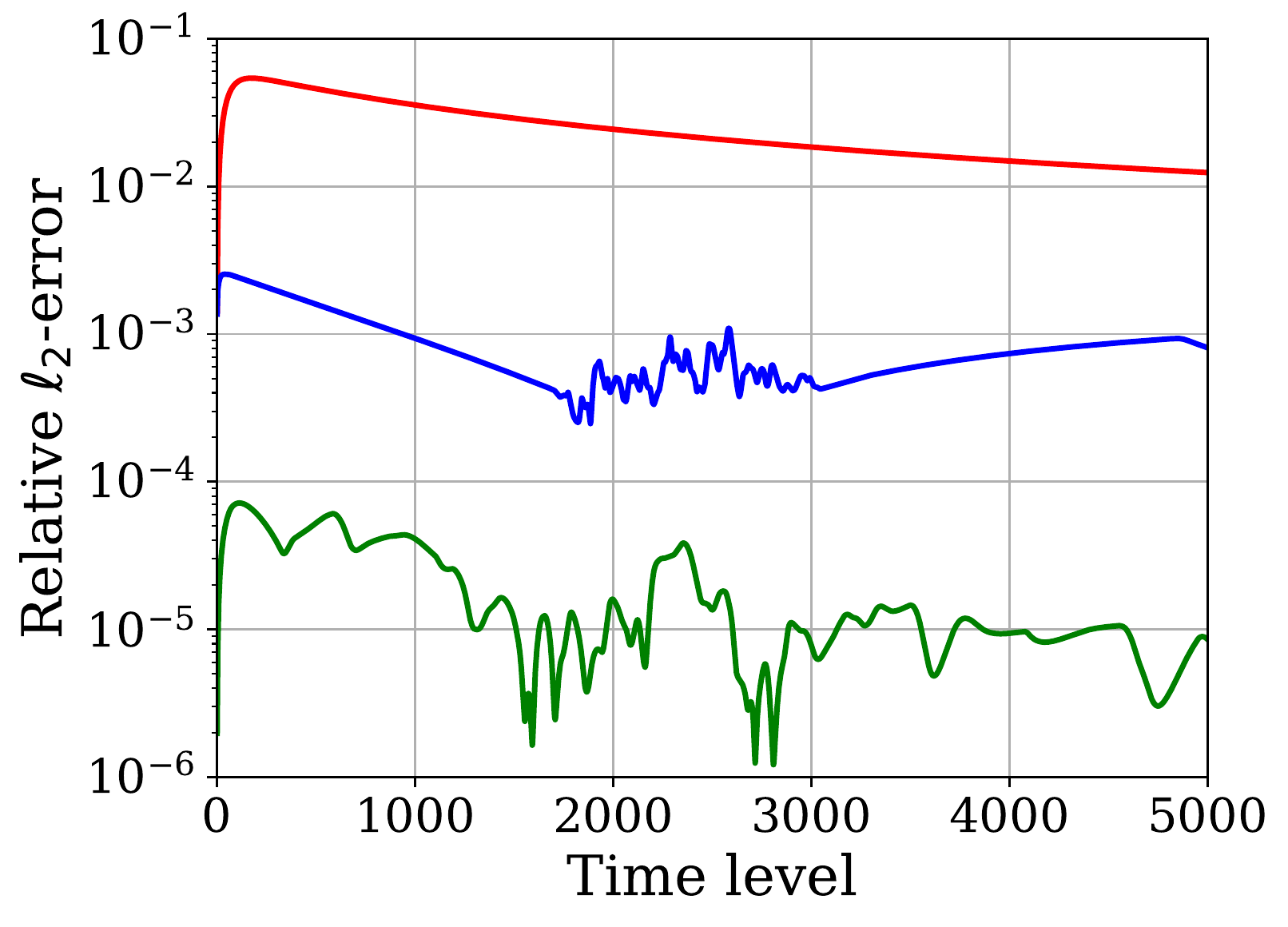}
		\vspace{-1.7em}
		\caption{2P1, $\alpha = 1.5$.}
		\vspace{0.4em}
		\label{subfig:2P1_error_a1.5}
	\end{subfigure}%
	\\
	\begin{subfigure}[b]{0.5\linewidth}
		\centering 
		\includegraphics[width=\textwidth]{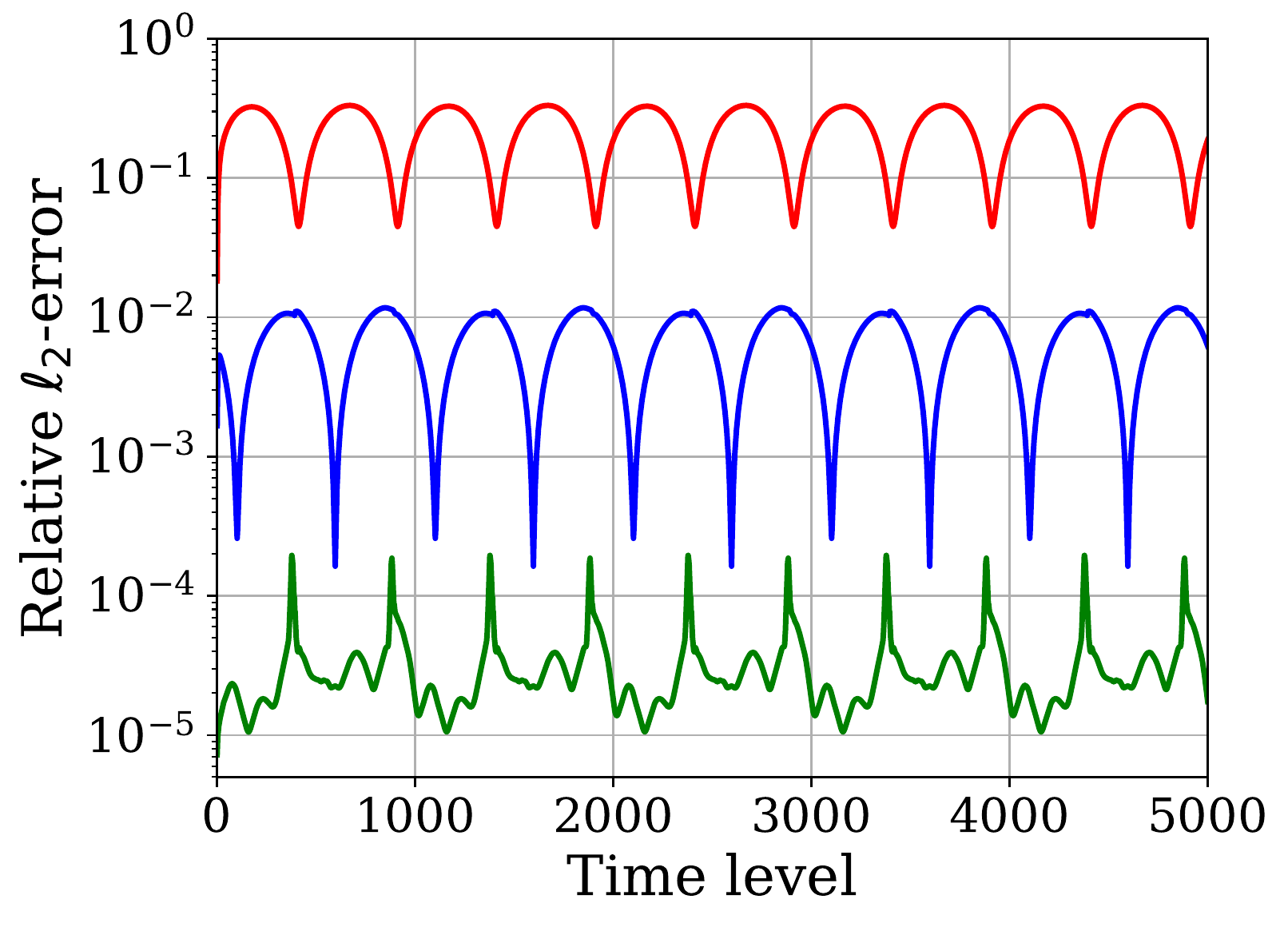}
		\vspace{-1.7em}
		\caption{2P2, $\alpha = 0.7$.}
		\label{subfig:2P2_error_a0.7}
	\end{subfigure}%
	\begin{subfigure}[b]{0.5\linewidth}
		\centering 
		\includegraphics[width=\textwidth]{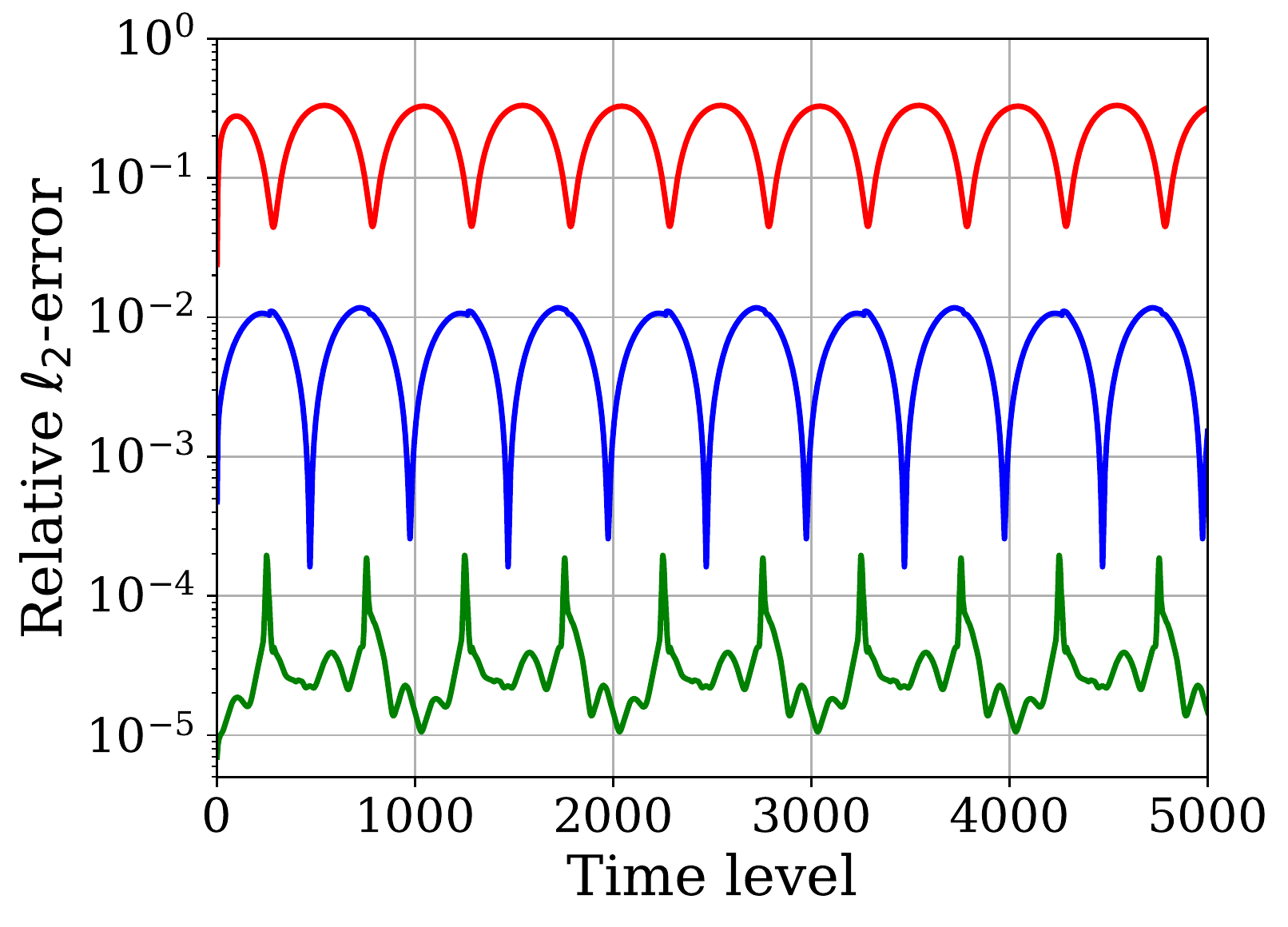}
		\vspace{-1.7em}
		\caption{2P2, $\alpha = 1.5$.}
		\label{subfig:2P2_error_a1.5}
	\end{subfigure}%
	\caption{Solutions 2P1 and 2P2, interpolation: Relative $\ell_2$-errors for $\alpha\in\{0.7,1.5\}$ (\redline~PBM, \blueline~DDM, \greenline~HAM).}
	\label{fig:2P_interp_errors}
	\vspace{7.0em}
\end{figure}

\begin{figure}
    \centering
    \includegraphics[width=\linewidth]{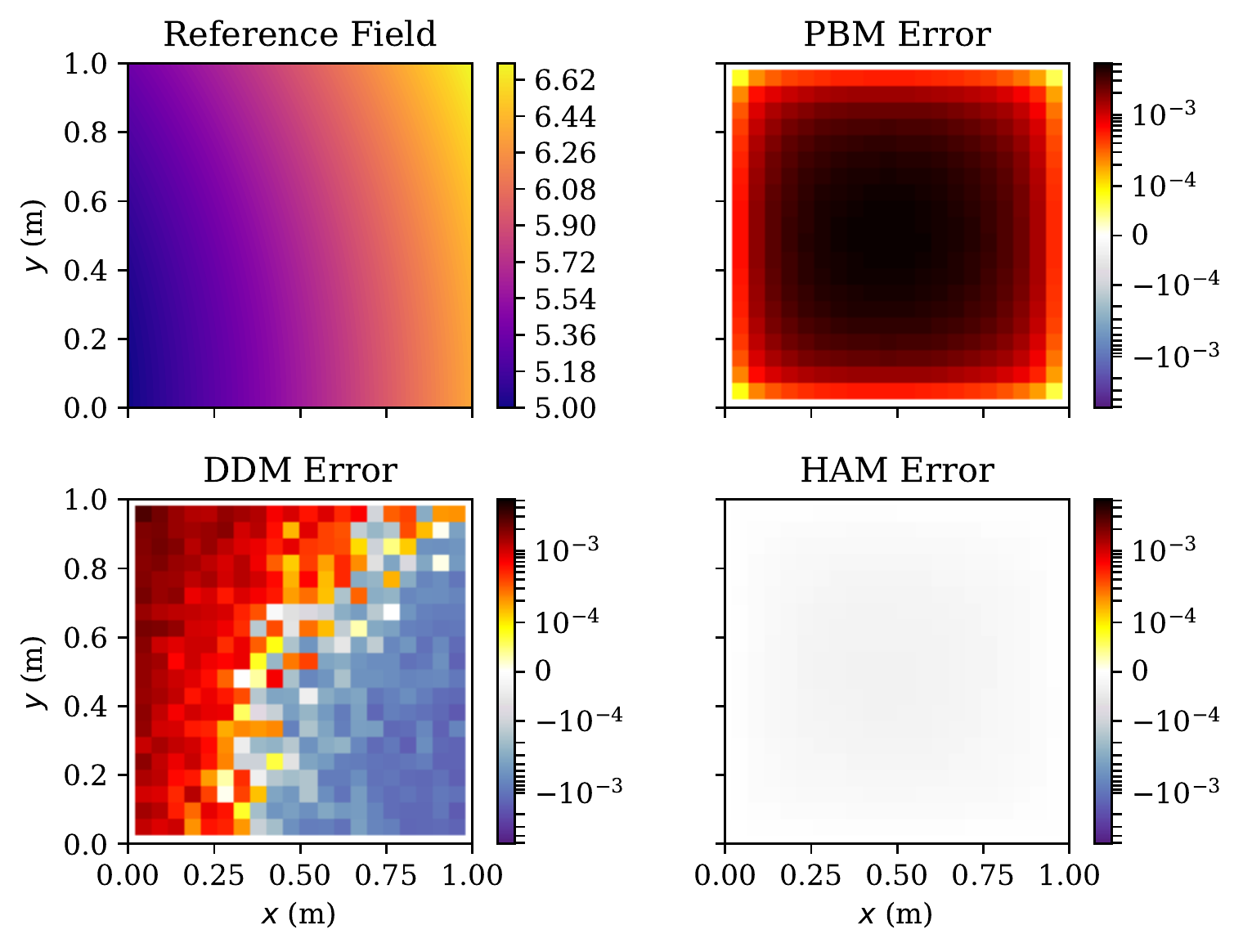}
    \caption{Solution 2P1, $\alpha=0.7$: Reference temperature field and relative $\ell_2$-errors of PBM, DDM and HAM.}
    \label{fig:2P1_profile_a0.7}
\end{figure}

\begin{figure}
    \centering
    \includegraphics[width=\linewidth]{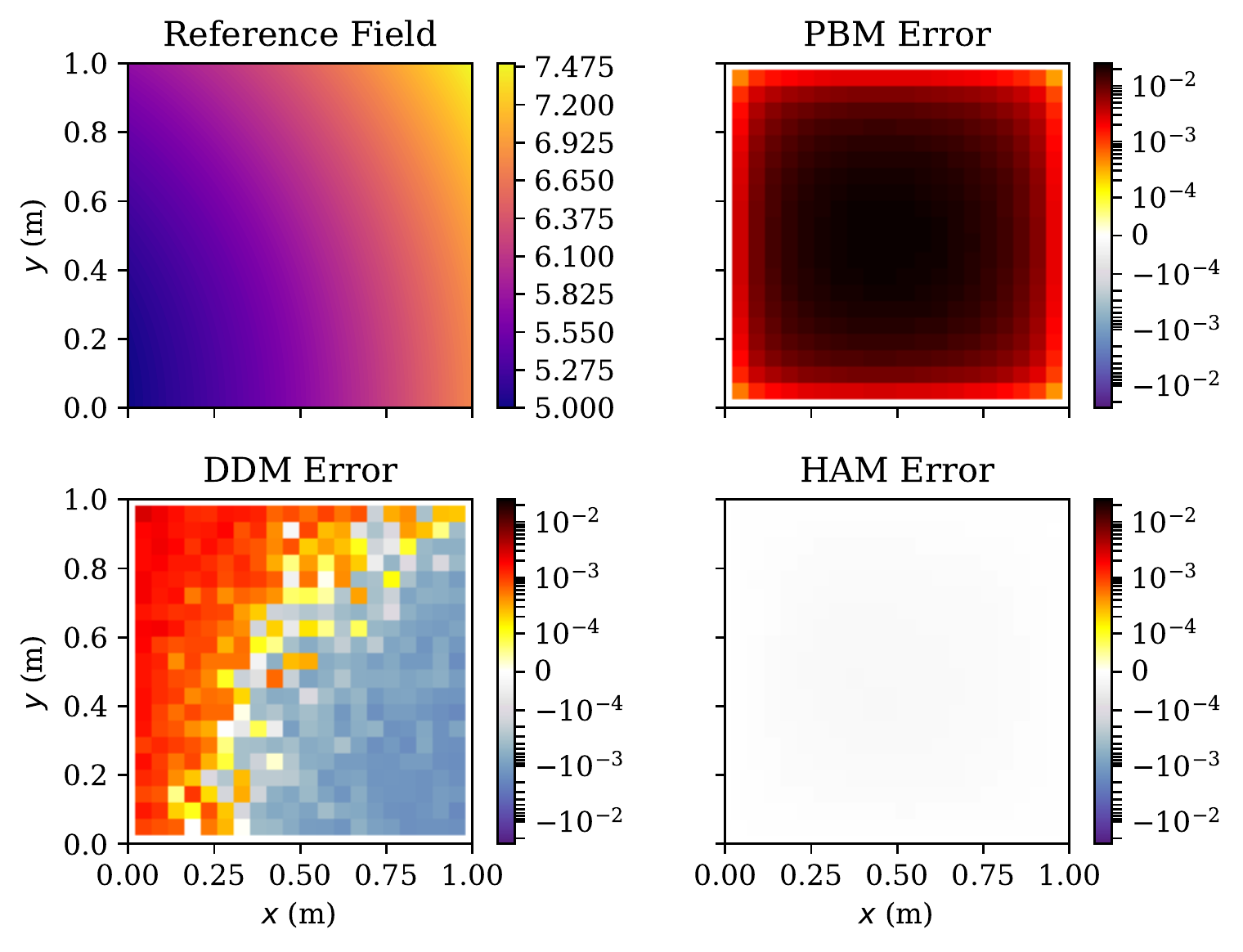}
    \caption{Solution 2P1, $\alpha=1.5$: Reference temperature field and relative $\ell_2$-errors of PBM, DDM and HAM.}
    \label{fig:2P1_profile_a1.5}
\end{figure}

\begin{figure}
    \centering
    \includegraphics[width=\linewidth]{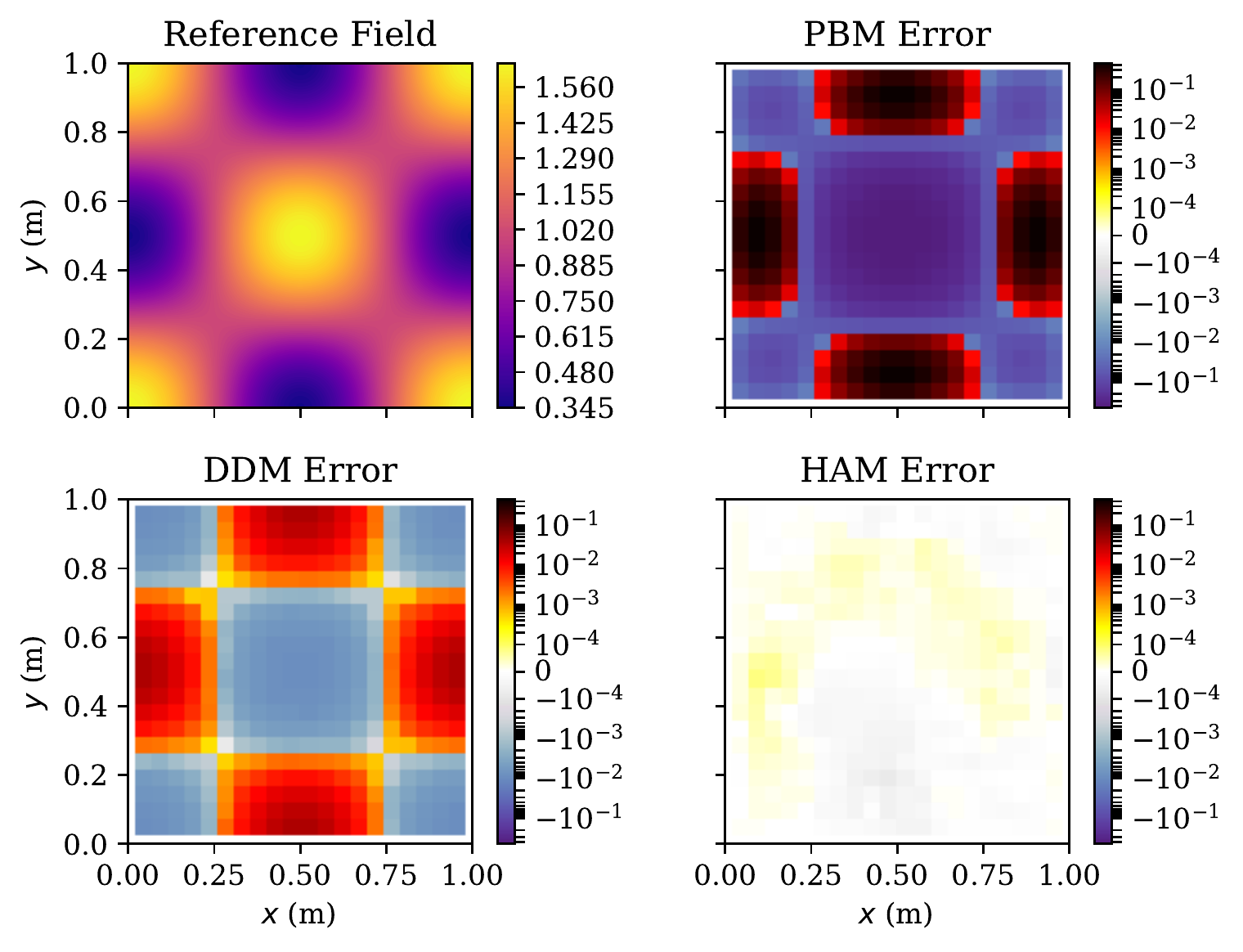}
    \caption{Solution 2P2, $\alpha=0.7$: Reference temperature field and relative $\ell_2$-errors of PBM, DDM and HAM.}
    \label{fig:2P2_profile_a0.7}
\end{figure}

\begin{figure}
    \centering
    \includegraphics[width=\linewidth]{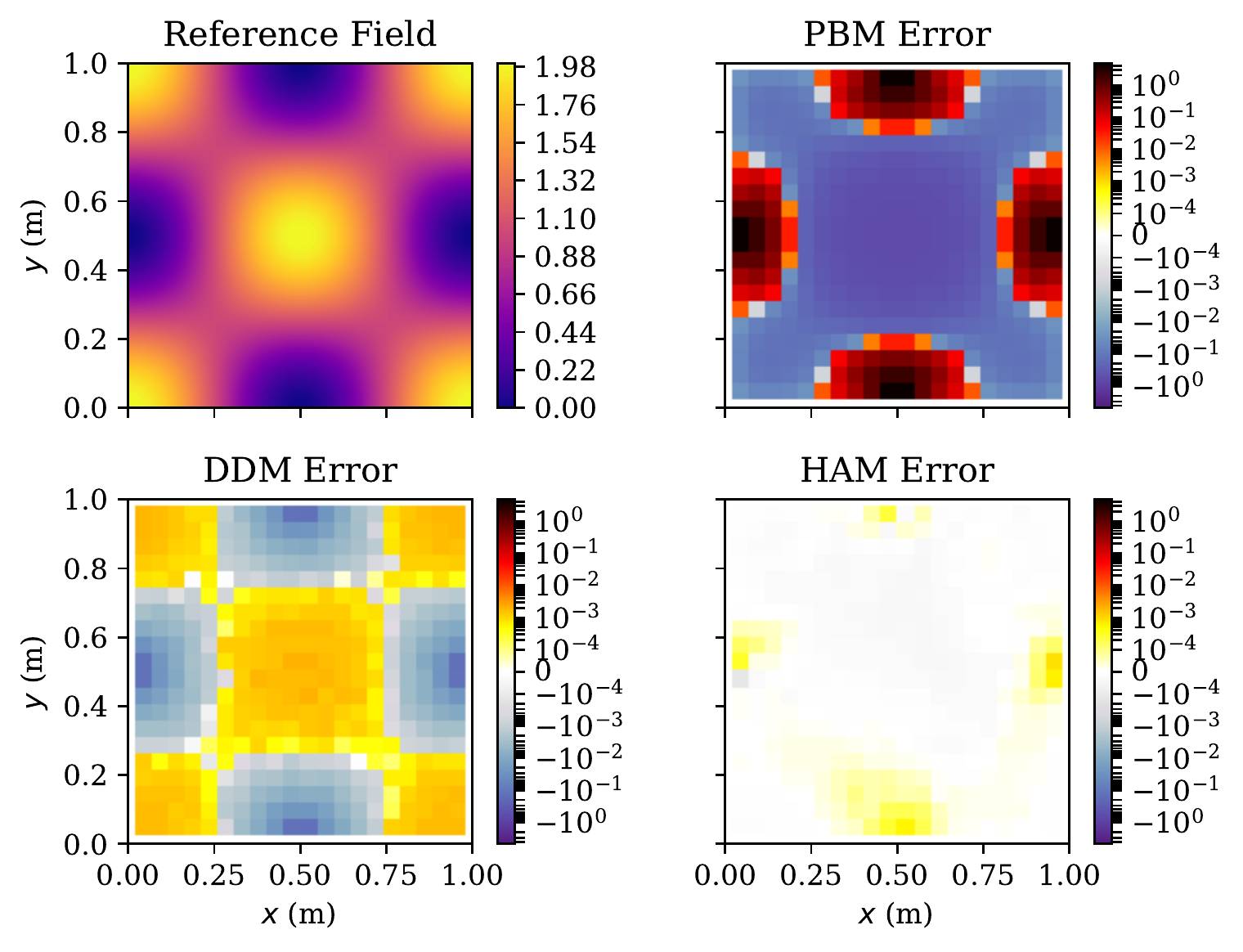}
    \caption{Solution 2P2, $\alpha=1.5$: Reference temperature field and relative $\ell_2$-errors of PBM, DDM and HAM.}
    \label{fig:2P2_profile_a1.5}
\end{figure}

The results for Solutions~2P1 and~2P2 in the interpolation scenarios are shown in Figures~\ref{fig:2P_interp_errors}--\ref{fig:2P2_profile_a1.5}. From the temporal development of the models' relative $\ell_2$-errors (Figure~\ref{fig:2P_interp_errors}), we see that the models follow a clear hierarchy in terms of accuracy. In all four cases, the PBM is the least accurate model. The PBM's accuracy is especially poor for Solution~2P2, for which it produces relative $\ell_2$-errors of up to 30\%. The DDM is the second most accurate model, producing relative $\ell_2$-errors which are, on average, roughly one order of magnitude smaller than those of the PBM. However, in all cases, the CoSTA-based HAM model is by far the most accurate model. We observe that the addition of the DNN-generated corrective source term yields an increase in accuracy of roughly three orders of magnitude compared to the unmodified PBM. We also observe that the CoSTA-based HAM model generally outperforms the DDM by more than one order of magnitude. The CoSTA model and the DDM use the same DNN using the same hyperparameters and the same training regime, so this result must imply that using a PBM to account for some physics, as is done in the CoSTA model, is more efficient than using the DNN to account for all physics, as is done in the DDM model. We notice that this holds true even when the accuracy of the PBM itself is poor (cf. the results for Solution~2P2).

%\subsubsection{Extrapolation Scenarios}

\begin{figure}[b!]
	\begin{subfigure}[b]{0.5\linewidth}
		\centering 
		\includegraphics[width=\textwidth]{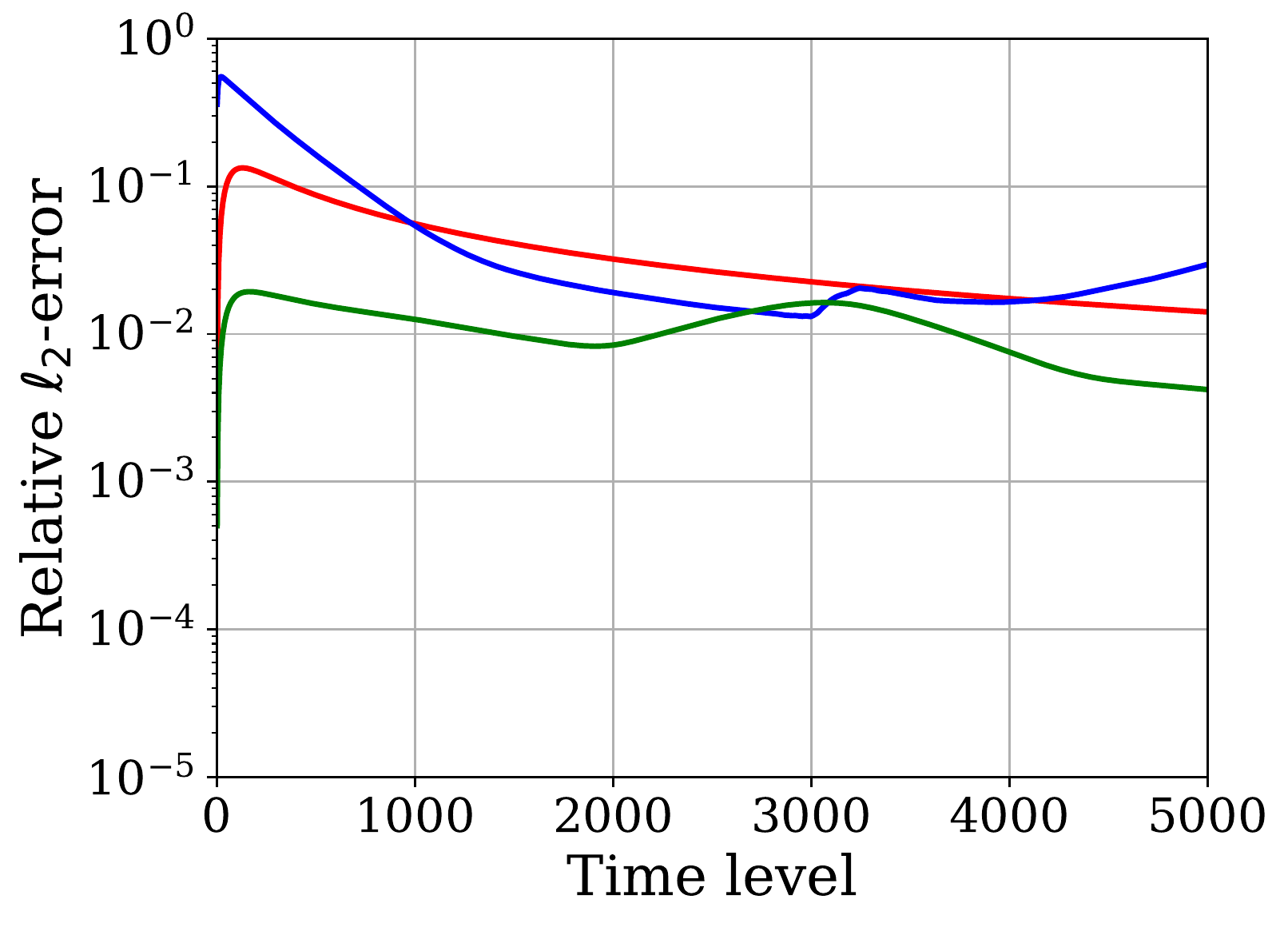}
		\vspace{-1.7em}
		\caption{2P1, $\alpha = -0.5$, relative errors.}
		\vspace{0.4em}
		\label{subfig:2P1_error_a-0.5}
	\end{subfigure}%
	\begin{subfigure}[b]{0.5\linewidth}
		\centering 
		\includegraphics[width=\textwidth]{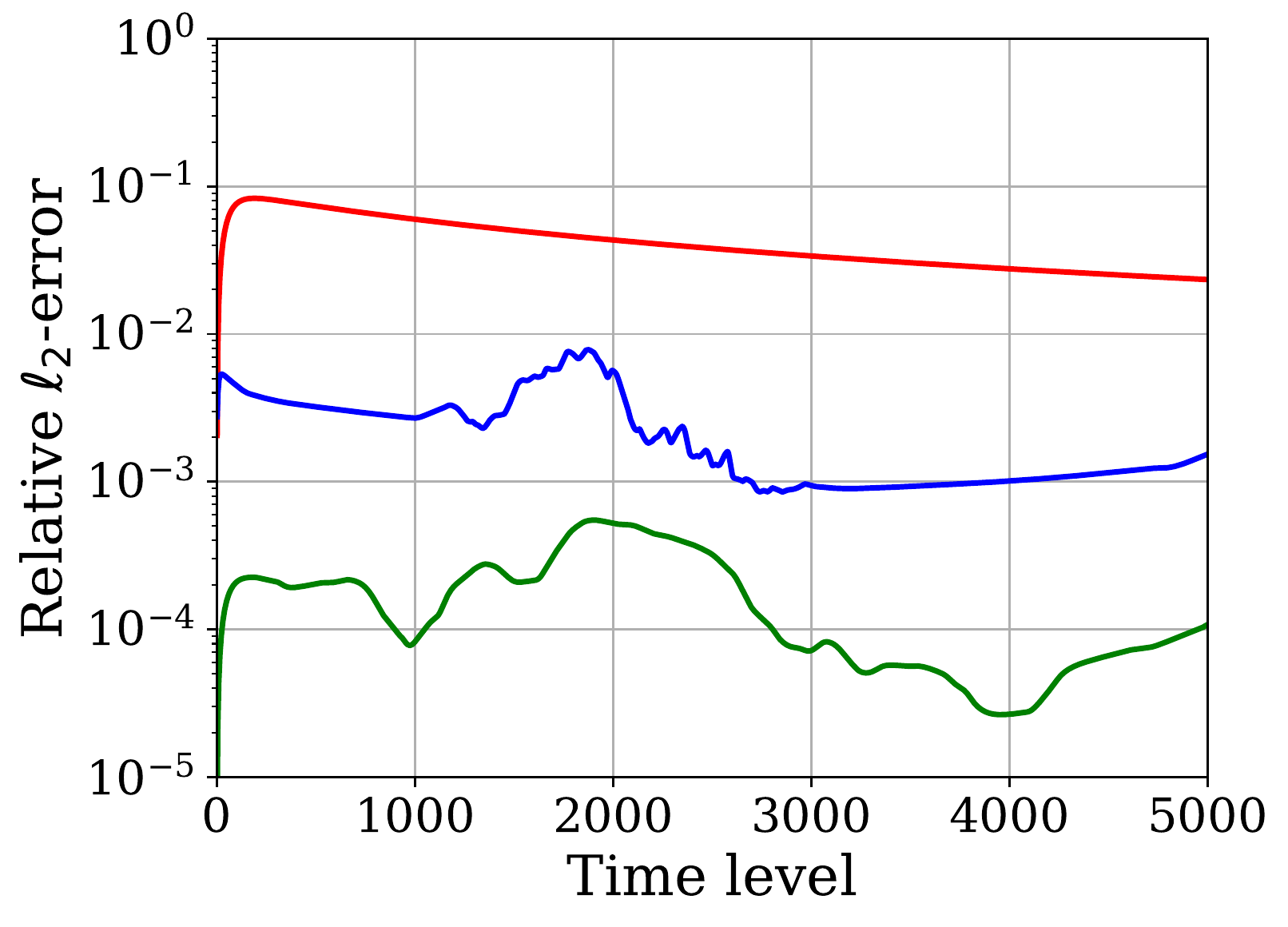}
		\vspace{-1.7em}
		\caption{2P1, $\alpha = 2.5$, relative errors.}
		\vspace{0.4em}
		\label{subfig:2P1_error_a2.5}
	\end{subfigure}%
	\\
	\begin{subfigure}[b]{0.5\linewidth}
		\centering 
		\includegraphics[width=\textwidth]{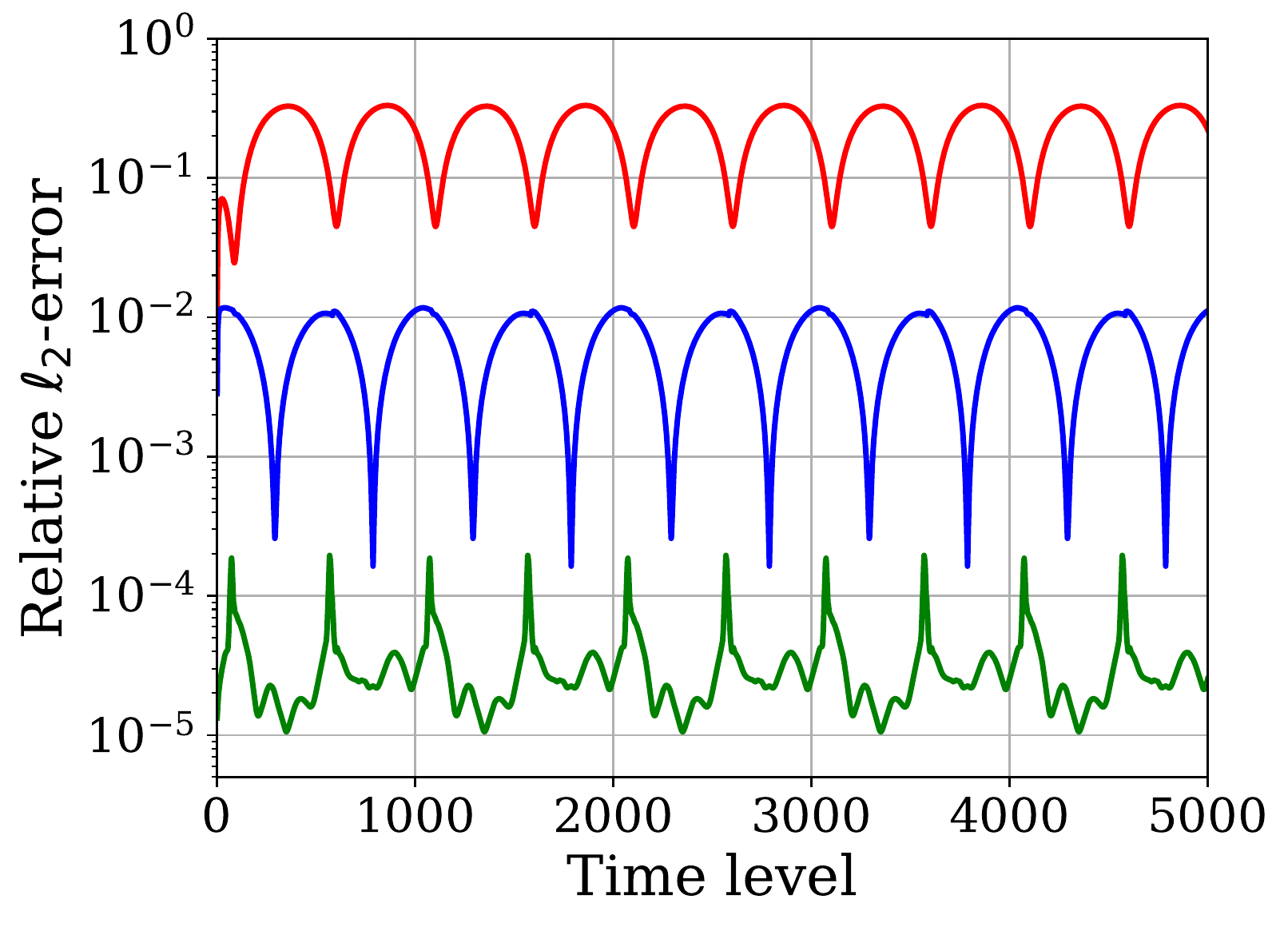}
		\vspace{-1.7em}
		\caption{2P2, $\alpha = -0.5$, relative errors.}
		\label{subfig:2P2_error_a-0.5}
	\end{subfigure}%
	\begin{subfigure}[b]{0.5\linewidth}
		\centering 
		\includegraphics[width=\textwidth]{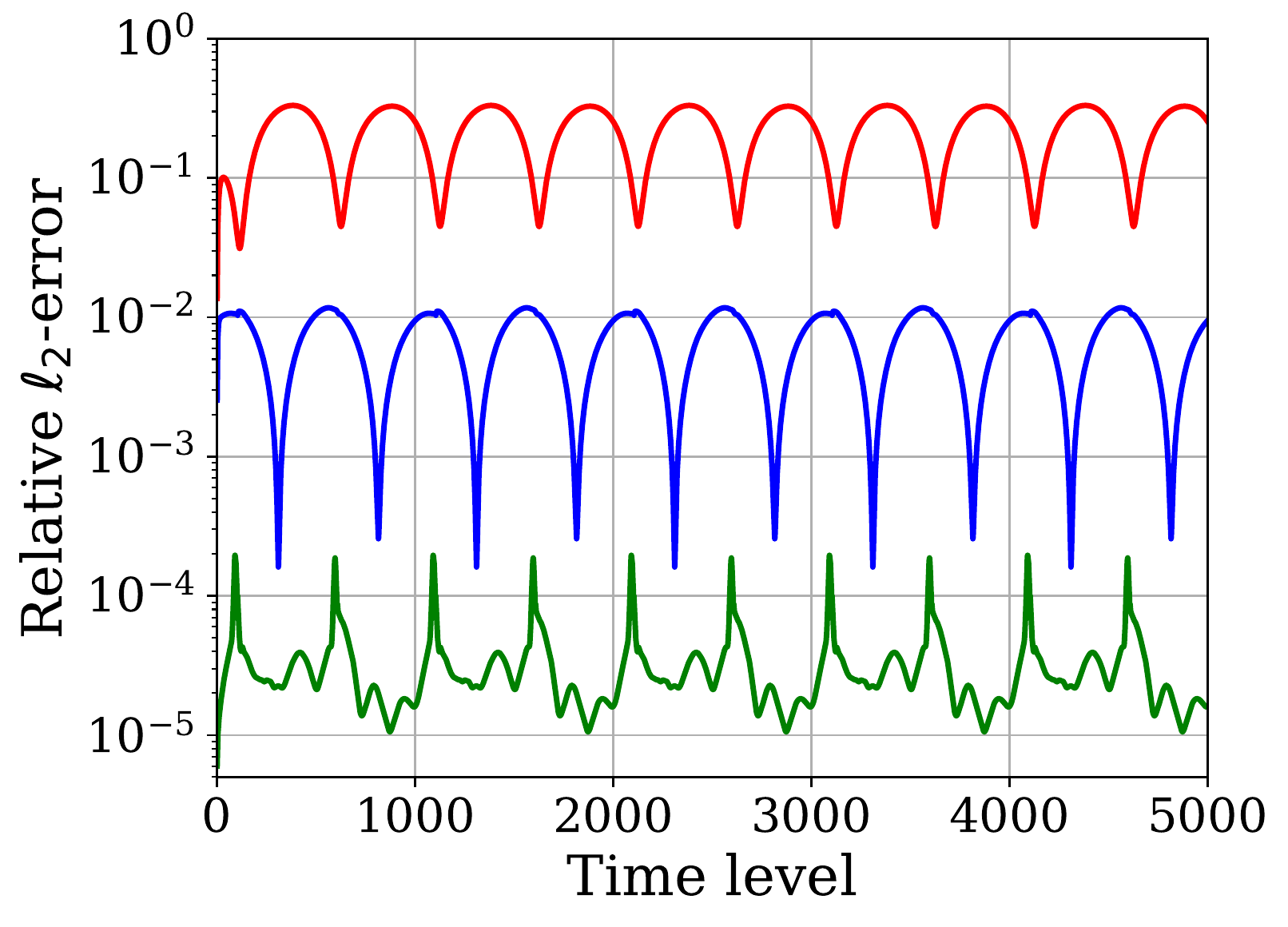}
		\vspace{-1.7em}
		\caption{2P2, $\alpha = 2.5$, relative errors.}
		\label{subfig:2P2_error_a2.5}
	\end{subfigure}%
	\caption{Solutions 2P1 and 2P2, extrapolation: Relative $\ell_2$-errors for $\alpha\in\{0.7,1.5\}$ (\redline~PBM, \blueline~DDM, \greenline~HAM).}
	\label{fig:2P_extrap_errors} 
\end{figure}

\begin{figure}
    \centering
    \includegraphics[width=\linewidth]{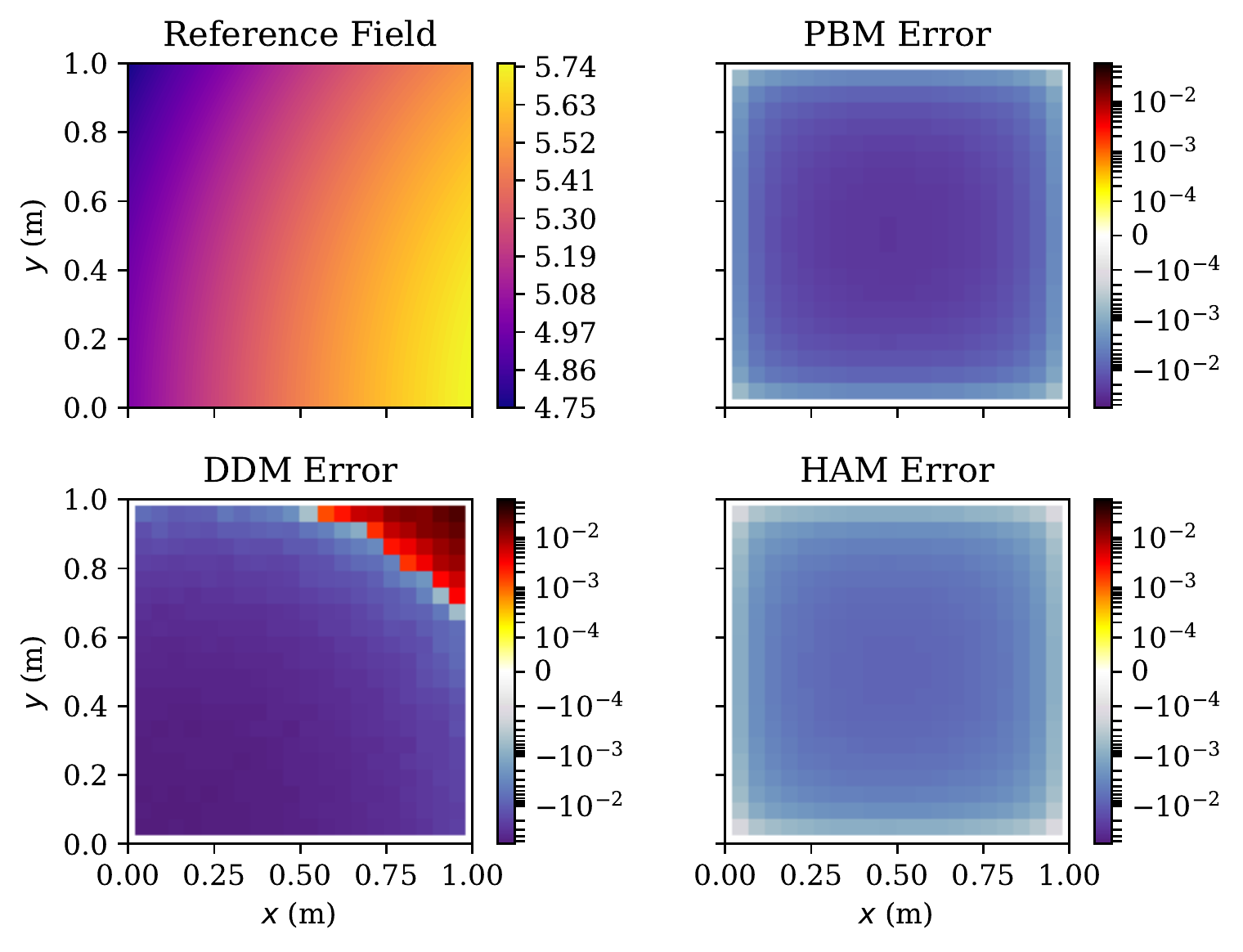}
    \caption{Solution 2P1, $\alpha=-0.5$: Reference temperature field and relative $\ell_2$-errors of PBM, DDM and HAM.}
    \label{fig:2P1_profile_a-0.5}
\end{figure}

\begin{figure}
    \centering
    \includegraphics[width=\linewidth]{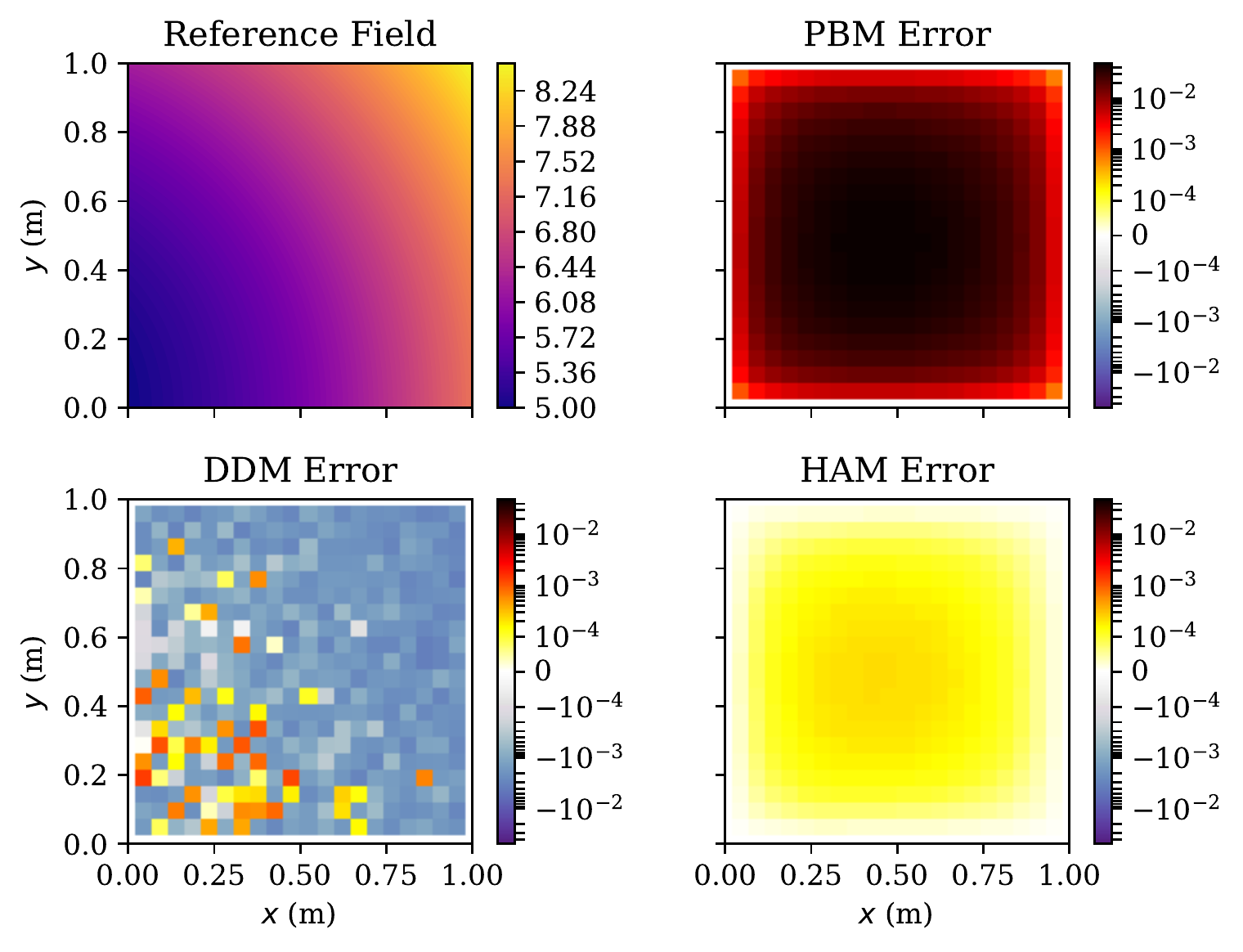}
    \caption{Solution 2P1, $\alpha=2.5$: Reference temperature field and relative $\ell_2$-errors of PBM, DDM and HAM.}
    \label{fig:2P1_profile_a2.5}
\end{figure}

\begin{figure}
    \centering
    \includegraphics[width=\linewidth]{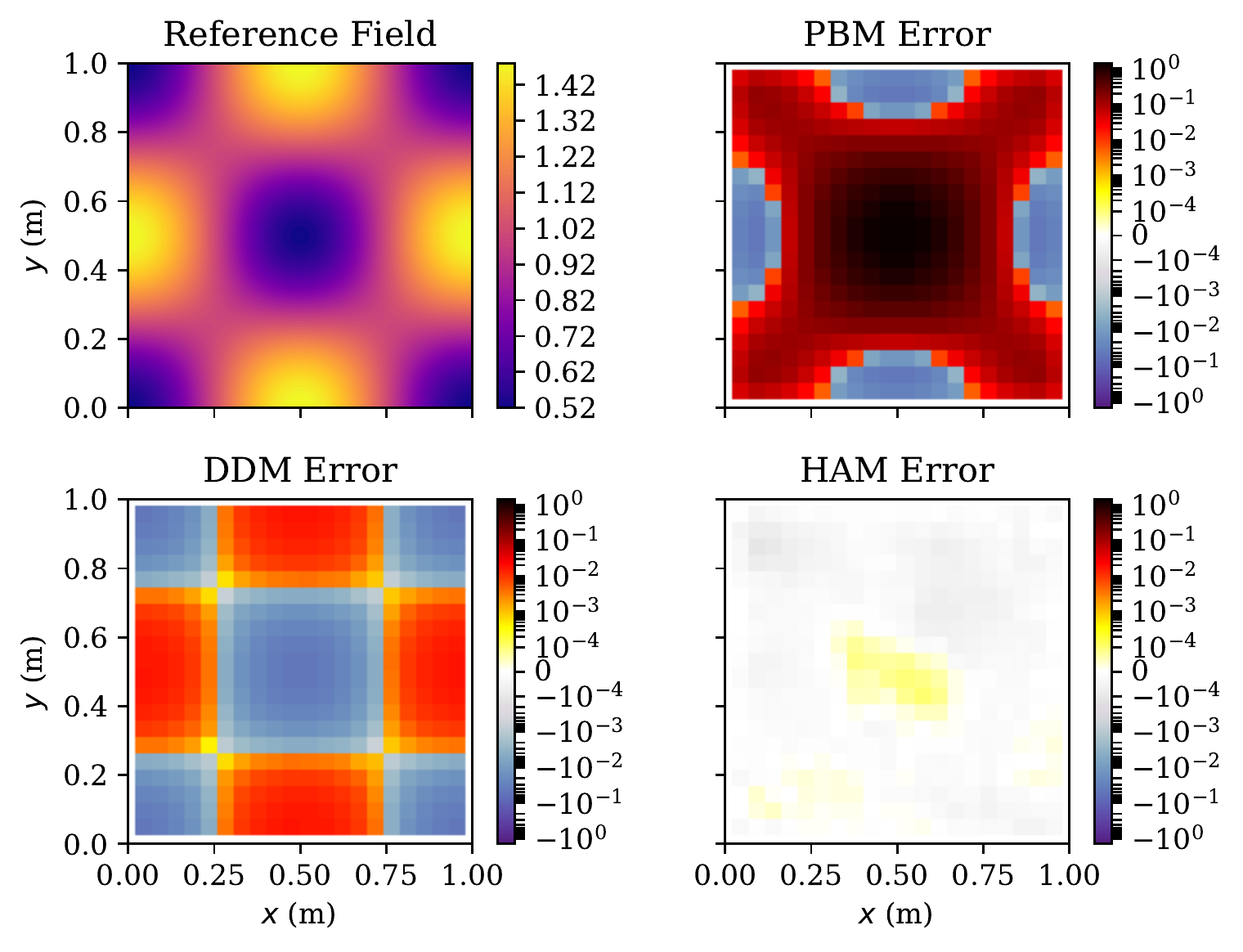}
    \caption{Solution 2P2, $\alpha=-0.5$: Reference temperature field and relative $\ell_2$-errors of PBM, DDM and HAM.}
    \label{fig:2P2_profile_a-0.5}
\end{figure}

\begin{figure}
    \centering
    \includegraphics[width=\linewidth]{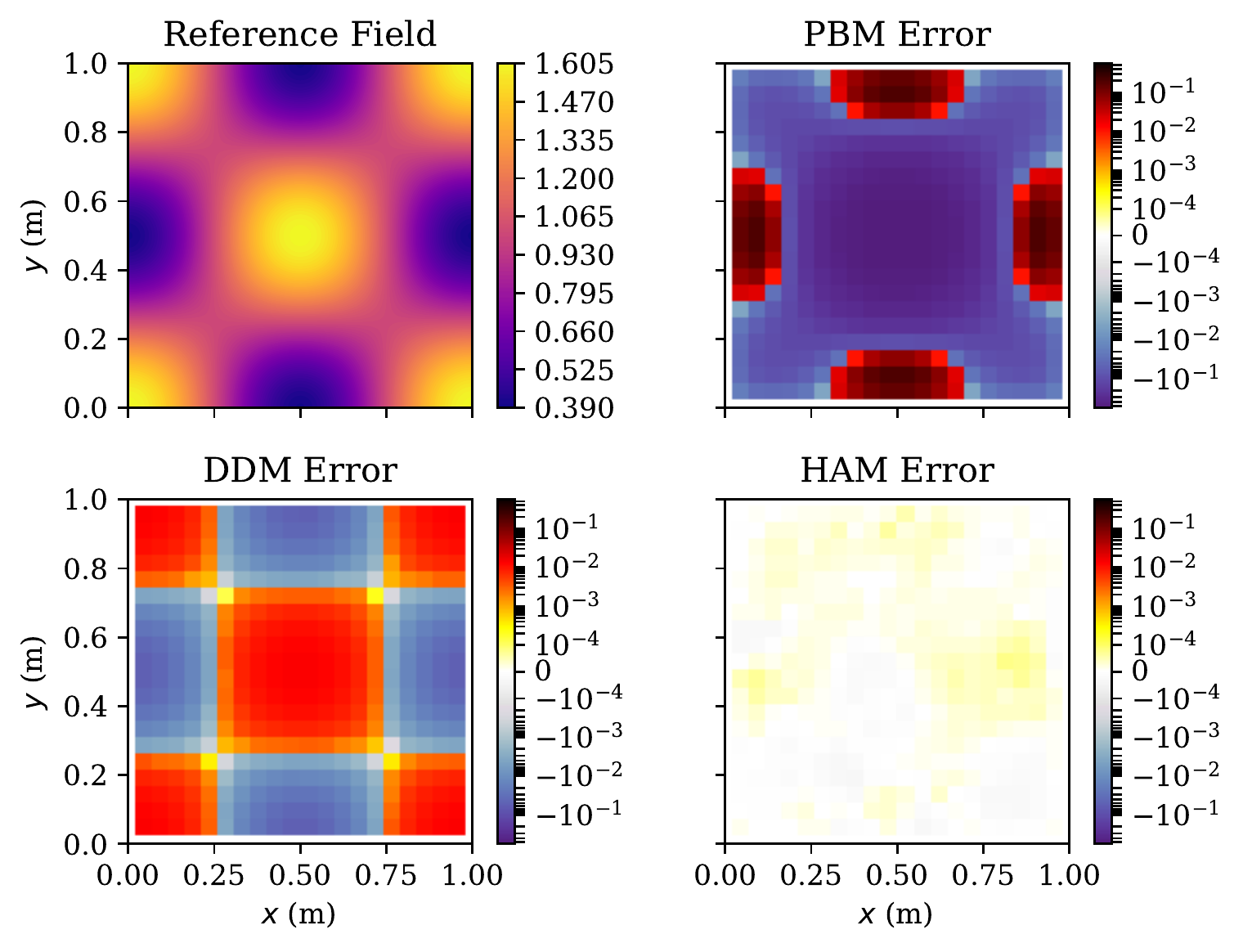}
    \caption{Solution 2P2, $\alpha=2.5$: Reference temperature field and relative $\ell_2$-errors of PBM, DDM and HAM.}
    \label{fig:2P2_profile_a2.5}
\end{figure}

The extrapolation scenario results for Solutions~2P1 and~2P2 are shown in Figures~\ref{fig:2P_extrap_errors}--\ref{fig:2P2_profile_a2.5}. By comparing the $\ell_2$-errors for the extrapolation scenarios (cf. Figure~\ref{fig:2P_extrap_errors}) and the interpolation scenarios (cf. Figure~\ref{fig:2P_interp_errors}), we see that PBM is the most generalizable model, in the sense that its results for the interpolation and extrapolation scenarios are the most similar. In fact, there is no significant difference between the PBM's accuracy in the extrapolation scenarios and in the interpolation scenarios. However, it should be noted that the PBM is still the least accurate model overall. For Solution~2P2, DDM and HAM exhibit roughly the same level of accuracy in the interpolation scenarios, and this can be explained by the fact that Solution~2P2 is not qualitatively different in the extrapolation scenarios than in the interpolation scenarios. The qualitative difference between the scenarios is much greater for Solution~2P1, and this is clearly reflected in the DDM and HAM results. Both DDM and HAM suffer significant accuracy reduction going from interpolation to extrapolation for this solution. However, CoSTA-based HAM is still the most accurate model overall.

It is of particular interest to study the error fields corresponding to Solution~2P1, which are illustrated in Figures~\ref{fig:2P1_profile_a-0.5} and~\ref{fig:2P1_profile_a2.5}. We observe that the PBM and HAM error fields are smooth and quite uniform throughout the spatial domain, while this is not the case for the DDM error fields. For $\alpha=-0.5$, the DDM prediction is significantly too hot in the top right corner. For $\alpha > 0$, that corner is the warmest corner, so this error illustrates a failure to generalize which is not observed for the PBM and CoSTA models. 
For $\alpha=2.5$, we observe that the DDM error field is noisy. Since the DNN is trained using an $\ell_2$ loss function which does not enforce smooth DNN output, this is not really surprising. However, it is worth noting that HAM is not affected by noise from the DNN to the same extent as the purely data-driven model.

Looking at the error curves for System~2P2 in Figures~\ref{fig:2P_interp_errors} and~\ref{fig:2P_extrap_errors}, the curious reader may wonder why the error curves of the PBM are ``out of phase'' in comparison to the DDM and HAM curves. A careful examination reveals that the low-points of the PBM error curves correspond to temporal locations where $P\approx0$. Since the PBM assumes $P=0$, this is sensible. Still, it may appear counter-intuitive that the DDM and HAM models perform worst for this simple situation. The reason is probably related to the training of the DNNs; most training examples seen by the DNNs correspond to curved temperature fields, so the DNNs are not well-trained for handling the special case $P\approx0$ corresponding to virtually flat temperature fields. If the case $P\approx0$ is of particular importance, data augmentation could be beneficial in obtaining more accurate predictions in this case.

\subsection{Experiments with Unknown Conductivity}\label{subsec:exp_2D_k}

We now move on to our two experiments where the conductivity $k$ is taken to be unknown. Such scenarios are found, for example, when studying composite systems, mixtures or other inhomogeneous systems. In such cases, accurately determining the thermal conductivity at all locations within the system can be forbiddingly challenging, or even unfeasible. In this section, we consider the two manufactured solutions~2k1 and~2k2 (cf.~\ref{tab:manufactured_solutions_2D}), whose corresponding conductivity profiles are linear and periodic in space, respectively. As in the previous section, we discuss the interpolation scenarios first and the extrapolation scenarios thereafter.

\begin{figure}
	\begin{subfigure}[b]{0.5\linewidth}
		\centering 
		\includegraphics[width=\textwidth]{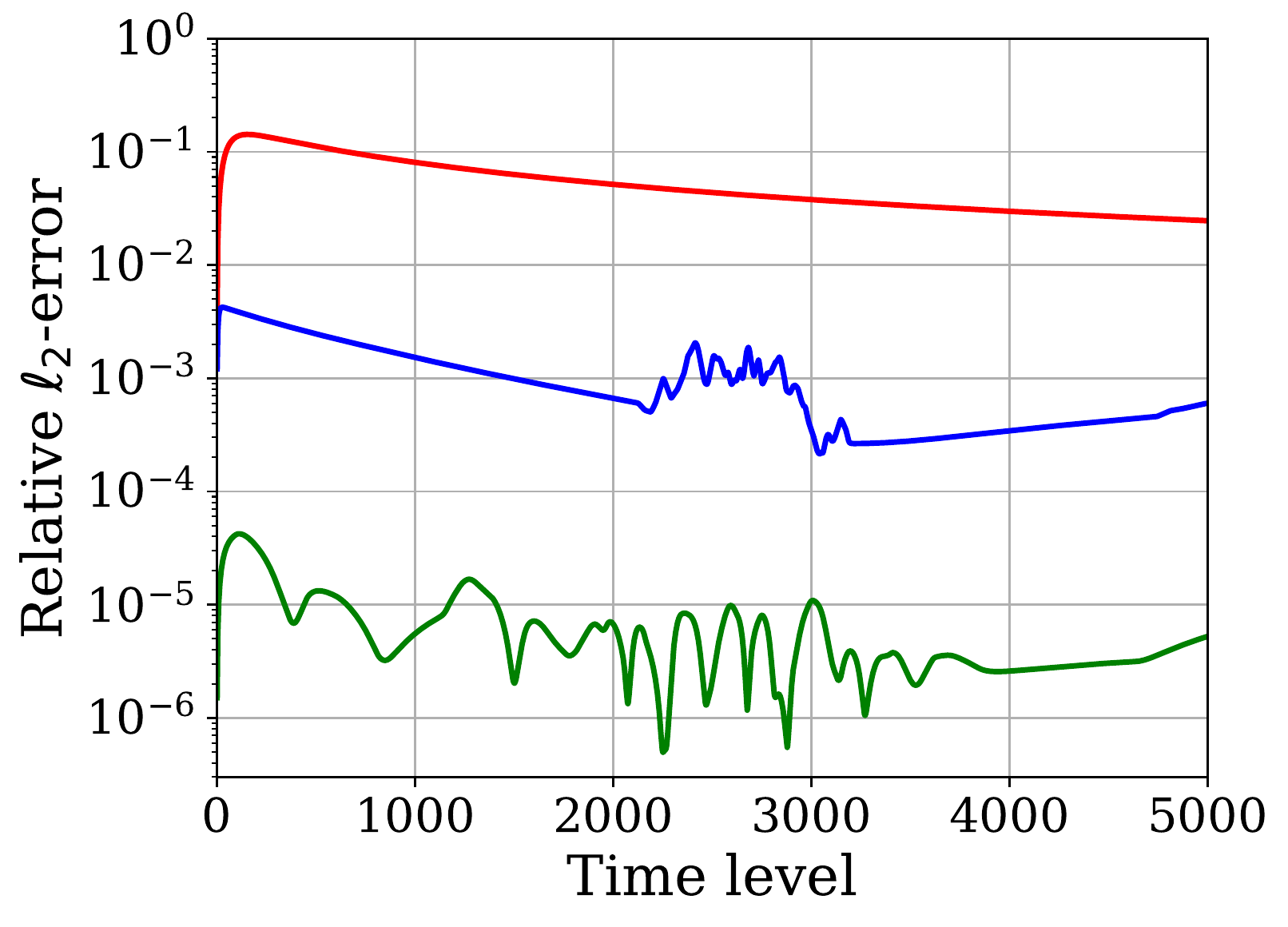}
		\vspace{-1.7em}
		\caption{2k1, $\alpha = 0.7$, relative errors.}
		\vspace{0.4em}
		\label{subfig:2k1_error_a0.7}
	\end{subfigure}%
	\begin{subfigure}[b]{0.5\linewidth}
		\centering 
		\includegraphics[width=\textwidth]{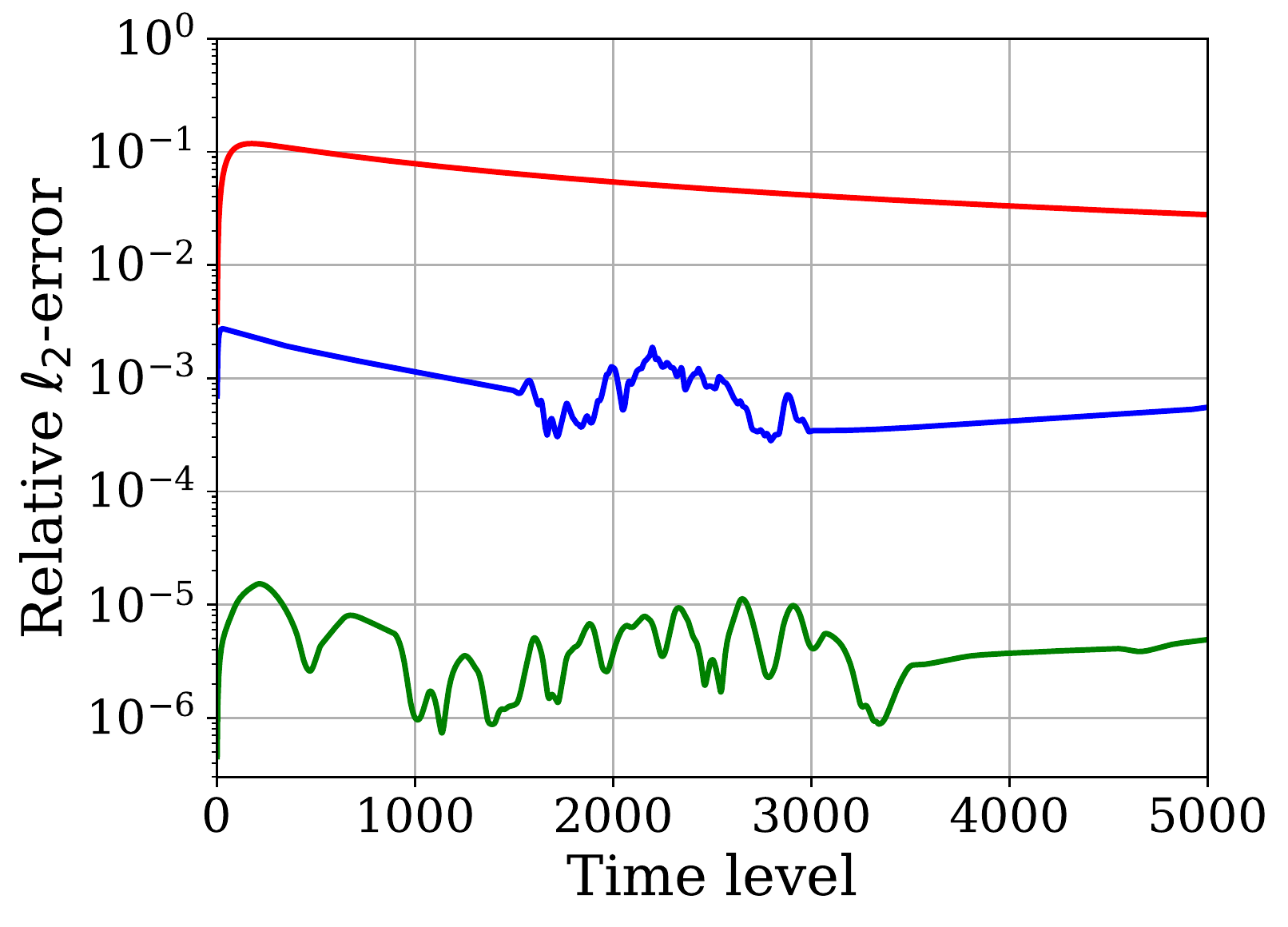}
		\vspace{-1.7em}
		\caption{2k1, $\alpha = 1.5$, relative errors.}
		\vspace{0.4em}
		\label{subfig:2k1_error_a1.5}
	\end{subfigure}%
	\\
	\begin{subfigure}[b]{0.5\linewidth}
		\centering 
		\includegraphics[width=\textwidth]{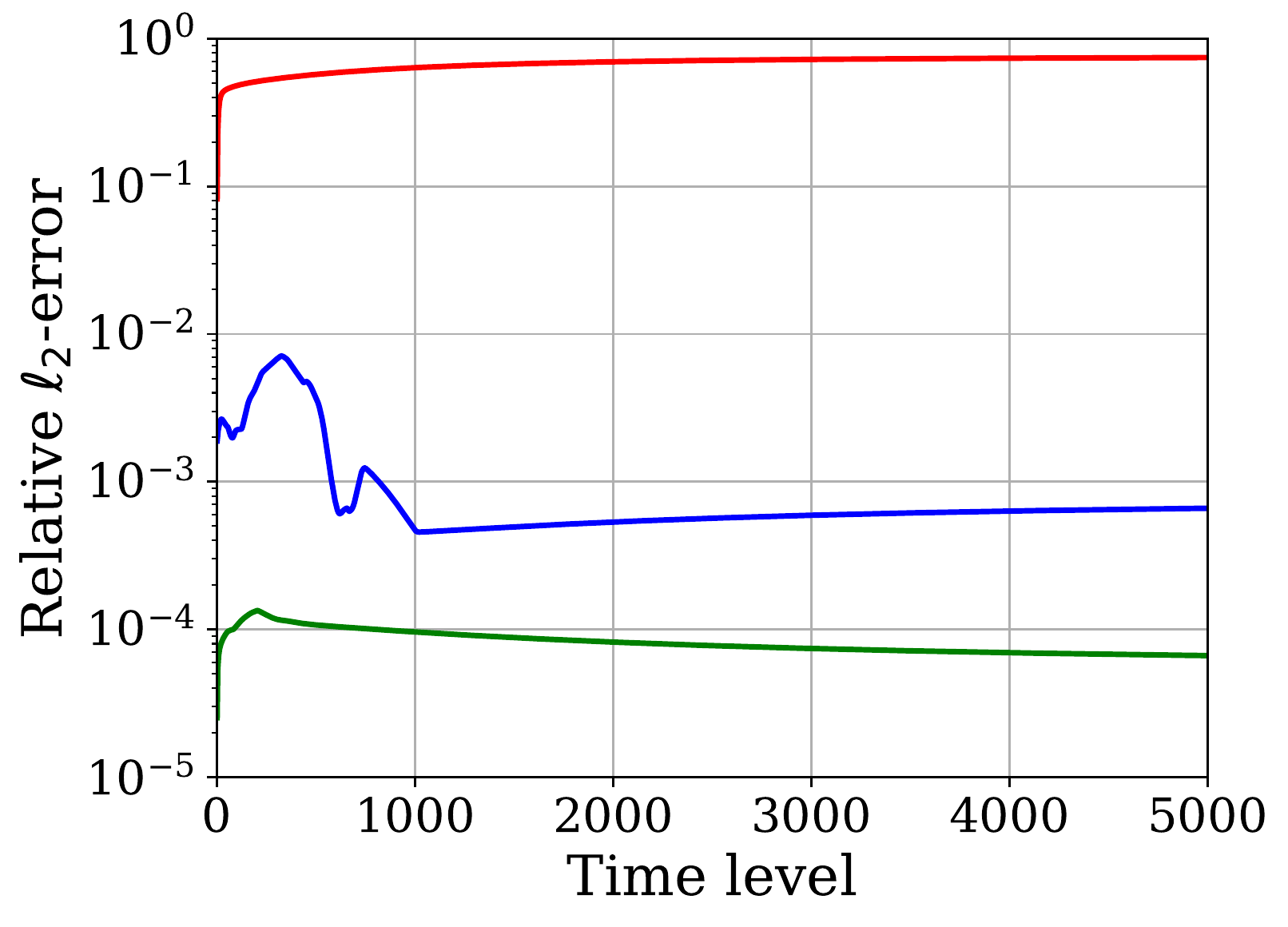}
		\vspace{-1.7em}
		\caption{2k2, $\alpha = 0.7$, relative errors.}
		\label{subfig:2k2_error_a0.7}
	\end{subfigure}%
	\begin{subfigure}[b]{0.5\linewidth}
		\centering 
		\includegraphics[width=\textwidth]{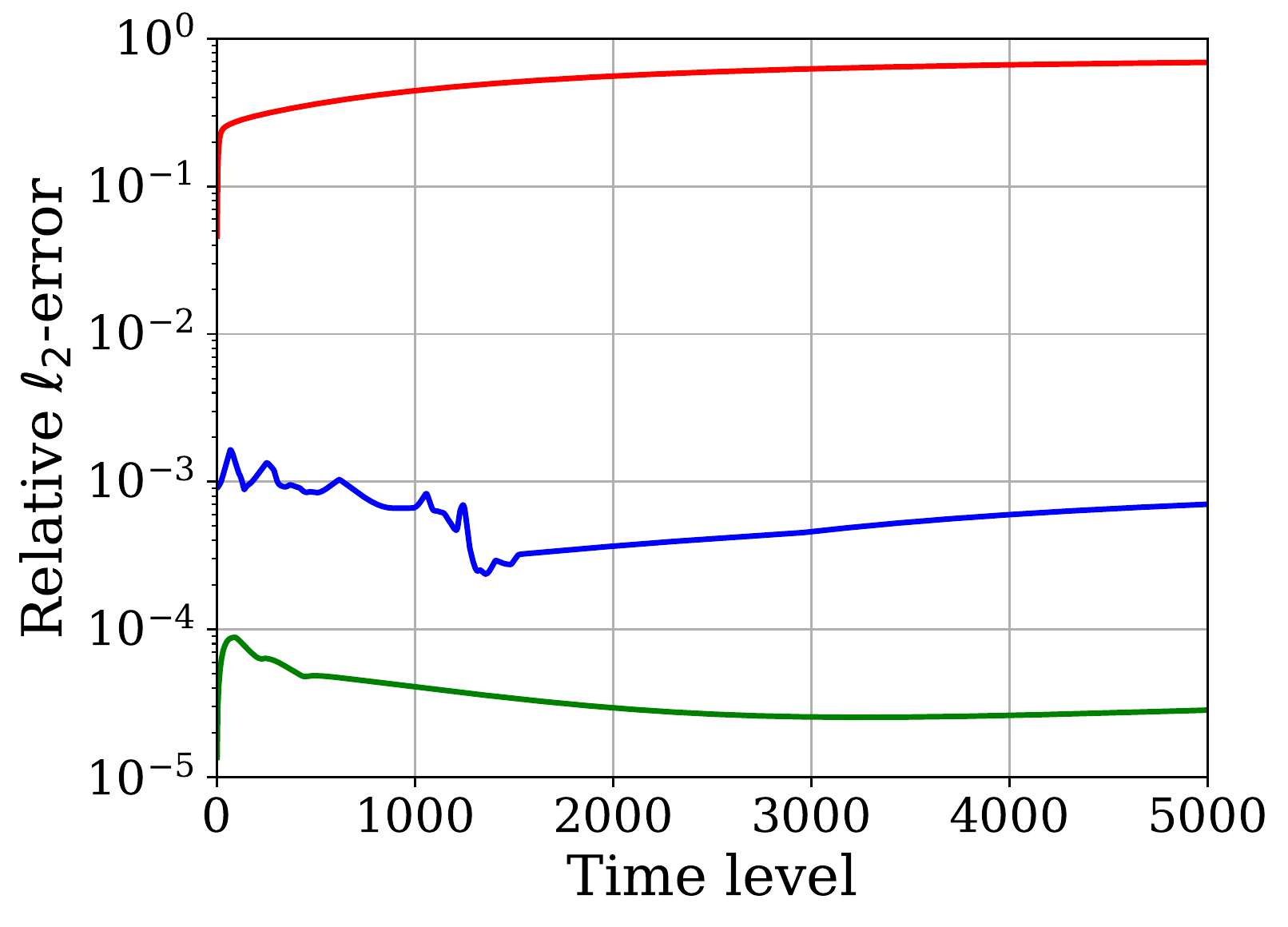}
		\vspace{-1.7em}
		\caption{2k2, $\alpha = 1.5$, relative errors.}
		\label{subfig:2k2_error_a1.5}
	\end{subfigure}%
	\caption{Solutions 2k1 and 2k2, interpolation: Relative $\ell_2$-errors for $\alpha\in\{0.7,1.5\}$ (\redline~PBM, \blueline~DDM, \greenline~HAM).}
	\label{fig:2k_interp_errors} 
\end{figure}

\begin{figure}
    \centering
    \includegraphics[width=\linewidth]{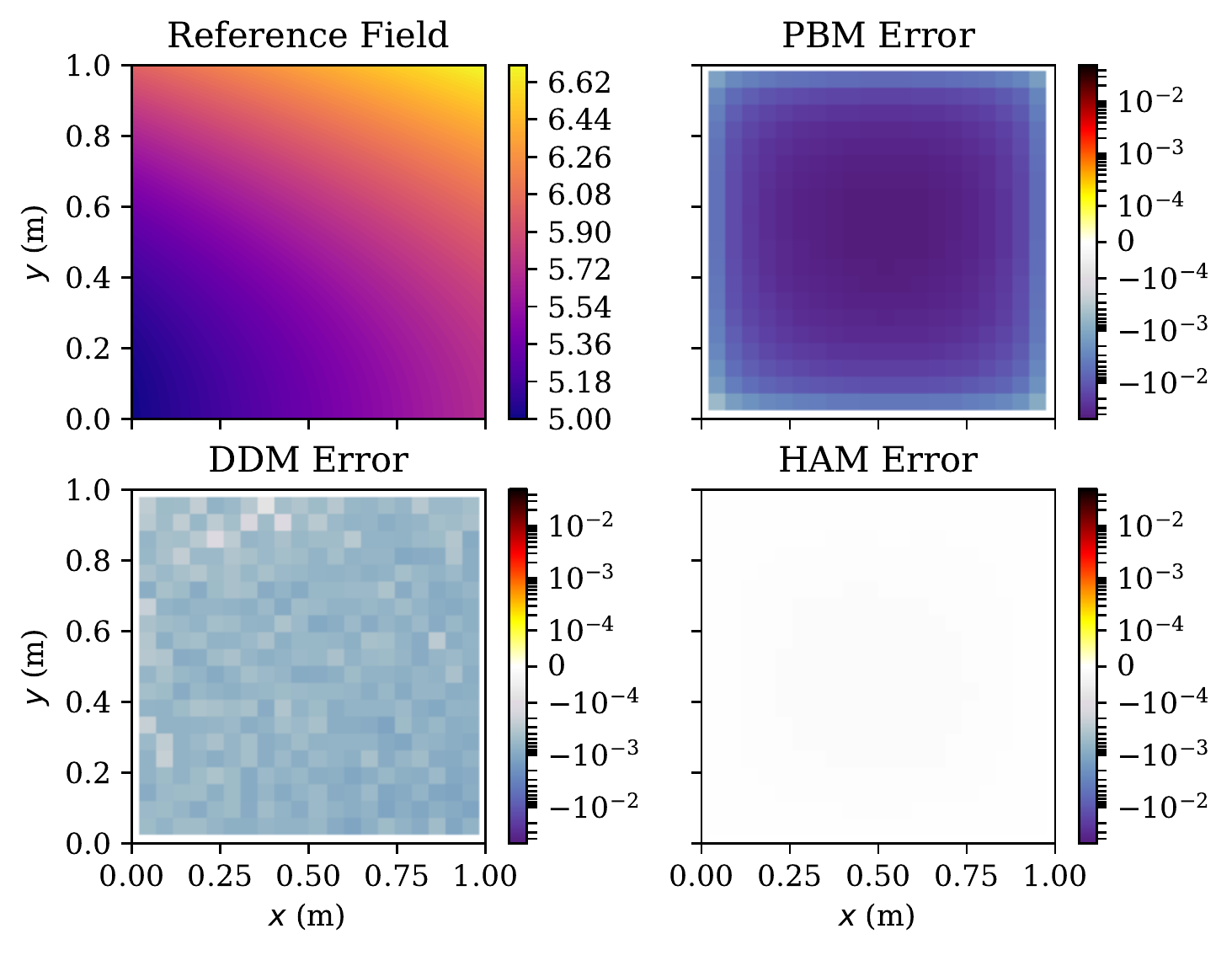}
    \caption{Solution 2k1, $\alpha=0.7$: Reference temperature field and relative $\ell_2$-errors of PBM, DDM and HAM.}
    \label{fig:2K1_profile_a0.7}
\end{figure}

\begin{figure}
    \centering
    \includegraphics[width=\linewidth]{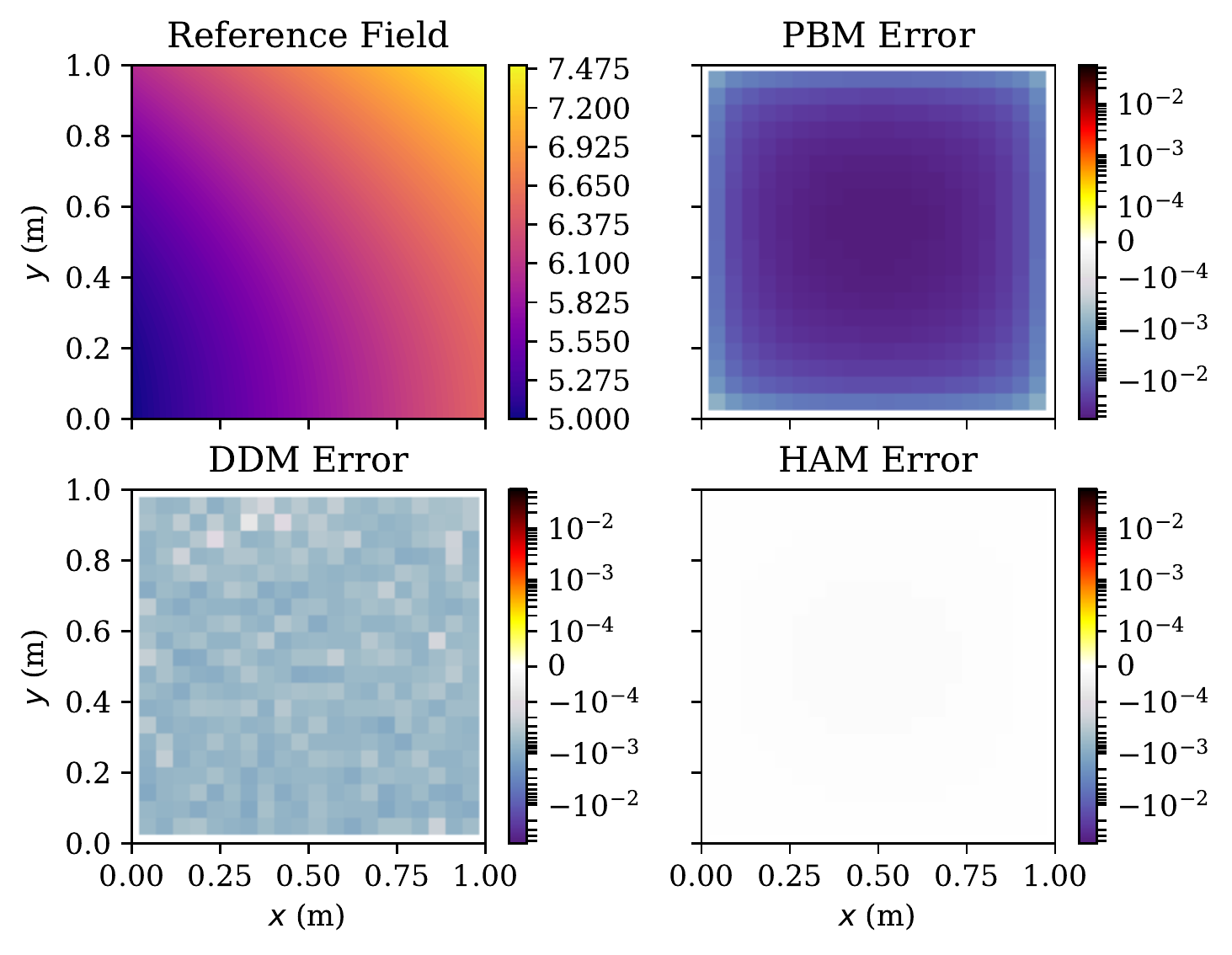}
    \caption{Solution 2k1, $\alpha=1.5$: Reference temperature field and relative $\ell_2$-errors of PBM, DDM and HAM.}
    \label{fig:2K1_profile_a1.5}
\end{figure}

\begin{figure}
    \centering
    \includegraphics[width=\linewidth]{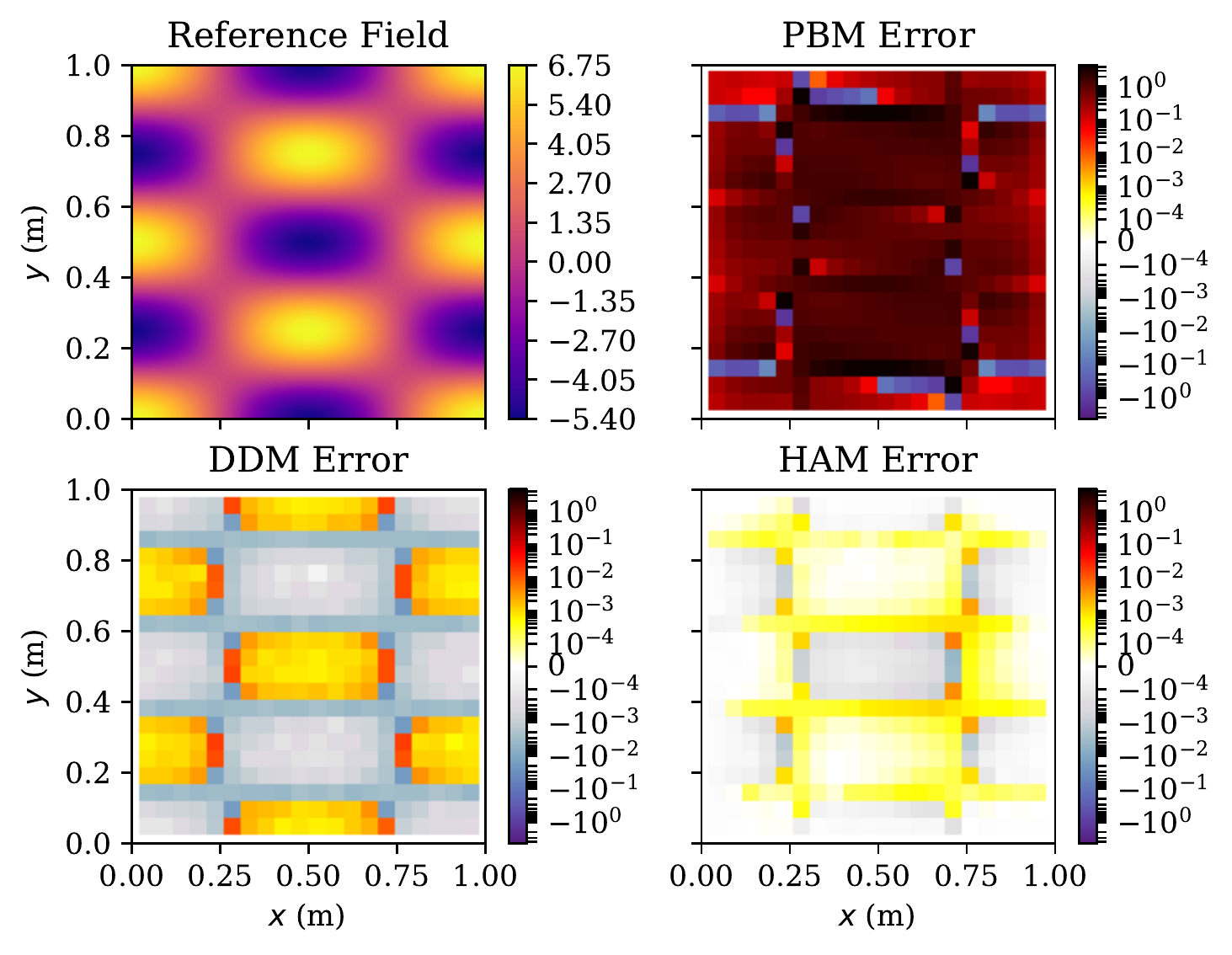}
    \caption{Solution 2k2, $\alpha=0.7$: Reference temperature field and relative $\ell_2$-errors of PBM, DDM and HAM.}
    \label{fig:2K2_profile_a0.7}
\end{figure}

\begin{figure}
    \centering
    \includegraphics[width=\linewidth]{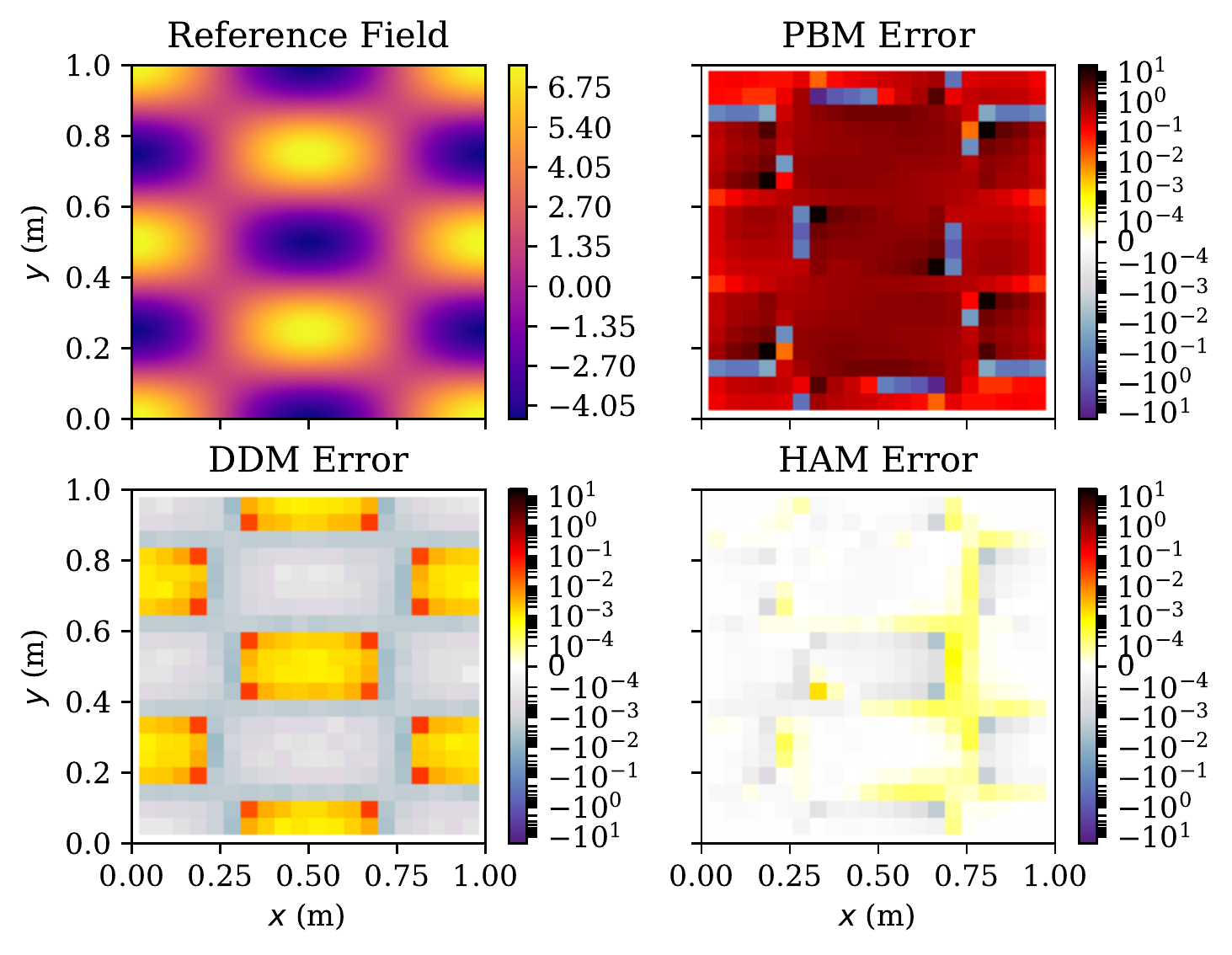}
    \caption{Solution 2k2, $\alpha=1.5$: Reference temperature field and relative $\ell_2$-errors of PBM, DDM and HAM.}
    \label{fig:2K2_profile_a1.5}
\end{figure}

The interpolation scenario results for Solutions~2k1 and~2k2 are shown in Figures~\ref{fig:2k_interp_errors}--\ref{fig:2K2_profile_a1.5}. As in the previous experiments, we see from Figure~\ref{fig:2k_interp_errors} that the CoSTA-based HAM model is the most accurate model. On the whole, CoSTA is at least one order of magnitude more accurate than DDM, which is in turn at least one order of magnitude more accurate than PBM. The difference in accuracy is particularly striking for Solution~2k1, for which the DNN-generated source term of CoSTA yields an accuracy increase of roughly four orders of magnitude in comparison to the uncorrected PBM.

\begin{figure}
	\begin{subfigure}[b]{0.5\linewidth}
		\centering 
		\includegraphics[width=\textwidth]{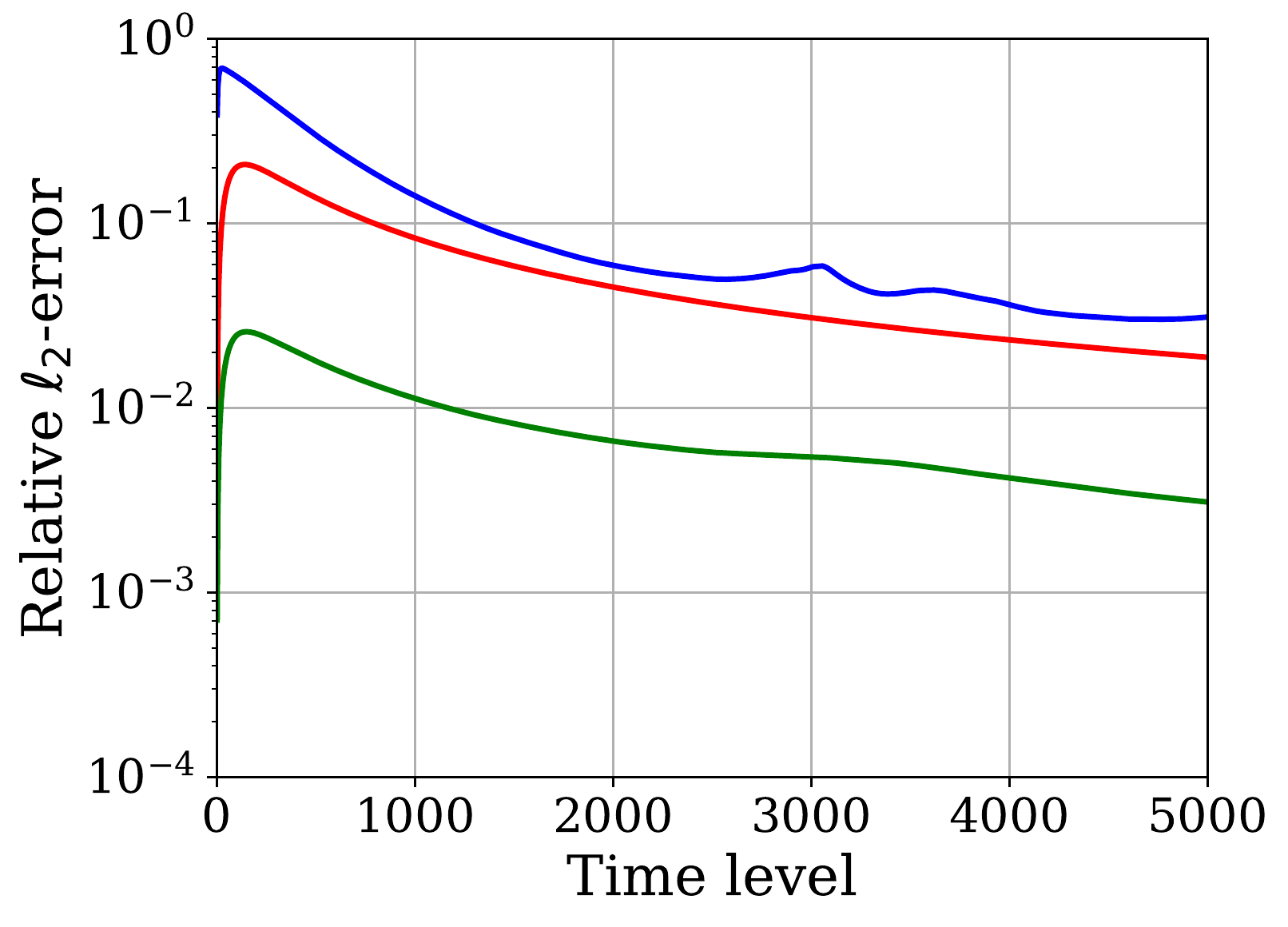}
		\vspace{-1.7em}
		\caption{2k1, $\alpha = -0.5$, relative errors.}
		\vspace{0.4em}
		\label{subfig:2k1_error_a-0.5}
	\end{subfigure}%
	\begin{subfigure}[b]{0.5\linewidth}
		\centering 
		\includegraphics[width=\textwidth]{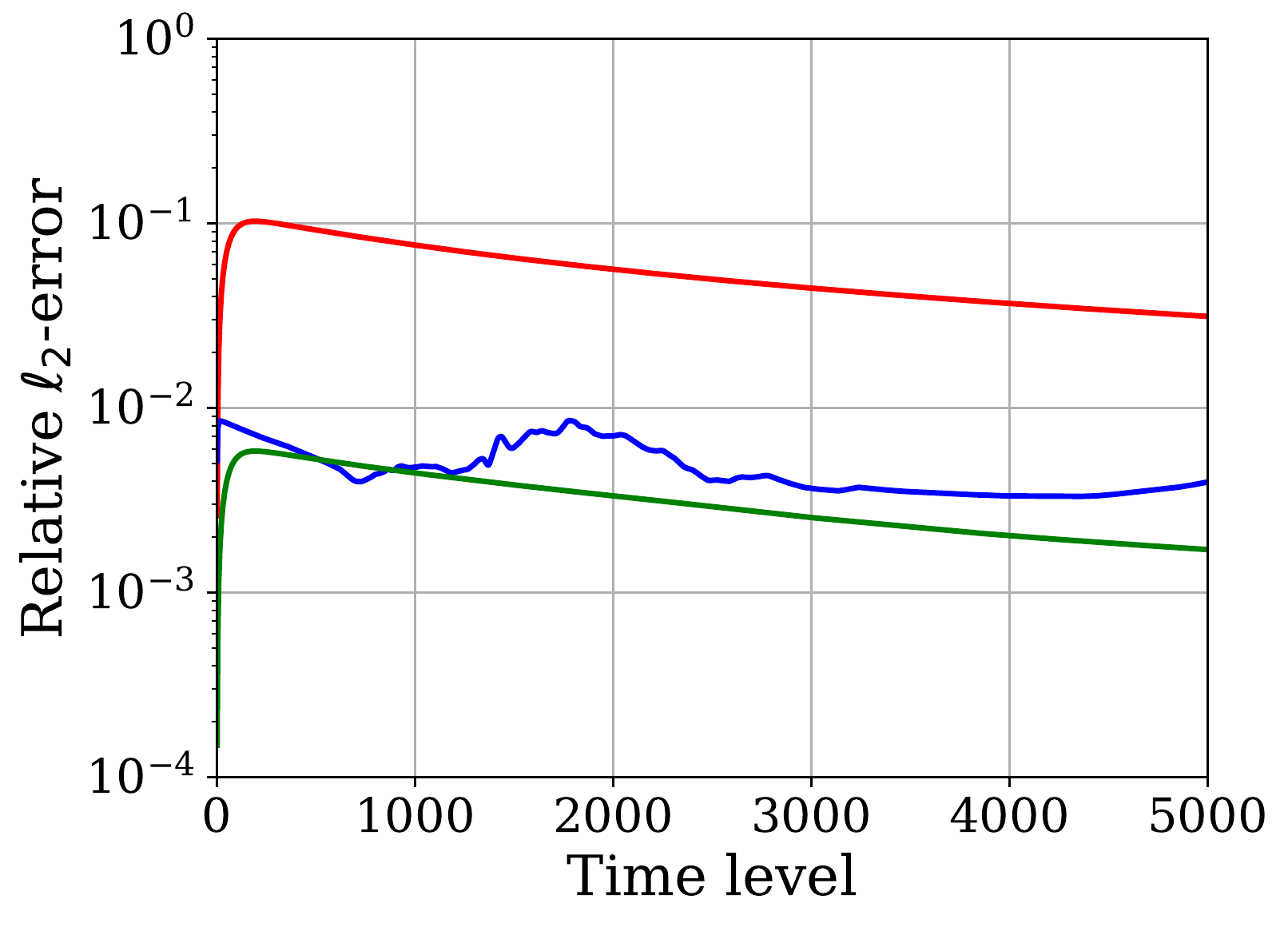}
		\vspace{-1.7em}
		\caption{2k1, $\alpha = 2.5$, relative errors.}
		\vspace{0.4em}
		\label{subfig:2k1_error_a2.5}
	\end{subfigure}%
	\\
	\begin{subfigure}[b]{0.5\linewidth}
		\centering 
		\includegraphics[width=\textwidth]{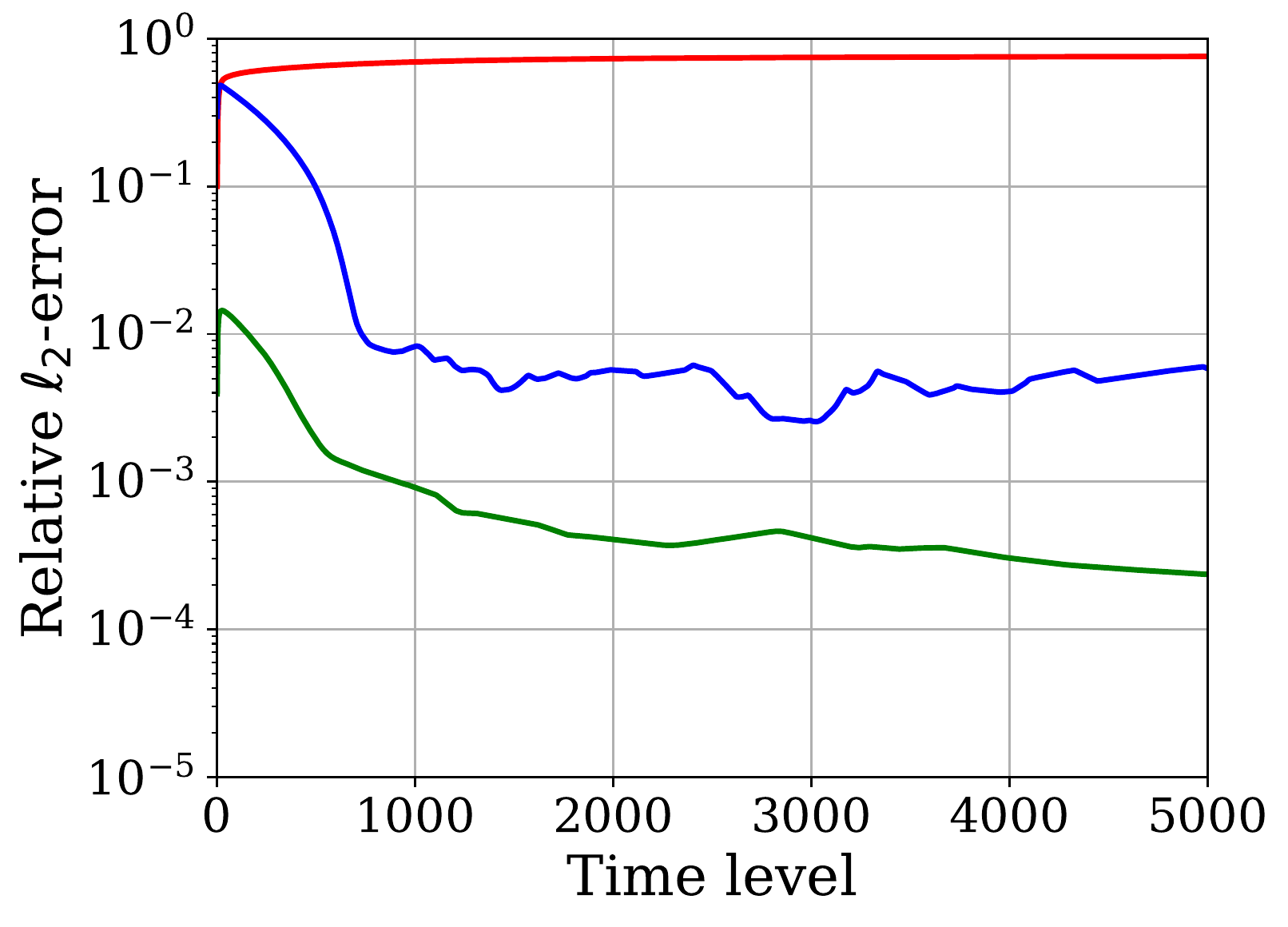}
		\vspace{-1.7em}
		\caption{2k2, $\alpha = -0.5$, relative errors.}
		\label{subfig:2k2_error_a-0.5}
	\end{subfigure}%
	\begin{subfigure}[b]{0.5\linewidth}
		\centering 
		\includegraphics[width=\textwidth]{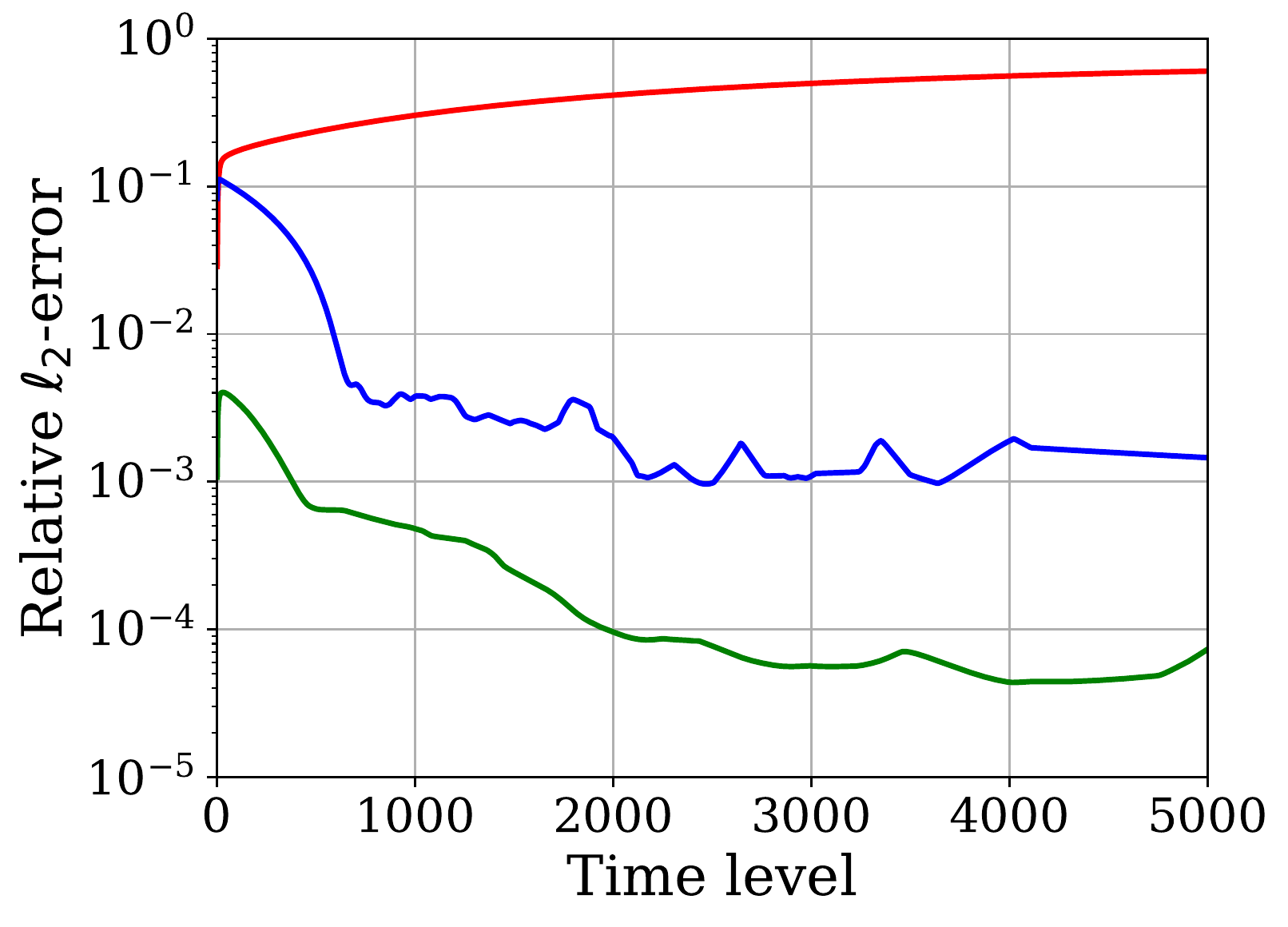}
		\vspace{-1.7em}
		\caption{2k2, $\alpha = 2.5$, relative errors.}
		\label{subfig:2k2_error_a2.5}
	\end{subfigure}%
	\caption{Solutions 2k1 and 2k2, extrapolation: Relative $\ell_2$-errors for $\alpha\in\{0.7,1.5\}$ (\redline~PBM, \blueline~DDM, \greenline~HAM).}
	\label{fig:2k_extrap_errors} 
\end{figure}

\begin{figure}
    \centering
    \includegraphics[width=\linewidth]{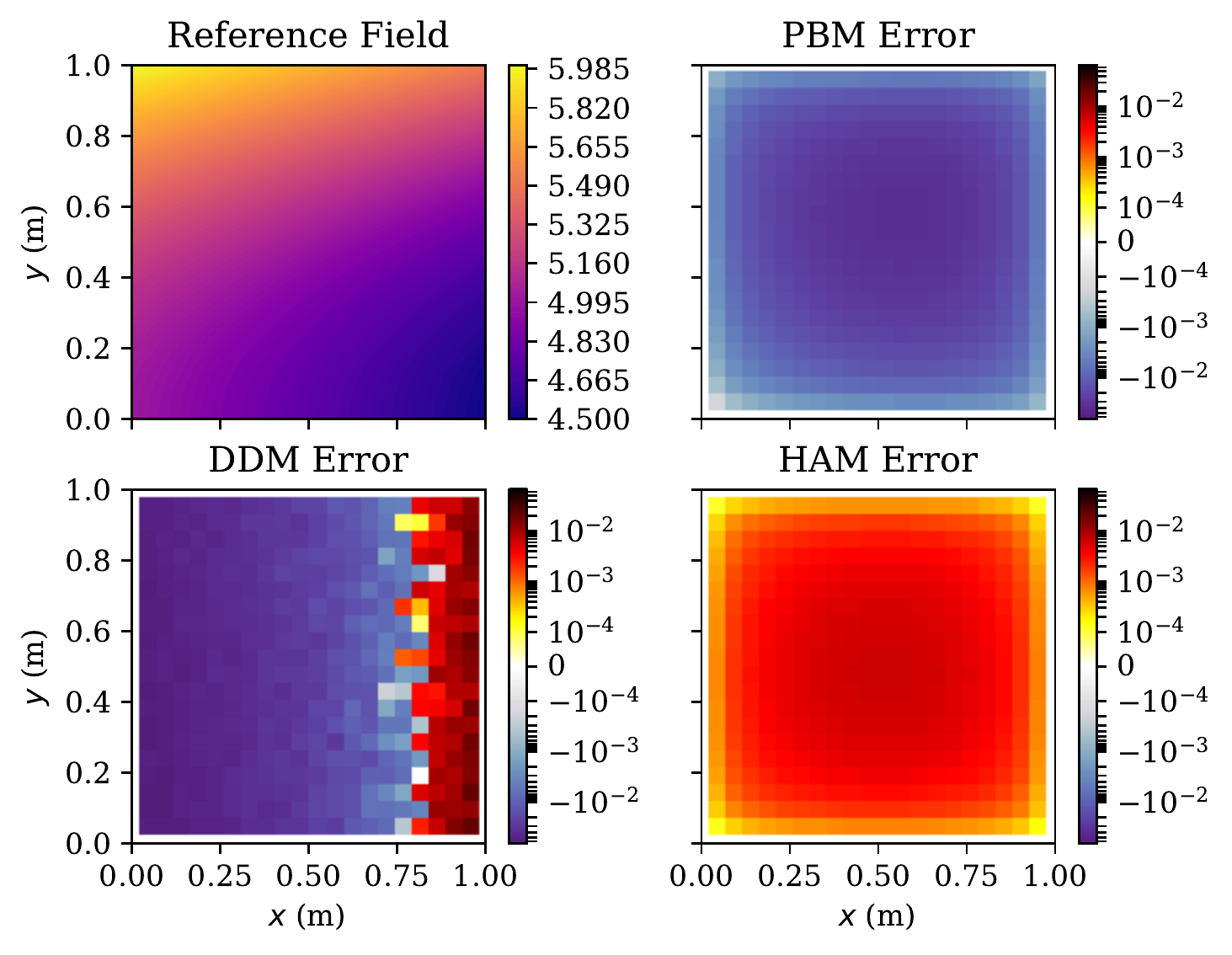}
    \caption{Solution 2k1, $\alpha=-0.5$: Reference temperature field and relative $\ell_2$-errors of PBM, DDM and HAM.}
    \label{fig:2K1_profile_a-0.5}
\end{figure}

\begin{figure}
    \centering
    \includegraphics[width=\linewidth]{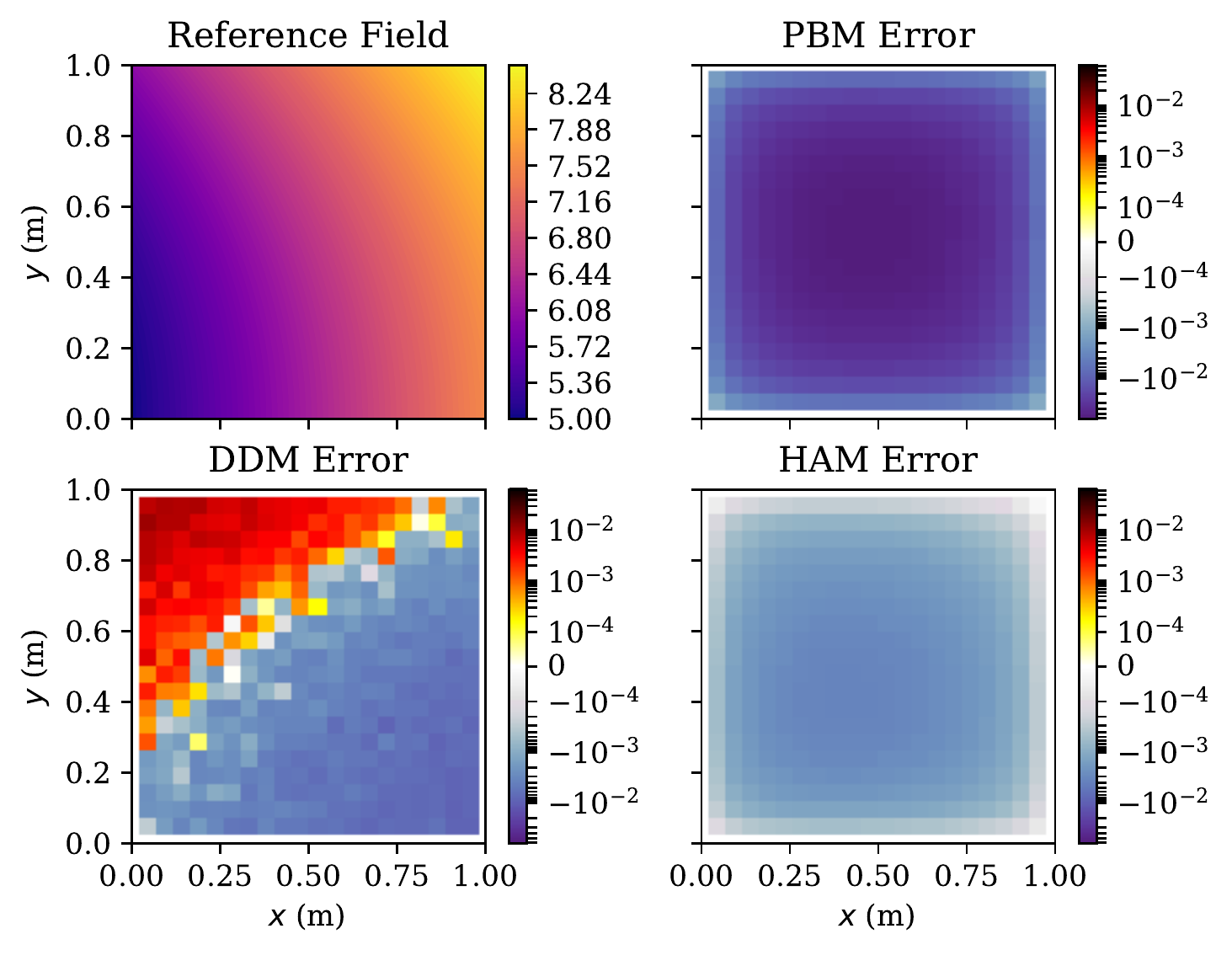}
    \caption{Solution 2k1, $\alpha=2.5$: Reference temperature field and relative $\ell_2$-errors of PBM, DDM and HAM.}
    \label{fig:2K1_profile_a2.5}
\end{figure}

\begin{figure}
    \centering
    \includegraphics[width=\linewidth]{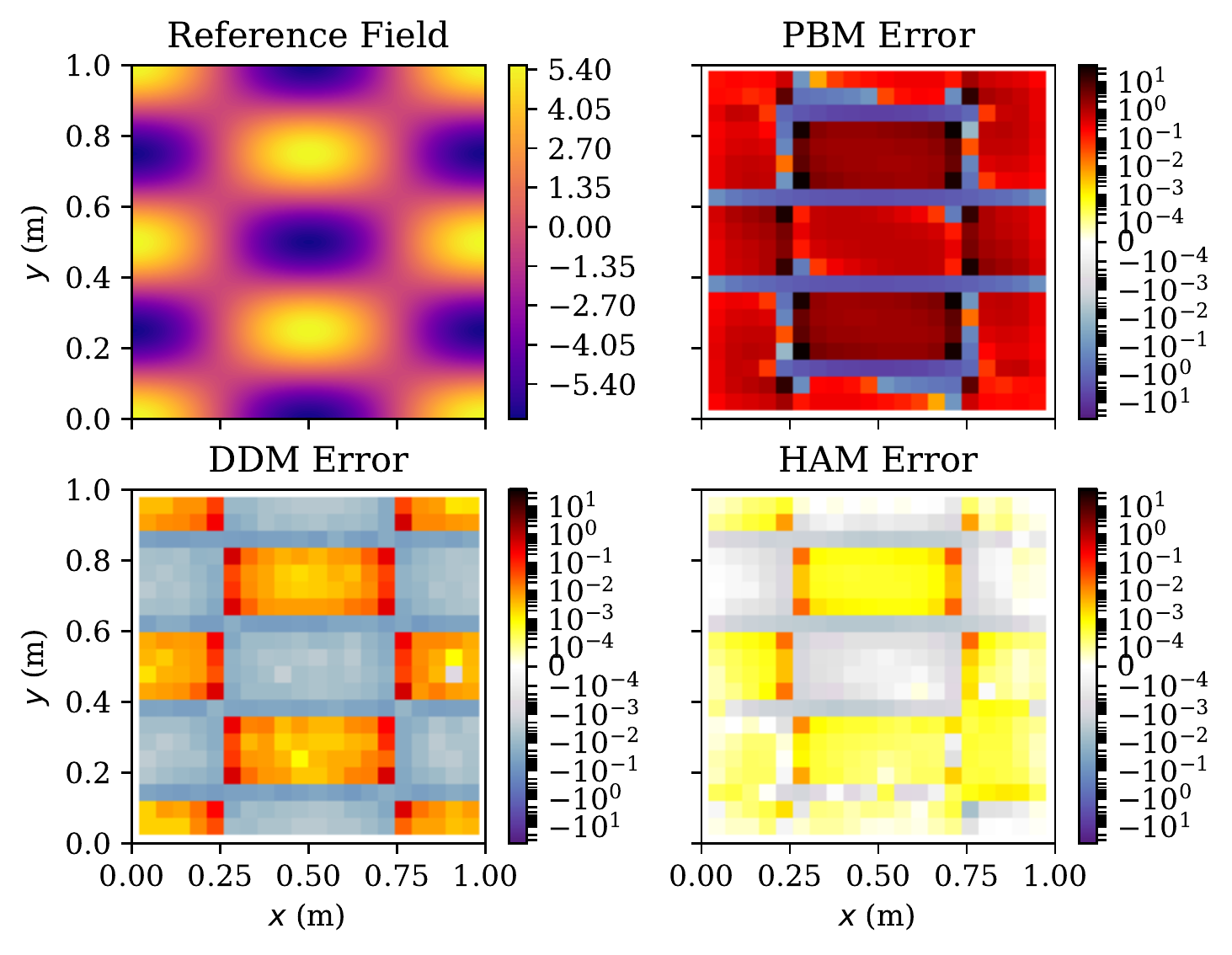}
    \caption{Solution 2k2, $\alpha=-0.5$: Reference temperature field and relative $\ell_2$-errors of PBM, DDM and HAM.}
    \label{fig:2K2_profile_a-0.5}
\end{figure}

\begin{figure}
    \centering
    \includegraphics[width=\linewidth]{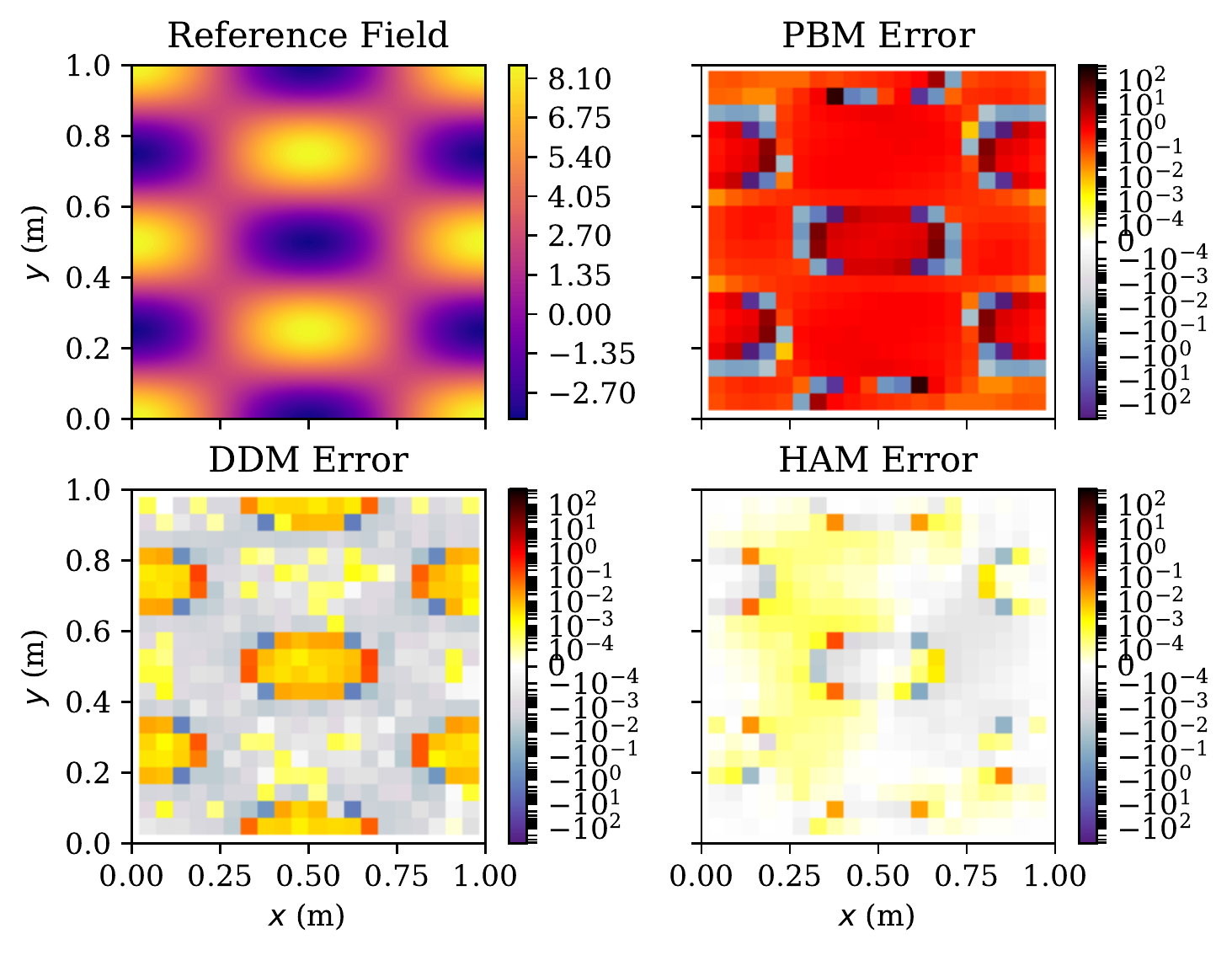}
    \caption{Solution 2k2, $\alpha=2.5$: Reference temperature field and relative $\ell_2$-errors of PBM, DDM and HAM.}
    \label{fig:2K2_profile_a2.5}
\end{figure}

The results for Solutions~2k1 and~2k2 in the extrapolation scenarios $\alpha \in \{-0.5, 2.5\}$ are shown in Figures~\ref{fig:2k_extrap_errors}--\ref{fig:2K2_profile_a2.5}. From the $\ell_2$-errors illustrated in Figure~\ref{fig:2k_extrap_errors}, we see that CoSTA maintains its position as the most accurate model. We also observe that the DDM model apparently generalizes well to the scenario $\alpha=2.5$ for Solution~2k1, almost matching the accuracy of CoSTA in that scenario. However, the DDM model is the least accurate model for the same solution with $\alpha=-0.5$, being more than one order of magnitude less accurate in the latter scenario than in the former. On the other hand, CoSTA maintains the same level of accuracy in both scenarios, thereby exhibiting better generalization than DDM. This conclusion is further strengthened by the error fields shown in Figure~\ref{fig:2K1_profile_a-0.5}. From that figure, we see that the prediction of the CoSTA-based HAM model is qualitatively correct but somewhat too hot over the entire domain. However, the DDM prediction is decidedly too cold over most of the domain while being too warm for $x \gtrsim \SI{0.75}{\meter}$. For the system at hand, there is nothing special occurring at this vertical line, so the error field indicates that the DDM predictions for this scenario are qualitatively incorrect and possibly even unphysical. The errors fields shown in Figure~\ref{fig:2K1_profile_a2.5} tell a similar story; the CoSTA prediction is too cold but otherwise qualitatively correct, while the DDM error field is noisy and has no clear connection to the reference temperature field.

The significant $\ell_2$-error drop observed for System~2k2 in Figure~\ref{fig:2k_extrap_errors} can possibly be explained by noting that, since the oscillation amplitude of $T_{\mathrm{ref}}$ increases with time, the importance of $\alpha$ (which defined the center of the oscillation) decreases with time. As such, the observed errors are consistent with the HAM and DDM models being able to model the spatial oscillation well while having more difficulties modeling  $\alpha$ accurately (with DDM struggling significantly more than HAM).

\FloatBarrier

\subsection{Interpretation of the Corrective Source Term}
\label{subsec:interpretation}

Interpretability of the DNN-generated corrective source term has been highlighted as one of CoSTA's major strong-points. In this section, we aim to substantiate these claims regarding interpretability by demonstrating possible ways of interpreting the corrective source term for Solutions~2P2 and~2k1 with $\alpha=-0.5$ and $\alpha=0.7$. We will also provide some general discussions related to interpretation of the corrective source term. 

Let us first consider the case where we want to model a system with unknown heat generation rate $P$. Suppose we approximate the true $P$ with $\widetilde{P} = P - \epsilon_{P}$, where $\epsilon_P \neq 0$ is the error of the approximation. (For Solutions~2P1 and~2P2 considered in the numerical experiments above, we used $\widetilde{P} = 0$, such that $\epsilon_P = P$.) Inserting $\widetilde{P}+\epsilon_{P}$ for $P$ into Equation~\eqref{eq:general_integral_form} for 2D systems, we obtain
\begin{align}
\begin{split}
    \int\limits_V \rho c_V \diffp{T}{t} \mathrm{d}V
    &= \left( kA \diffp{T}{x} \right)_e - \left( kA \diffp{T}{x} \right)_w + \left( kA \diffp{T}{y} \right)_n - \left( kA \diffp{T}{y} \right)_s \\
    & +  \int\limits_V P\ \mathrm{d}V \\
    &=\left( kA \diffp{T}{x} \right)_e - \left( kA \diffp{T}{x} \right)_w + \left( kA \diffp{T}{y} \right)_n - \left( kA \diffp{T}{y} \right)_s \\ &+ \int\limits_V \widetilde{P}\ \mathrm{d}V + \int\limits_V \epsilon_P\ \mathrm{d}V.
\end{split}
\end{align}
Following the discretization procedure used to derive the Implicit Euler FVM~\eqref{eq:2D_matrix_form},\footnote{This discretization procedure is considered in detail in Section~2.2.3 of \citet{blakseth2021ica}.} we can discretize the above equation as follows:
\begin{align}
    \mathbb{A}\vect{T}^{n+1} = \vect{b} \left( \vect{T}^n \right) + \Delta t \vect{\tilde{\sigma}}_P.
    \label{eq:FVM_with_inc_P}
\end{align}
Here, $\mathbb{A}$, $\vect{T}$ and $\vect{b}$ are defined as in Equation~\eqref{eq:2D_matrix_form}, and $\vect{\tilde{\sigma}}_P = \vect{\epsilon}_P/(\rho c_V)$, where $\vect{\epsilon}_P = [\epsilon_P(x_1,y_1), \dots, \epsilon_P(x_{N_j}, y_{N_i})]$ and $(x_1, y_1), \dots, (x_{N_j}, y_{N_i})$ are the grid nodes used to discretize the spatial domain. Comparing Equation~\eqref{eq:FVM_with_inc_P} to Equation~\eqref{eq:FVM_mod}, we can see that there is a clear connection between $\vect{\tilde{\sigma}}_P$ and the corrective source term $\vect{\hat{\sigma}}$ used in CoSTA.
One might be tempted to simply write $\Delta t \vect{\tilde{\sigma}}_P = \vect{\hat{\sigma}}$, but this equality does not hold true in general. The reason for this is that $\vect{\hat{\sigma}}$ accounts for \emph{all} error in the PBM, while $\vect{\tilde{\sigma}}_P$ accounts only for incorrect modelling of $P$. However, we do have
\begin{equation}
    \Delta t \vect{\tilde{\sigma}}_P \approx \vect{\hat{\sigma}}
    \label{eq:P_approx}
\end{equation}
under the assumption that incorrect modelling of $P$ dominates all other sources of error in the PBM (including discretization error).

\begin{figure}
\centering
	\begin{subfigure}[b]{\linewidth}
		\centering 
		\includegraphics[width=\textwidth]{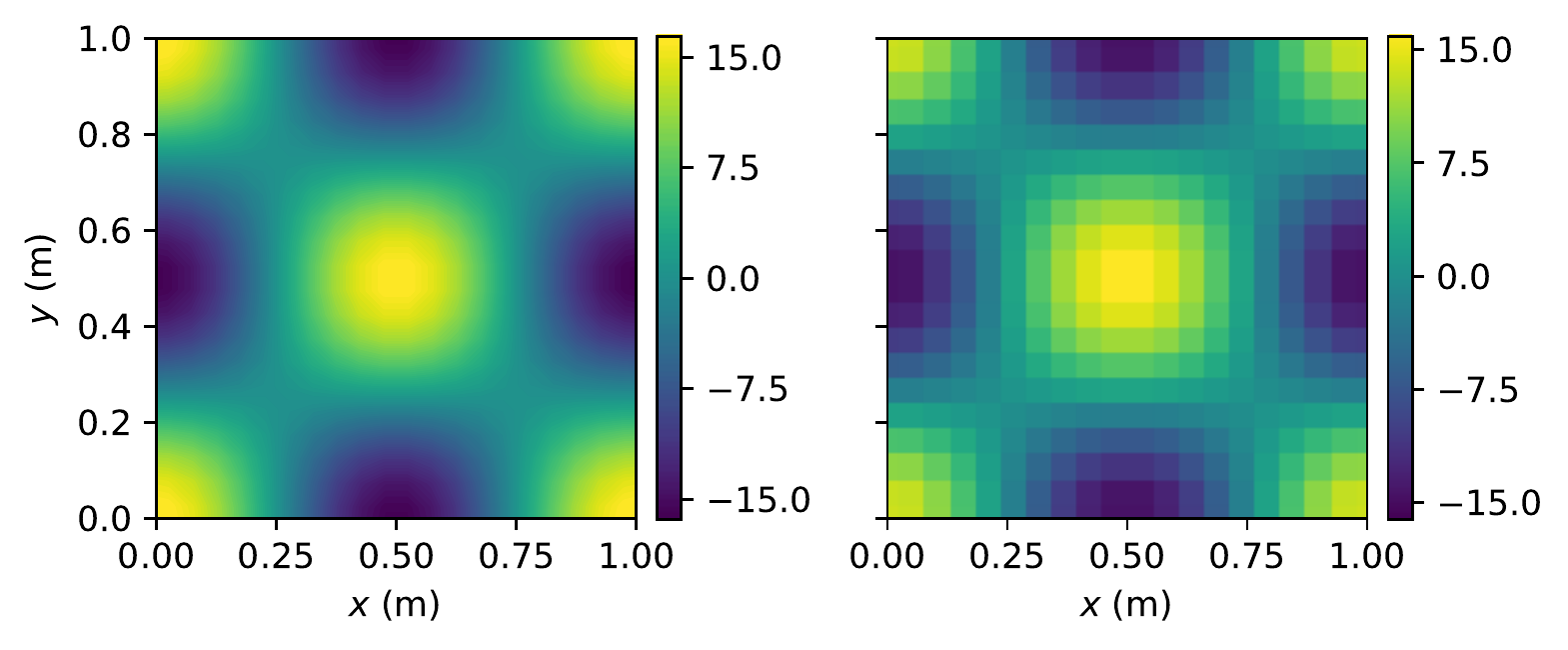}
		\vspace{-1.7em}
		\caption{$\alpha = -0.5$}
		\vspace{0.4em}
		\label{subfig:2P2_interpret_a-0.5}
	\end{subfigure}%
	\\
	\begin{subfigure}[b]{\linewidth}
		\centering 
		\includegraphics[width=\textwidth]{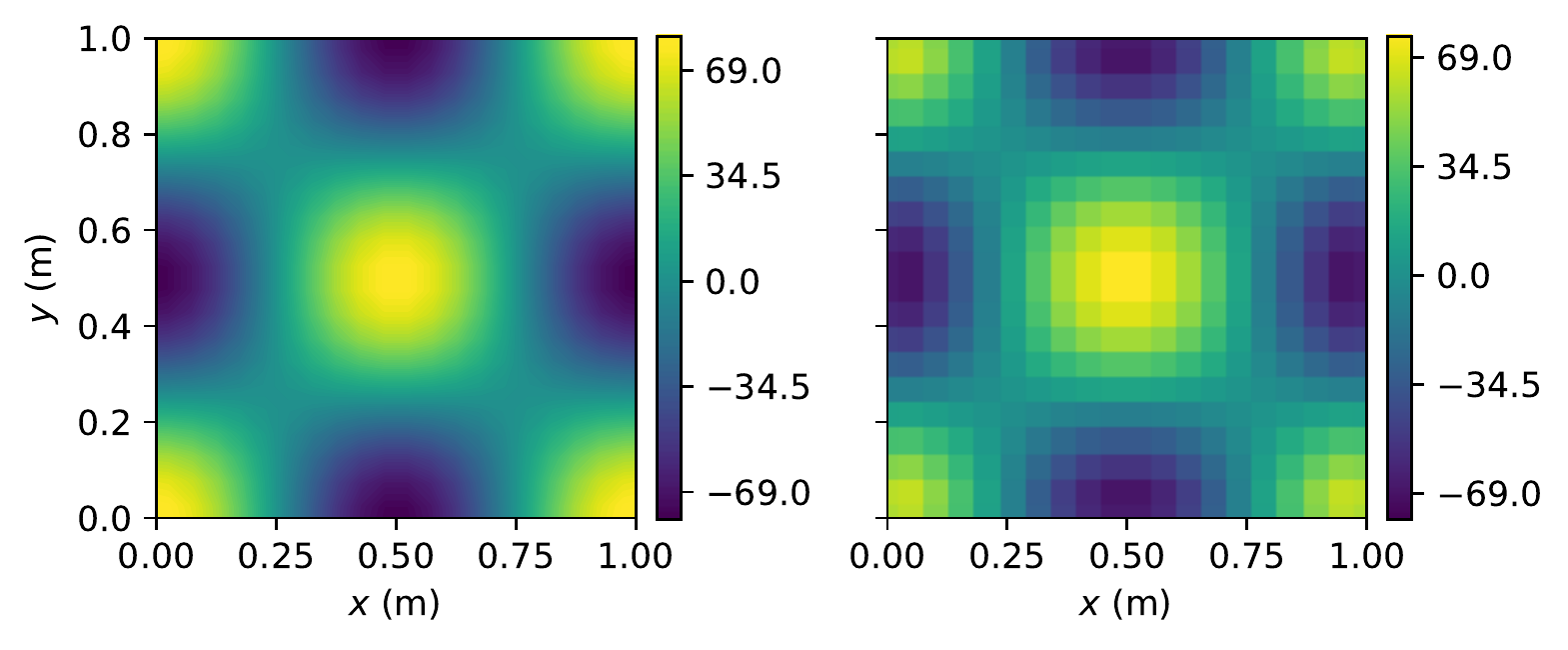}
		\vspace{-1.7em}
		\caption{$\alpha = 0.7$}
		\label{subfig:2P2_interpret_a0.7}
	\end{subfigure}%
	\caption{Comparison of $\vect{\hat{\sigma}}$ (left) and $\Delta t \vect{\hat{\sigma}}_P$ (right) for Solution~2P2 at $t=\SI{0.1}{\second}$.}
	\label{fig:2P2_interpret} 
\end{figure}

In Figure~\ref{fig:2P2_interpret}, we illustrate that the approximation~\eqref{eq:P_approx} is valid for Solution~2P2 with $\alpha \in \{0.7, 1.5\}$. The right-hand side of the figure illustrates the true corrective source term $\vect{\hat{\sigma}}^{n+1}$, as defined in Equation~\eqref{eq:FVM_corr_src_def}, at the time $t=\SI{0.1}{\second}$. The figure's left-hand side illustrates $\Delta t \hat{\sigma}_P = \Delta t \epsilon_P / (\rho c_V)$, which is equal to $\Delta t P$ for the choices of $\widetilde{P}$, $\rho$ and $c_V$ used in the present work. As can be seen from the figure, the top and bottom pairs are visually indistinguishable.\footnote{With the obvious exception that $\vect{\hat{\sigma}}^{n+1}$ is discrete while we have shown $\Delta t \hat{\sigma}_P$ as a continuous field since $P$, and thus also $\hat{\sigma}_P$, is known analytically everywhere.} This indicates that Equation~\eqref{eq:P_approx} is a reasonable approximation when it is known that a PBM suffers from significantly incorrect modelling of $P$.

We now consider the case where we model a system with unknown conductivity $k$. We write $k = \tilde{k} + \epsilon_k$, where $\tilde{k}$ is our estimate of the system's conductivity ($\tilde{k} = 1$ for the numerical experiments considered in the present work), and $\epsilon_k$ is the error of the estimate. Inserting into the 2D version of Equation~\eqref{eq:general_integral_form} yields
\begin{align*}
    \int\limits_V \rho c_V \diffp{T}{t} \mathrm{d}V &= \left( kA \diffp{T}{x} \right)_e - \left( kA \diffp{T}{x} \right)_w \\ &+ \left( kA \diffp{T}{y} \right)_n - \left( kA \diffp{T}{y} \right)_s + \int\limits_V P\, \mathrm{d}V \\
    &= \left( (\tilde{k} + \epsilon_k)A \diffp{T}{x} \right)_e - \left( (\tilde{k} + \epsilon_k)A \diffp{T}{x} \right)_w \\
    & + \left( (\tilde{k} + \epsilon_k)A \diffp{T}{y} \right)_n - \left( (\tilde{k} + \epsilon_k)A \diffp{T}{y} \right)_s + \int\limits_V P\, \mathrm{d}V \\
    &= \left( \tilde{k}A \diffp{T}{x} \right)_e - \left( \tilde{k}A \diffp{T}{x} \right)_w \\
    &+ \left( \tilde{k}A \diffp{T}{y} \right)_n - \left( \tilde{k}A \diffp{T}{y} \right)_s + \int\limits_V P\, \mathrm{d}V \\
    \quad + \left( \epsilon_kA \diffp{T}{x}\right)_e& - \left( \epsilon_kA \diffp{T}{x} \right)_w + \left( \epsilon_kA \diffp{T}{y} \right)_n - \left( \epsilon_kA \diffp{T}{y} \right)_s
\end{align*}

For ease of notation, we define
\begin{equation}
    \xi = \left( \epsilon_kA \diffp{T}{x} \right)_e - \left( \epsilon_kA \diffp{T}{x} \right)_w + \left( \epsilon_kA \diffp{T}{y} \right)_n - \left( \epsilon_kA \diffp{T}{y} \right)_s,
    \label{eq:ksi}
\end{equation}
such that we get
\begin{equation}
\begin{split}
    \int\limits_V \rho c_V \diffp{T}{t} \mathrm{d}V &= \left( \tilde{k}A \diffp{T}{x} \right)_e - \left( \tilde{k}A \diffp{T}{x} \right)_w + \left( \tilde{k}A \diffp{T}{y} \right)_n - \left( \tilde{k}A \diffp{T}{y} \right)_s \\ &+ \int\limits_V P\, \mathrm{d}V + \xi.
\end{split}    
    \label{eq:intermediate_k_interpret}
\end{equation}
To enable a comparison with Equation~\eqref{eq:FVM_corr_src_def}, we need to discretize the equation above. To this end, we discretize the spatial domain into a grid of $N_j \cdot N_i$ grid cells, where integral and half-integral indices are used to denote quantities evaluated at cell centers and cell faces, respectively. As such, $\xi$ for a grid cell centered at $(x_j, y_i)$ is
\begin{align*}
    \xi_{j,i} &= \left( \epsilon_kA \diffp{T}{x} \right)_{j+1/2,i} - \left( \epsilon_kA \diffp{T}{x} \right)_{j-1/2,i} \\ &+
    \left( \epsilon_kA \diffp{T}{y} \right)_{j,i+1/2} - \left( \epsilon_kA \diffp{T}{y} \right)_{j,i-1/2}.
\end{align*}
We now make the following approximations
\begin{alignat*}{2}
    (\epsilon_k)_{j+1/2,i} &\approx (\epsilon_k)_{j,i}, \quad (\epsilon_k)_{j-1/2,i} &&\approx (\epsilon_k)_{j,i}, \\ (\epsilon_k)_{j,i+1/2} &\approx (\epsilon_k)_{j,i}, \quad (\epsilon_k)_{j,i-1/2} &&\approx (\epsilon_k)_{j,i}
\end{alignat*}
\begin{align*}
    \left( \diffp{T}{x} \right)_{j+1/2,i} &\approx \frac{T_{j+1,i} - T_{j,i}}{x_{j+1} - x_j}, \quad \left( \diffp{T}{x} \right)_{j-1/2,i} &&\approx \frac{T_{j,i} - T_{j-1,i}}{x_{j} - x_{j-1}} \\
    \left( \diffp{T}{y} \right)_{j,i+1/2} &\approx \frac{T_{j,i+1} - T_{j,i}}{y_{i+1} - y_i}, \quad \left( \diffp{T}{y} \right)_{j,i-1/2} &&\approx \frac{T_{j,i} - T_{j,i-1}}{y_{i} - y_{i-1}}
\end{align*}
which yield
\begin{equation}
\begin{split}
    \xi_{j,i} = (\epsilon_k A)_{j,i} & \Biggl( \frac{T_{j+1,i} - T_{j,i}}{x_{j+1}-x_j} -  \frac{T_{j,i} - T_{j-1,i}}{x_{j}-x_{j-1}} \\
    &+\frac{T_{j,i+1} - T_{j,i}}{y_{i+1}-y_i} -  \frac{T_{j,i} - T_{j,i-1}}{y_{i}-y_{i-1}} \Biggr).
\end{split}
\end{equation}

Using the above discretization for $\xi$ and the standard Implicit Euler FVM discretization for the other terms of Equation~\eqref{eq:intermediate_k_interpret}, we obtain
\begin{equation}
    \mathbb{A} \vect{T}^{n+1} = \vect{b} ( \vect{T}^n ) + \Delta t \vect{\hat{\sigma}}_k,
\end{equation}
with $\mathbb{A}$ and $\vect{b}$ as in Equation~\eqref{eq:2D_matrix_form} and $\vect{\hat{\sigma}}_k = \vect{\xi} / (\rho c_V)$. Finally, we observe that we have
\begin{equation}
    \Delta t \vect{\hat{\sigma}}_k \approx \vect{\hat{\sigma}}
    \label{eq:k_approx}
\end{equation}
when incorrect modelling of $k$ is the dominant source of error in the PBM.

\begin{figure}
\centering
	\begin{subfigure}[b]{\linewidth}
		\centering 
		\includegraphics[width=\textwidth]{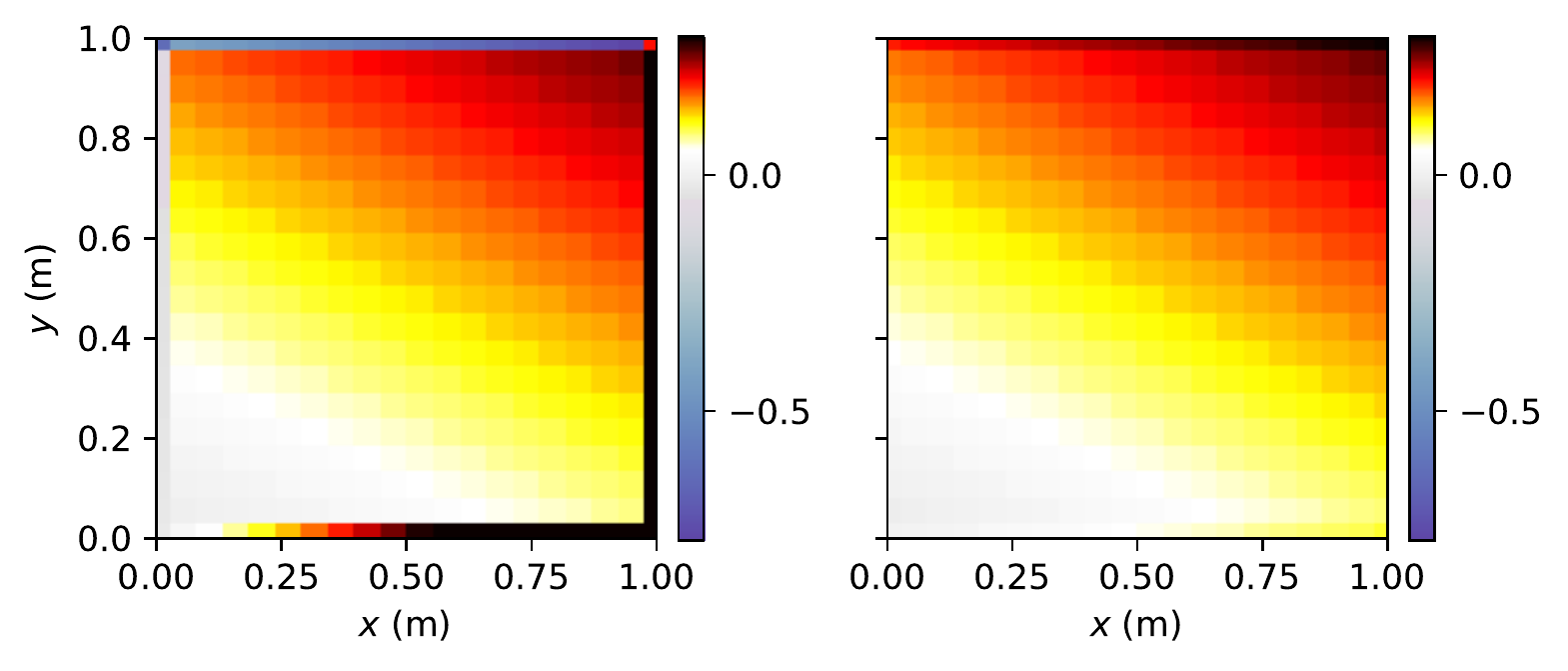}
		\vspace{-1.7em}
		\caption{$\alpha = -0.5$}
		\vspace{0.4em}
		\label{subfig:2k1_interpret_a-0.5}
	\end{subfigure}%
	\\
	\begin{subfigure}[b]{\linewidth}
		\centering 
		\includegraphics[width=\textwidth]{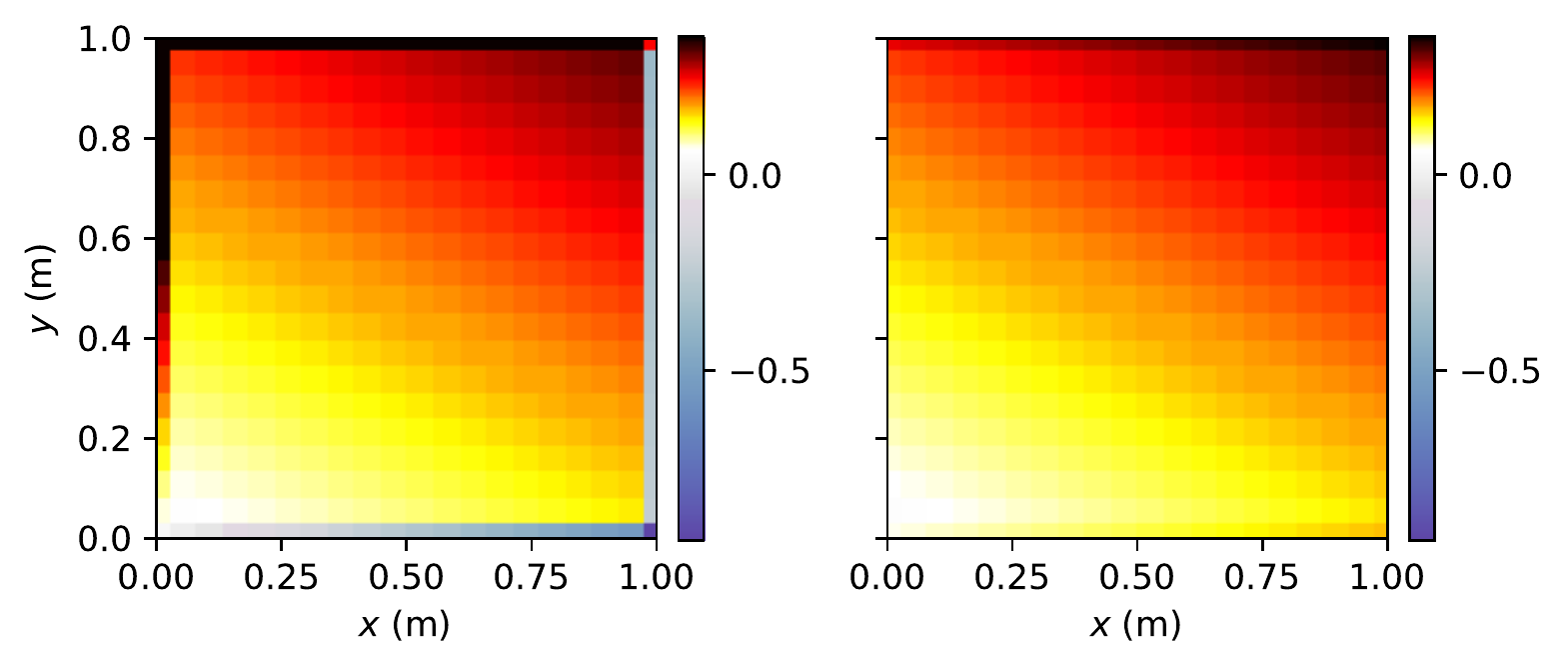}
		\vspace{-1.7em}
		\caption{$\alpha = 0.7$}
		\label{subfig:2k1_interpret_a0.7}
	\end{subfigure}%
	\caption{Comparison of $\vect{\hat{\sigma}}$ (left) and $\Delta t \vect{\hat{\sigma}}_k$ (right) for Solution~2k1 at $t=\SI{0.1}{\second}$.}
	\label{fig:2k1_interpret} 
\end{figure}

The true corrective source term $\vect{\hat{\sigma}}$ is compared to $\Delta t \vect{\hat{\sigma}}_k$ in Figure~\ref{fig:2k1_interpret} for Solution~2k1 with $\alpha=-0.5$ and $\alpha = 0.7$ at time $t=\SI{0.1}{\second}$. From the figure, it is clear that the approximation~\eqref{eq:k_approx} holds well in the interior of the domain. However, at the domain boundaries, comparatively large discrepancies are visible. There are two main contributions to these discrepancies. The first is the approximations made when discretizing $\xi$, and the other is the influence of the discretization error in the original PBM on $\vect{\hat{\sigma}}$. Irrespective of where the discrepancies originate from, their presence illustrates that it is generally advisable to perform interpretations of the corrective source term in the interior of the domain, such as to avoid the influence of boundary effects. 

Equations~\eqref{eq:P_approx} and~\eqref{eq:k_approx} have two main areas of application: model improvement and DNN sanity checks. We first consider model improvement, which can be performed by conducting some kind of regression analysis on the true corrective source term $\vect{\hat{\sigma}}$ to approximate it with some analytic expression. For example, the symbolic regression techniques studied by \cite{vaddireddy2020fes} are prime candidate for performing this kind of regression analysis. Once a regressed expression for $\vect{\hat{\sigma}}$ has been identified, this expression can be related to the modelling errors $\epsilon_P$ and $\epsilon_k$ using the applicable approximation~\eqref{eq:P_approx} or~\eqref{eq:k_approx}. The modelling of $P$ or $k$ in the PBM-component of the CoSTA model can then be updated according to estimated modelling error. Since CoSTA benefits from improved PBM accuracy, this would increase the accuracy of the CoSTA model.\footnote{The regression analysis and PBM update can be performed before the DNN of the CoSTA model is trained, such that the proposed improvement scheme does not incur any extra DNN training cost.}

For sanity checks, we observe first that a successfully trained DNN$_{\sigma}$ is -- by definition -- a DNN$_{\sigma}$ for which we have $\vect{\hat{\sigma}}_{\textsc{nn}} \approx \vect{\hat{\sigma}}$. So let us insert $\vect{\hat{\sigma}}_{\textsc{nn}}$ into Equation~\eqref{eq:P_approx} or~\eqref{eq:k_approx} (depending on which is applicable). For systems whose PBM is dominated by a single error source, we expect the applicable approximation to hold well even with this modification. This allows us to obtain a relation between $\vect{\hat{\sigma}}_{\textsc{nn}}$ and $\epsilon_P$ or $\epsilon_k$.
Of course, for \emph{a priori} predictions, the modelling error $\epsilon_P$ or $\epsilon_k$ is not known exactly, but it may be known that they are bounded within some range.
As a simple example, suppose the we know that a room is heated by some heater whose precise power is unknown. Moreover, suppose that the heater manufacturer specifies that their heaters output $\SI{50}{\watt\per\meter^3} \pm 10\%$. When modelling the effect of the heater on the room temperature, it is then natural to use $\widetilde{P} = \SI{50}{\watt\per\meter^3}$. Then, $\epsilon_P$ should lie within the range $[-\SI{5}{\watt\per\meter^3}, \SI{5}{\watt\per\meter^3}]$. Equation~\eqref{eq:P_approx} then gives us a corresponding bound for $\vect{\hat{\sigma}}_{\textsc{nn}}$. This is generally not a hard bound, since $\vect{\hat{\sigma}}_{\textsc{nn}}$ is also influenced by any other errors in the PBM, but it gives a ball park estimate for $\vect{\hat{\sigma}}_{\textsc{nn}}$. If we suddenly observe that DNN$_{\sigma}$ produces a $\vect{\hat{\sigma}}_{\textsc{nn}}$ well outside the estimated ball park (e.g. a $\vect{\hat{\sigma}}_{\textsc{nn}}$ corresponding to negative $\epsilon_k$, in our example), this is a clear sign of DNN misbehaviour. As such, physic-based interpretation of $\vect{\hat{\sigma}}_{\textsc{nn}}$ using the framework presented in this section provides a valuable sanity-check for the DNN of the CoSTA model.

We conclude the this section with two remarks on the applicability of the interpretability approach outlined above. First of all, if two or more error source contribute significantly to the overall error of a PBM, the approximations underpinning the analyses above do not hold. Also, it is not straightforward to separate the contributions from the different error source. Interpretation of the corrective source term using the suggested approach will then be limited. However, this is a general problem with inverse solution methods, so the issue is not unique to CoSTA. Secondly, we highlight that, when a single error source is dominating, the approach above is not limited to capturing the effects of heat generation or conductivity in the heat equation. To the contrary, analogous calculations can equally well be carried out for other governing equations with other parameters.

\subsection{The Impact of Noise}

Just like other modeling approaches utilizing data-driven techniques, CoSTA is potentially vulnerable to noise. Indeed, if the reference data ($\vect{T}_{\mathrm{ref}}$ in our case) contains noise, this noise will be embedded in the corrective source term $\hat{\sigma}$ from Equation~\eqref{eq:PBM_mod_in}. It is therefore beneficial to remove as much noise as possible from the reference data before it is used to train the DNN that is part of the CoSTA model. In particular, it is advised to minimize high-frequency noise, since this noise will have the greatest impact on the magnitude of the corrective source term. If the nature of noise is statistically similar across the snapshots in time, conducting a principal component analysis for denoising the data could be helpful. Still, it is not necessarily so that all noise must be removed for CoSTA to be usable. If the noise is truly random, proper regularization of the DNN should ensure that $\hat{\sigma}_{\textsc{nn}}$ is not detrimentally affected by the noise. Using the DNN in a CoSTA model does not restrict the choice of regularization techniques, so standard regularization techniques like dropout, weight regularization and early stopping are all applicable.

\section{Conclusion}
\label{sec:conclusionandfuturework}
In this work, we presented the Corrective Source Term Approach (CoSTA) to Hybrid Analysis and Modeling (HAM). The method exploits the universal approximation properties of a deep neural network (DNN) to generate a correction term that compensates for the unmodeled/unresolved physics in a physics-based model (PBM). In a series of numerical experiments on two-dimensional heat diffusion problems, we compared the performance of the CoSTA-based HAM to PBM and the data-driven model (DDM). The two major conclusions from the study are as follows:

\begin{itemize}
    \item In terms of predictive accuracy, CoSTA for two dimensional heat diffusion problems involving unknown physics is several orders of magnitude more accurate than comparable PBM and DDM both for interpolation as well as extrapolation cases. 
    \item It is also demonstrated that the CoSTA-generated corrective source term can be subjected to physical interpretation leading to a better understanding of the underlying physics. In fact, physical laws (like the conservation of energy) can be used to put a sanity check on the predictions of the DDM-part of CoSTA models. Such sanity checks are foreseen to result in more reliable models, resulting in increased penetration of DDM in high-stakes applications.
\end{itemize}

Despite the demonstrated strengths of CoSTA, there are still some areas for improvement. For example, the DDM used does not consider all known information of the scenario in which the model is applied. More specifically, we do not inform the DDM of the parameter $\alpha$, even if it might be reasonable to assume that $\alpha$ is known. One of the reasons for not exploiting the knowledge about the parameter's value is that the vanilla neural network architecture used in this work downweights the importance of the parameter if it is fed at the input layer. Based on our recent research work in \cite{haakon2022pgnn,pawar2021pgml, pawar2021msw}, it could be desirable to inject this knowledge in an intermediate hidden layer leading to smoother and more certain solutions. Another knowledge that is not yet exploited is the temporal correlation of the time series. This can be addressed through the use of long short-term memory (LSTM) network. A potential future extension of the work is proposed in these two directions. Lastly, as explained earlier, we chose to demonstrate the effectiveness of the proposed approach using synthetic data which was devoid of any noise. Since we claim that the approach can be used in the context of predictive digital twins, it would be valuable to apply the approach on real temperature data collected using e.g.\ high-resolution thermal cameras.

\section*{Acknowledgments}
The second and third author are grateful for the support received by the Research Council of Norway and the industrial partners of the following projects: EXAIGON--{\em Explainable AI systems for gradual industry adoption\/} (grant no. 304843), {\em Hole cleaning monitoring in drilling with distributed sensors and hybrid methods\/} (grant no. 308823), and RaPiD--{\em Reciprocal Physics and Data-driven models\/} (grant no. 313909). The fourth author gratefully acknowledges the Early Career Research Program (ECRP) support of the U.S. Department of Energy, Office of Science, Office of Advanced Scientific Computing Research under Award Number DE-SC0019290.

%\section{Bibliography styles}
%\section*{References}
\bibliographystyle{elsarticle-harv}
\bibliography{references}

\begin{thebibliography}{22}
\expandafter\ifx\csname natexlab\endcsname\relax\def\natexlab#1{#1}\fi
\providecommand{\url}[1]{\texttt{#1}}
\providecommand{\href}[2]{#2}
\providecommand{\path}[1]{#1}
\providecommand{\DOIprefix}{doi:}
\providecommand{\ArXivprefix}{arXiv:}
\providecommand{\URLprefix}{URL: }
\providecommand{\Pubmedprefix}{pmid:}
\providecommand{\doi}[1]{\href{http://dx.doi.org/#1}{\path{#1}}}
\providecommand{\Pubmed}[1]{\href{pmid:#1}{\path{#1}}}
\providecommand{\bibinfo}[2]{#2}
\ifx\xfnm\relax \def\xfnm[#1]{\unskip,\space#1}\fi
%Type = Article
\bibitem[{Ahmed et~al.(2021)Ahmed, Pawar, San, Rasheed, Iliescu and
  Noack}]{shady2021ocf}
\bibinfo{author}{Ahmed, S.E.}, \bibinfo{author}{Pawar, S.},
  \bibinfo{author}{San, O.}, \bibinfo{author}{Rasheed, A.},
  \bibinfo{author}{Iliescu, T.}, \bibinfo{author}{Noack, B.R.},
  \bibinfo{year}{2021}.
\newblock \bibinfo{title}{On closures for reduced order models—a spectrum of
  first-principle to machine-learned avenues}.
\newblock \bibinfo{journal}{Physics of Fluids} \bibinfo{volume}{33},
  \bibinfo{pages}{091301}.
\newblock \URLprefix \url{https://doi.org/10.1063/5.0061577},
  \DOIprefix\doi{10.1063/5.0061577},
  \href{http://arxiv.org/abs/https://doi.org/10.1063/5.0061577}{{\tt
  arXiv:https://doi.org/10.1063/5.0061577}}.
%Type = Inproceedings
\bibitem[{Amos and Kolter(2017)}]{amos_2017_optnet}
\bibinfo{author}{Amos, B.}, \bibinfo{author}{Kolter, J.Z.},
  \bibinfo{year}{2017}.
\newblock \bibinfo{title}{{OptNet}: {Differentiable} {Optimization} as a
  {Layer} in {Neural} {Networks}}, in: \bibinfo{booktitle}{International
  {Conference} on {Machine} {Learning}}, \bibinfo{publisher}{PMLR}. pp.
  \bibinfo{pages}{136--145}.
\newblock \URLprefix \url{https://proceedings.mlr.press/v70/amos17a.html}.
%Type = Inproceedings
\bibitem[{de~Avila Belbute-Peres et~al.(2018)de~Avila Belbute-Peres, Smith,
  Allen, Tenenbaum and Kolter}]{avilabelbuteperes_2018_end}
\bibinfo{author}{de~Avila Belbute-Peres, F.}, \bibinfo{author}{Smith, K.},
  \bibinfo{author}{Allen, K.}, \bibinfo{author}{Tenenbaum, J.},
  \bibinfo{author}{Kolter, J.Z.}, \bibinfo{year}{2018}.
\newblock \bibinfo{title}{End-to-{End} {Differentiable} {Physics} for
  {Learning} and {Control}}, in: \bibinfo{booktitle}{Advances in {Neural}
  {Information} {Processing} {Systems}}, \bibinfo{publisher}{Curran Associates,
  Inc.}
\newblock \URLprefix
  \url{https://papers.nips.cc/paper/2018/hash/842424a1d0595b76ec4fa03c46e8d755-Abstract.html}.
%Type = Article
\bibitem[{Bakarji and Tartakovsky(2021)}]{BAKARJI2021110219}
\bibinfo{author}{Bakarji, J.}, \bibinfo{author}{Tartakovsky, D.M.},
  \bibinfo{year}{2021}.
\newblock \bibinfo{title}{Data-driven discovery of coarse-grained equations}.
\newblock \bibinfo{journal}{Journal of Computational Physics}
  \bibinfo{volume}{434}, \bibinfo{pages}{110219}.
\newblock \URLprefix
  \url{https://www.sciencedirect.com/science/article/pii/S0021999121001145},
  \DOIprefix\doi{https://doi.org/10.1016/j.jcp.2021.110219}.
%Type = Masterthesis
\bibitem[{Blakseth(2021)}]{blakseth2021ica}
\bibinfo{author}{Blakseth, S.S.}, \bibinfo{year}{2021}.
\newblock \bibinfo{title}{Introducing {CoSTA}: A Deep Neural Network Enabled
  Approach to Improving Physics-Based Numerical Simulations}.
\newblock Master's thesis. NTNU.
%Type = Article
\bibitem[{Blakseth et~al.(2022)Blakseth, Rasheed, Kvamsdal and
  San}]{blakseth2022dnn}
\bibinfo{author}{Blakseth, S.S.}, \bibinfo{author}{Rasheed, A.},
  \bibinfo{author}{Kvamsdal, T.}, \bibinfo{author}{San, O.},
  \bibinfo{year}{2022}.
\newblock \bibinfo{title}{Deep neural network enabled corrective source term
  approach to hybrid analysis and modeling}.
\newblock \bibinfo{journal}{Neural Networks} \bibinfo{volume}{146},
  \bibinfo{pages}{181--199}.
\newblock \URLprefix
  \url{https://www.sciencedirect.com/science/article/pii/S0893608021004494},
  \DOIprefix\doi{https://doi.org/10.1016/j.neunet.2021.11.021}.
%Type = Article
\bibitem[{Champion et~al.(2019)Champion, Lusch, Kutz and
  Brunton}]{Champion22445}
\bibinfo{author}{Champion, K.}, \bibinfo{author}{Lusch, B.},
  \bibinfo{author}{Kutz, J.N.}, \bibinfo{author}{Brunton, S.L.},
  \bibinfo{year}{2019}.
\newblock \bibinfo{title}{Data-driven discovery of coordinates and governing
  equations}.
\newblock \bibinfo{journal}{Proceedings of the National Academy of Sciences}
  \bibinfo{volume}{116}, \bibinfo{pages}{22445--22451}.
\newblock \URLprefix \url{https://www.pnas.org/content/116/45/22445},
  \DOIprefix\doi{10.1073/pnas.1906995116},
  \href{http://arxiv.org/abs/https://www.pnas.org/content/116/45/22445.full.pdf}{{\tt
  arXiv:https://www.pnas.org/content/116/45/22445.full.pdf}}.
%Type = Article
\bibitem[{Georgaka et~al.(2020)Georgaka, Stabile, Star, Rozza and
  Bluck}]{GEORGAKA2020104615}
\bibinfo{author}{Georgaka, S.}, \bibinfo{author}{Stabile, G.},
  \bibinfo{author}{Star, K.}, \bibinfo{author}{Rozza, G.},
  \bibinfo{author}{Bluck, M.J.}, \bibinfo{year}{2020}.
\newblock \bibinfo{title}{A hybrid reduced order method for modelling turbulent
  heat transfer problems}.
\newblock \bibinfo{journal}{Computers \& Fluids} \bibinfo{volume}{208},
  \bibinfo{pages}{104615}.
\newblock \URLprefix
  \url{https://www.sciencedirect.com/science/article/pii/S0045793020301870},
  \DOIprefix\doi{https://doi.org/10.1016/j.compfluid.2020.104615}.
%Type = Article
\bibitem[{He et~al.(2021)He, Ni, Wang and Zhang}]{HE2021102719}
\bibinfo{author}{He, Z.}, \bibinfo{author}{Ni, F.}, \bibinfo{author}{Wang, W.},
  \bibinfo{author}{Zhang, J.}, \bibinfo{year}{2021}.
\newblock \bibinfo{title}{A physics-informed deep learning method for solving
  direct and inverse heat conduction problems of materials}.
\newblock \bibinfo{journal}{Materials Today Communications}
  \bibinfo{volume}{28}, \bibinfo{pages}{102719}.
\newblock \URLprefix
  \url{https://www.sciencedirect.com/science/article/pii/S235249282100711X},
  \DOIprefix\doi{https://doi.org/10.1016/j.mtcomm.2021.102719}.
%Type = Book
\bibitem[{{LeVeque}(2002)}]{leveque2002fvm}
\bibinfo{author}{{LeVeque}, R.J.}, \bibinfo{year}{2002}.
\newblock \bibinfo{title}{Finite-Volume Methods for Hyperbolic Problems}.
\newblock \bibinfo{edition}{1st} ed., \bibinfo{publisher}{Cambridge University
  Press}.
%Type = Article
\bibitem[{Li et~al.(2020)Li, Gao, Han, Yang and Yu}]{LI2020118783}
\bibinfo{author}{Li, T.}, \bibinfo{author}{Gao, Y.}, \bibinfo{author}{Han, D.},
  \bibinfo{author}{Yang, F.}, \bibinfo{author}{Yu, B.}, \bibinfo{year}{2020}.
\newblock \bibinfo{title}{A novel pod reduced-order model based on edfm for
  steady-state and transient heat transfer in fractured geothermal reservoir}.
\newblock \bibinfo{journal}{International Journal of Heat and Mass Transfer}
  \bibinfo{volume}{146}, \bibinfo{pages}{118783}.
\newblock \URLprefix
  \url{https://www.sciencedirect.com/science/article/pii/S0017931019325219},
  \DOIprefix\doi{https://doi.org/10.1016/j.ijheatmasstransfer.2019.118783}.
%Type = Article
\bibitem[{Pawar et~al.(2021a)Pawar, San, Aksoylu, Rasheed and
  Kvamsdal}]{pawar2021pgml}
\bibinfo{author}{Pawar, S.}, \bibinfo{author}{San, O.},
  \bibinfo{author}{Aksoylu, B.}, \bibinfo{author}{Rasheed, A.},
  \bibinfo{author}{Kvamsdal, T.}, \bibinfo{year}{2021}a.
\newblock \bibinfo{title}{Physics guided machine learning using simplified
  theories}.
\newblock \bibinfo{journal}{Physics of Fluids} \bibinfo{volume}{33},
  \bibinfo{pages}{011701}.
%Type = Article
\bibitem[{Pawar et~al.(2021b)Pawar, San, N., Rasheed and
  Kvamsdal}]{pawar2021msw}
\bibinfo{author}{Pawar, S.}, \bibinfo{author}{San, O.}, \bibinfo{author}{N.,
  A.}, \bibinfo{author}{Rasheed, A.}, \bibinfo{author}{Kvamsdal, T.},
  \bibinfo{year}{2021}b.
\newblock \bibinfo{title}{Model fusion with physics-guided machine learning:
  projection based reduced order modeling.}
\newblock \bibinfo{journal}{Physics of Fluids} \bibinfo{volume}{33},
  \bibinfo{pages}{067123}.
%Type = Article
\bibitem[{Penwarden et~al.(2021)Penwarden, Zhe, Narayan and
  Kirby}]{PENWARDEN2021110844}
\bibinfo{author}{Penwarden, M.}, \bibinfo{author}{Zhe, S.},
  \bibinfo{author}{Narayan, A.}, \bibinfo{author}{Kirby, R.M.},
  \bibinfo{year}{2021}.
\newblock \bibinfo{title}{Multifidelity modeling for physics-informed neural
  networks (pinns)}.
\newblock \bibinfo{journal}{Journal of Computational Physics} ,
  \bibinfo{pages}{110844}\URLprefix
  \url{https://www.sciencedirect.com/science/article/pii/S0021999121007397},
  \DOIprefix\doi{https://doi.org/10.1016/j.jcp.2021.110844}.
%Type = Book
\bibitem[{Quarteroni and Rozza(2014)}]{quarteroni2014reduced}
\bibinfo{author}{Quarteroni, A.}, \bibinfo{author}{Rozza, G.},
  \bibinfo{year}{2014}.
\newblock \bibinfo{title}{Reduced order methods for modeling and computational
  reduction}. volume~\bibinfo{volume}{9}.
\newblock \bibinfo{publisher}{Springer, New York}.
%Type = Article
\bibitem[{Raissi et~al.(2019)Raissi, Perdikaris and
  Karniadakis}]{raissi2019physics}
\bibinfo{author}{Raissi, M.}, \bibinfo{author}{Perdikaris, P.},
  \bibinfo{author}{Karniadakis, G.E.}, \bibinfo{year}{2019}.
\newblock \bibinfo{title}{Physics-informed neural networks: A deep learning
  framework for solving forward and inverse problems involving nonlinear
  partial differential equations}.
\newblock \bibinfo{journal}{Journal of Computational Physics}
  \bibinfo{volume}{378}, \bibinfo{pages}{686--707}.
%Type = Article
\bibitem[{Rasheed et~al.(2020)Rasheed, San and Kvamsdal}]{rasheed2020dtv}
\bibinfo{author}{Rasheed, A.}, \bibinfo{author}{San, O.},
  \bibinfo{author}{Kvamsdal, T.}, \bibinfo{year}{2020}.
\newblock \bibinfo{title}{Digital twin: Values, challenges and enablers from a
  modeling perspective}.
\newblock \bibinfo{journal}{IEEE Access} \bibinfo{volume}{8},
  \bibinfo{pages}{21980--22012}.
\newblock \DOIprefix\doi{https://doi.org/10.1109/ACCESS.2020.2970143}.
%Type = Misc
\bibitem[{Robinson et~al.(2022)Robinson, Pawar, Rasheed and
  San}]{haakon2022pgnn}
\bibinfo{author}{Robinson, H.}, \bibinfo{author}{Pawar, S.},
  \bibinfo{author}{Rasheed, A.}, \bibinfo{author}{San, O.},
  \bibinfo{year}{2022}.
\newblock \bibinfo{title}{Physics guided neural networks for modelling of
  non-linear dynamics}.
\newblock \URLprefix \url{https://arxiv.org/abs/2205.06858},
  \DOIprefix\doi{10.48550/ARXIV.2205.06858}.
%Type = Article
\bibitem[{San et~al.(2021)San, Rasheed and Kvamsdal}]{san2021hybrid}
\bibinfo{author}{San, O.}, \bibinfo{author}{Rasheed, A.},
  \bibinfo{author}{Kvamsdal, T.}, \bibinfo{year}{2021}.
\newblock \bibinfo{title}{Hybrid analysis and modeling, eclecticism, and
  multifidelity computing toward digital twin revolution}.
\newblock \bibinfo{journal}{GAMM-Mitteilungen} \bibinfo{volume}{44},
  \bibinfo{pages}{e202100007}.
\newblock \DOIprefix\doi{https://doi.org/10.1002/gamm.202100007}.
%Type = Article
\bibitem[{Vaddireddy et~al.(2020)Vaddireddy, Rasheed, Staples and
  San}]{vaddireddy2020fes}
\bibinfo{author}{Vaddireddy, H.}, \bibinfo{author}{Rasheed, A.},
  \bibinfo{author}{Staples, A.E.}, \bibinfo{author}{San, O.},
  \bibinfo{year}{2020}.
\newblock \bibinfo{title}{Feature engineering and symbolic regression methods
  for detecting hidden physics from sparse sensors}.
\newblock \bibinfo{journal}{Physics of Fluids, Editor's pick}
  \bibinfo{volume}{32}, \bibinfo{pages}{015113}.
\newblock \DOIprefix\doi{https://doi.org/10.1063/1.5136351}.
%Type = Article
\bibitem[{Xiang et~al.(2022)Xiang, Lee, Zikanov and Hsu}]{XIANG2022117641}
\bibinfo{author}{Xiang, L.}, \bibinfo{author}{Lee, C.W.},
  \bibinfo{author}{Zikanov, O.}, \bibinfo{author}{Hsu, C.C.},
  \bibinfo{year}{2022}.
\newblock \bibinfo{title}{Efficient reduced order model for heat transfer in a
  battery pack of an electric vehicle}.
\newblock \bibinfo{journal}{Applied Thermal Engineering} \bibinfo{volume}{201},
  \bibinfo{pages}{117641}.
\newblock \URLprefix
  \url{https://www.sciencedirect.com/science/article/pii/S135943112101067X},
  \DOIprefix\doi{https://doi.org/10.1016/j.applthermaleng.2021.117641}.
%Type = Article
\bibitem[{Xu et~al.(2021)Xu, Zhang and Wang}]{XU2021110592}
\bibinfo{author}{Xu, H.}, \bibinfo{author}{Zhang, D.}, \bibinfo{author}{Wang,
  N.}, \bibinfo{year}{2021}.
\newblock \bibinfo{title}{Deep-learning based discovery of partial differential
  equations in integral form from sparse and noisy data}.
\newblock \bibinfo{journal}{Journal of Computational Physics}
  \bibinfo{volume}{445}, \bibinfo{pages}{110592}.
\newblock \URLprefix
  \url{https://www.sciencedirect.com/science/article/pii/S0021999121004873},
  \DOIprefix\doi{https://doi.org/10.1016/j.jcp.2021.110592}.

\end{thebibliography}
\end{document}